\documentclass{article}

\usepackage[preprint]{neurips_2026}

\usepackage[utf8]{inputenc} 
\usepackage[T1]{fontenc}    
\usepackage{hyperref}       
\usepackage{url}            
\usepackage{booktabs}       
\usepackage{amsfonts}       
\usepackage{nicefrac}       
\usepackage{microtype}      
\usepackage{xcolor}         
\usepackage{graphicx}
\usepackage{tabularx}
\usepackage{amsmath}
\usepackage[textsize=tiny]{todonotes}

\title{On the Robustness of Multilingual Text Embedding Rankings Across Learning Tasks, Languages, and Benchmark Datasets}

\author{
  Ana Gjorgjevikj\\
  Computer Systems Department\\
  Jožef Stefan Institute\\
  Ljubljana, Slovenia\\
  \texttt{ana.gjorgjevikj@ijs.si} \\
  \AND
  Barbara Koroušić Seljak \\
  Computer Systems Department\\
  Jožef Stefan Institute\\
  Ljubljana, Slovenia\\
  \texttt{barbara.korousic@ijs.si} \\
  \AND
  Tome Eftimov \\
  Computer Systems Department\\
  Jožef Stefan Institute\\
  Ljubljana, Slovenia\\
  \texttt{tome.eftimov@ijs.si} \\
}

\begin{document}

\maketitle

\begin{abstract}
Large-scale multilingual text embedding models play crucial role in both research and industry, yet their behavior in language-specific, multi-task settings remains insufficiently understood. Although benchmarking platforms such as MTEB report results across more than 250 languages, conclusions about model superiority often depend on implicit choices of dataset compositions and performance aggregation methods. To address this gap, we present a meta-study of multilingual model performance robustness in MTEB, applying a diverse set of multi-criteria decision-making ranking schemes and introducing two robustness indicators: dataset-composition robustness (sensitivity of rankings to changing dataset compositions) and ranking-scheme robustness (sensitivity to aggregation method change). They enable systematic sensitivity analysis of whether benchmarking conclusions remain stable under different evaluation designs. We conduct an in-depth analysis on five languages (English, French, German, Hindi, and Spanish) across nine tasks (e.g., classification, clustering, retrieval) and release results for approximately 230 additional languages. The task-specific analyses show that large-scale LLM-based models are often robust top performers, though not uniformly (e.g., in retrieval task), while task-agnostic results reveal that only a small subset of models remains consistently strong across tasks, ranking schemes, and data subsamples.

\end{abstract}
\section{Introduction}
\label{sec:intro}

Text embeddings constitute a core component of modern natural language processing (NLP) systems, underpinning tasks\footnote{MTEB uses ``task" to refer to a dataset with its own evaluation protocol and groups tasks in ``task categories" (e.g., STS is a task category). In this paper, we use ``task" to denote the MTEB's task category and ``dataset" to denote MTEB's task.} such as classification, information retrieval, and semantic textual similarity (STS). They are also integral to retrieval-augmented generation (RAG) pipelines, in which embeddings are used to retrieve relevant external knowledge prior to large language model (LLM) output generation~\cite{zhao2026retrieval}. As retrieval increases computational cost and latency~\cite{huang2025embedding}, understanding which embedding models are suitable for the target task and language is crucial to achieve efficiency and effectiveness.

Despite rapid advances, learning embedding models that generalize reliably across both tasks and languages remains a significant challenge. Model performance can vary substantially across downstream tasks and linguistic settings, even when evaluated under comparable conditions. Linguistic diversity in syntactic and semantic structure, shaped by historical and cultural factors~\cite{enevoldsen2025mmteb}, complicates cross-lingual generalization. Although multilingual embedding models facilitate knowledge transfer from high-resource to low-resource languages, such transfer is not consistently reliable, and performance gains observed in one language may not generalize to others~\cite{enevoldsen2024scandinavian}. However, before discussing generalization, a crucial missing aspect is assessing how robust the performance gains reported in benchmarking studies are when the datasets and evaluation strategies change. Understanding this robustness is essential to properly set up and interpret subsequent generalization experiments.

Large-scale multilingual benchmarks such as MTEB~\cite{enevoldsen2025mmteb} provide performance evaluations across numerous tasks, datasets, and languages. However, they face a fundamental aggregation challenge: datasets are evaluated using heterogeneous metrics (e.g., accuracy, v-measure, nDCG), yet model rankings are typically derived via simple averaging. Such averaging implicitly assumes metric comparability, neglects dataset correlation, and can bias rankings toward overrepresented tasks/languages. These issues are exacerbated in multilingual settings, where dataset availability and task coverage differ substantially across languages. Consequently, average-based leaderboards may obscure meaningful differences in language-specific robustness and cross-task generalization. Recent studies use rank aggregation approaches inspired by social choice theory, in which datasets are treated as voters and models as candidates~\cite{colombo2022best,rofin2023vote}. While conceptually appealing, they generally assume equal relevance of datasets and do not account for correlations or application-driven preferences. In practice, datasets differ in linguistic properties and task difficulty, making robust model comparison non-trivial.

This work studies the robustness of benchmarking results for multilingual text embedding models and makes two contributions to multilingual evaluation methodologies. First, we introduce two \emph{robustness indicators} that quantify previously unmeasured sources of evaluation uncertainty. \emph{Dataset-composition robustness} measures the sensitivity of model rankings to the choice of benchmark datasets via correlation-aware subsampling. \emph{Ranking-scheme robustness} measures the sensitivity of rankings to the choice of score aggregation methodology by comparing multiple multi-criteria decision-making (MCDM) methods. Together, these indicators distinguish whether observed rankings reflect genuine model performance or artifacts of evaluation design (an issue not captured by standard leaderboard averages). Second, we apply these indicators at the \emph{language} level, analyzing robustness both within tasks (language-specific, task-specific) and across tasks (language-specific, task-agnostic). This yields, to our knowledge, the first systematic study of robustness in multilingual embedding benchmarks at language granularity. We evaluate our framework on MTEB Multilingual v2~\cite{enevoldsen2025mmteb}, covering 500+ datasets and 250+ languages, with detailed analyses for five languages (English, French, German, Hindi, and Spanish) and released artifacts for approximately 230 languages.

\section{Preliminaries}
\label{sec:preliminaries}

\textbf{Multilingual Text Embedding Models.}
Text embedding models map variable-length text to fixed-dimensional vectors, enabling similarity computation, clustering, and downstream prediction. Contemporary multilingual models broadly belong to two families. \textbf{Encoder-based embedding models} are typically based on multilingual Transformer encoders (e.g., XLM-RoBERTa) trained using contrastive objectives and instruction-style supervision. \textbf{LLM-derived embedding models} adapt decoder-only LLMs into embedding models via contrastive and instruction tuning, often using pooling over hidden states. This includes models unifying generation and embedding, e.g., \textit{GritLM}~\cite{muennighoff2024generative}, and such converting pretrained LLMs into encoders with minimal architectural changes, e.g., \textit{LLM2Vec}~\cite{behnamghader2024llm2vec}.

\textbf{Multilingual Embedding Benchmarks and Leaderboards.}
Multilingual embedding models are commonly evaluated using benchmark suites and public leaderboards. MTEB introduced a unified evaluation framework spanning multiple tasks and datasets, showing that no single model consistently dominates across all tasks~\cite{muennighoff2022mteb}. Its multilingual extension, MTEB Multilingual (MMTEB), increases language coverage and task diversity, enabling fine-grained language--task evaluation~\cite{enevoldsen2025mmteb}. In this work, we use MMTEB leaderboard v2 as a source of per-dataset performance scores. As multilingual benchmarks aggregate scores across datasets and tasks that rely on heterogeneous evaluation metrics, the model rankings depend critically on the aggregation procedure used to derive them. The dominant practice is simple average across datasets, which is sensitive to dataset redundancy and task imbalance. Alternative approaches based on social choice theory treat datasets as voters and models as candidates~\cite{colombo2022best,rofin2023vote,enevoldsen2025mmteb}. In this work, we adopt a complementary strategy. Instead of committing to a single aggregation rule, we generate multiple rankings using diverse MCDM schemes and analyze their sensitivity to both ranking methodology and dataset composition changes through correlation-aware subsampling.

\section{Methodology}

\subsection{Language-Aware Problem Formulation}

Let $\mathcal{L}=\{1,\ldots,L\}$ denote the set of languages and
$\mathcal{T}=\{1,\ldots,Q\}$ the set of downstream tasks (e.g., classification,
retrieval). For a task $q \in \mathcal{T}$ and language $\ell \in \mathcal{L}$,
we evaluate $m_\ell$ embedding models on $n_{q,\ell}$ datasets
written in language $\ell$. The evaluation results are summarized by a
performance matrix $P^{(q,\ell)} \in \mathbb{R}^{m_\ell \times n_{q,\ell}}$, where $P^{(q,\ell)}_{i,j}$ is the performance of model $i$ on dataset
$j$ according to a task-specific metric (e.g., accuracy, v-measure).
Rows correspond to models and columns to datasets.

\paragraph{Ranking Schemes.}
To compare models, we consider $K$ ranking schemes (e.g., different
MCDM methods). Each scheme defines a ranking
operator $R^{(s)} : \mathbb{R}^{m \times n} \rightarrow \{1,\ldots,m\}^{m}$, which maps a performance matrix to a vector of model ranks
(smaller ranks indicate better performance). For task $q$ and language $\ell$, the ranking produced by scheme $s$ is $r^{(s,q,\ell)} = R^{(s)}(P^{(q,\ell)})$.

\paragraph{Dataset Subsampling.}
To analyze robustness with respect to dataset composition,
we construct subsamples of datasets with low pairwise
correlation across model performance profiles. For dataset $j$, we define its performance profile across models
as column vector $p^{(q,\ell)}_j = P^{(q,\ell)}_{\cdot,j} \in \mathbb{R}^{m_\ell}$. Datasets whose performance profiles are strongly correlated provide redundant information. Therefore, those with correlation larger than a threshold $\tau$ are clustered together. From each cluster we randomly select one
representative dataset, generating multiple uncorrelated subsamples. Let $\mathcal{S}^{(q,\ell)}=\{1,\ldots,S_{q,\ell}\}$ index these subsamples. Each subsample defines a restricted performance matrix $P^{(q,\ell,t)} \in \mathbb{R}^{m_\ell \times n_{q,\ell,t}}$, obtained by selecting the corresponding columns of the complete performance matrix $P^{(q,\ell)}$.

\subsection{Multi-Ranking Across Schemes and Dataset Subsamples}

For each ranking scheme $s$ and subsample $t$, model ranks are computed as
$r^{(s,q,\ell,t)} = R^{(s)}(P^{(q,\ell,t)})$. This produces a family of rankings for each task--language pair $(q,\ell)$,
allowing us to analyze ranking robustness under different evaluation
conditions. We then evaluate ranking robustness with respect to two sources of variation:
ranking scheme choice and dataset composition, where low variation (sensitivity to changes) indicate stabile rankings across ranking schemes and dataset compositions.

\paragraph{Sensitivity Across Ranking Schemes.}

For a fixed dataset subsample $t$, we measure how sensitive model
rankings are to ranking scheme change through the rank variance across schemes as per Eq.~\ref{eq:sn_schemes}.

\begin{equation}
\label{eq:sn_schemes}
\mathrm{ST}^{(q,\ell,t)}(i)
=
\mathrm{Var}_{s=1,\ldots,K}
\left(r^{(s,q,\ell,t)}_i\right)
\end{equation}

\paragraph{Sensitivity to Dataset Composition.}

For a fixed ranking scheme $s$, we measure the sensitivity
of rankings to dataset composition by computing the variance of ranks
across subsamples with Eq.~\ref{eq:sn_datasets}.

\begin{equation}
\label{eq:sn_datasets}
\mathrm{SN}^{(q,\ell,s)}(i)
=
\mathrm{Var}_{t \in \mathcal{S}^{(q,\ell)}}
\left(r^{(s,q,\ell,t)}_i\right)
\end{equation}

\subsection{Language-Specific Cross-Task Consistency}

While previous analysis compares models within a task and
language, we also evaluate model consistency across tasks within a
language. 

For each ranking $r^{(s,q,\ell,t)}$, we define the set of top-$\eta$
models as $\mathcal{M}^{(s,q,\ell,t)}_\eta = \{ i \mid r^{(s,q,\ell,t)}_i \le \eta \}$. Let $\mathcal{U}^{(\ell)}_\eta$ denote the set of models that appear
in the top-$\eta$ list for at least one task, scheme, or subsample
within language $\ell$. For each model $i$ we define a cross-task consistency score with Eq.~\ref{eq:consistency_score}, which measures in how many tasks the model consistently appears
among the top-$\eta$ models.

\begin{equation}
\label{eq:consistency_score}
\mathrm{CT}^{(s,\ell)}(\eta,i)
=
\sum_{q=1}^{Q}
\frac{1}{|\mathcal{S}^{(q,\ell)}|}
\sum_{t \in \mathcal{S}^{(q,\ell)}}
\mathbf{1}\!\left(i \in \mathcal{M}^{(s,q,\ell,t)}_\eta\right)
\end{equation}

\section{Experimental Design}

The experimental data used in this paper are obtained from the MTEB Multilingual leaderboard v2\footnote{https://huggingface.co/spaces/mteb/leaderboard}, accessed on Dec 26, 2025\footnote{Results correspond to the MTEB results repository snapshot (commit \texttt{1ddac93}).}. The benchmark provides performance scores for embedding models across nine learning tasks: (1) classification, (2) clustering, (3) retrieval, (4) reranking, (5) pair classification, (6) STS, (7) multiclass classification, (8) instruction reranking, and (9) bitext mining. Based on the available model metadata, we restrict our analysis to open and 100\% zero-shot models within tasks, indicating that they were not trained on instances drawn from the corresponding task datasets\footnote{The MTEB scores should be interpreted as a metric that offers a coarse estimate of how ``in-domain” or ``out-of-domain” a model is with respect to the benchmark~\cite{chung2025maintaining}.}. Each dataset is associated with one or more languages, and we leverage this metadata to select datasets to be analyzed for each language. The distribution of datasets by language-task is reported in Appendix~\ref{appendix:dataset_distribution} and the project repository. These statistics indicate that out of 1,037 languages, 798 are associated with only a single dataset and are therefore excluded from our analysis. We analyze the remaining approximately 230 languages, represented by at least two datasets.

For each task–language pair, we vary the correlation threshold $\tau \in \{0.8, 0.85, 0.9\}$. Two datasets having a correlation above $\tau$ are considered correlated and assigned to the same cluster. Thresholds below the tested range frequently collapse datasets into a single cluster, as many dataset pairs satisfy the correlation criterion and are merged together, thus preventing meaningful analysis. In the main paper, we report results for $\tau = 0.9$, while for sensitivity analyses with the other thresholds, see Appendix~\ref{sec:sensitivty_correlation}. To obtain an uncorrelated subset of datasets for robustness analysis, we perform random sampling by selecting one dataset from each cluster (repeated three times), resulting in three dataset compositions. The correlation threshold determines how datasets are clustered, allowing robustness analysis of rankings with respect to dataset compositions. 
Practical guidelines for the selection of the parameter $\tau$ are given in Appendix~\ref{appendix:hyperparameters}. The sensitivity analysis (Appendix~\ref{sec:sensitivty_correlation}) shows that the ranking procedure remains stable across correlation threshold variations, indicating that the underlying signal driving the rankings is largely independent of the specific clustering configuration.

To rank models within a specific language, task, and dataset composition, we apply four MCDM methods, i.e., Weighted Sum Model (WSM), Technique for Order Preference by Similarity to an Ideal Solution (TOPSIS)~\cite{chen1992fuzzy}, VlseKriterijumska Optimizacija I Kompromisno Resenje (VIKOR)~\cite{opricovic1998multicriteria}, and Ranking Organization Method for Enrichment Evaluations (PROMETHEE II)~\cite{brans1982ingenierie}. They originate from distinct methodological categories, ensuring ranking schemes diversity. For details on methods and their complementarity, see Appendices~\ref{appendix:rankingMethods},~\ref{appendix:weightingMethods}, and~\ref{app:portfolio}. PROMETHEE II is used with two preference functions, usual and Gaussian. We combine the MCDM methods with three objective criteria-weighting approaches, i.e., Equal Weighting, the Method based on the Removal Effects of Criteria (MEREC)~\cite{keshavarz2021determination}, and CRiteria Importance Through Inter-criteria Correlation (CRITIC)~\cite{diakoulaki1995determining}, which compute criteria weights from performance matrix. This gives 15 ranking schemes by combining five MCDM methods (WSM, TOPSIS, VIKOR, PROMETHEE II–usual, PROMETHEE II–Gaussian) with three weighting strategies (Equal, MEREC, CRITIC). Here criteria correspond to datasets in a given language, task, and dataset composition. For a guide on selecting ranking schemes, see Appendix~\ref{appendix:hyperparameters}. The experiments run on Intel Core i5-11320H CPU, with default system configuration.

\section{Results}
\label{sec:results}

\subsection{Language-Specific, Task-Specific Multi-Ranking}

Model robustness is evaluated with respect to both dataset composition (DS) and ranking scheme (RS), which we denote as full robustness. When only a single dataset is available, or when datasets do not meet the correlation threshold and thus form separate clusters for a given language–task pair, robustness is assessed only across ranking schemes. This quantifies how model rankings change with the choice of benchmark datasets and/or evaluation methodology.

\begin{figure*}[!ht]
    \centering
    \includegraphics[width=0.85\linewidth]{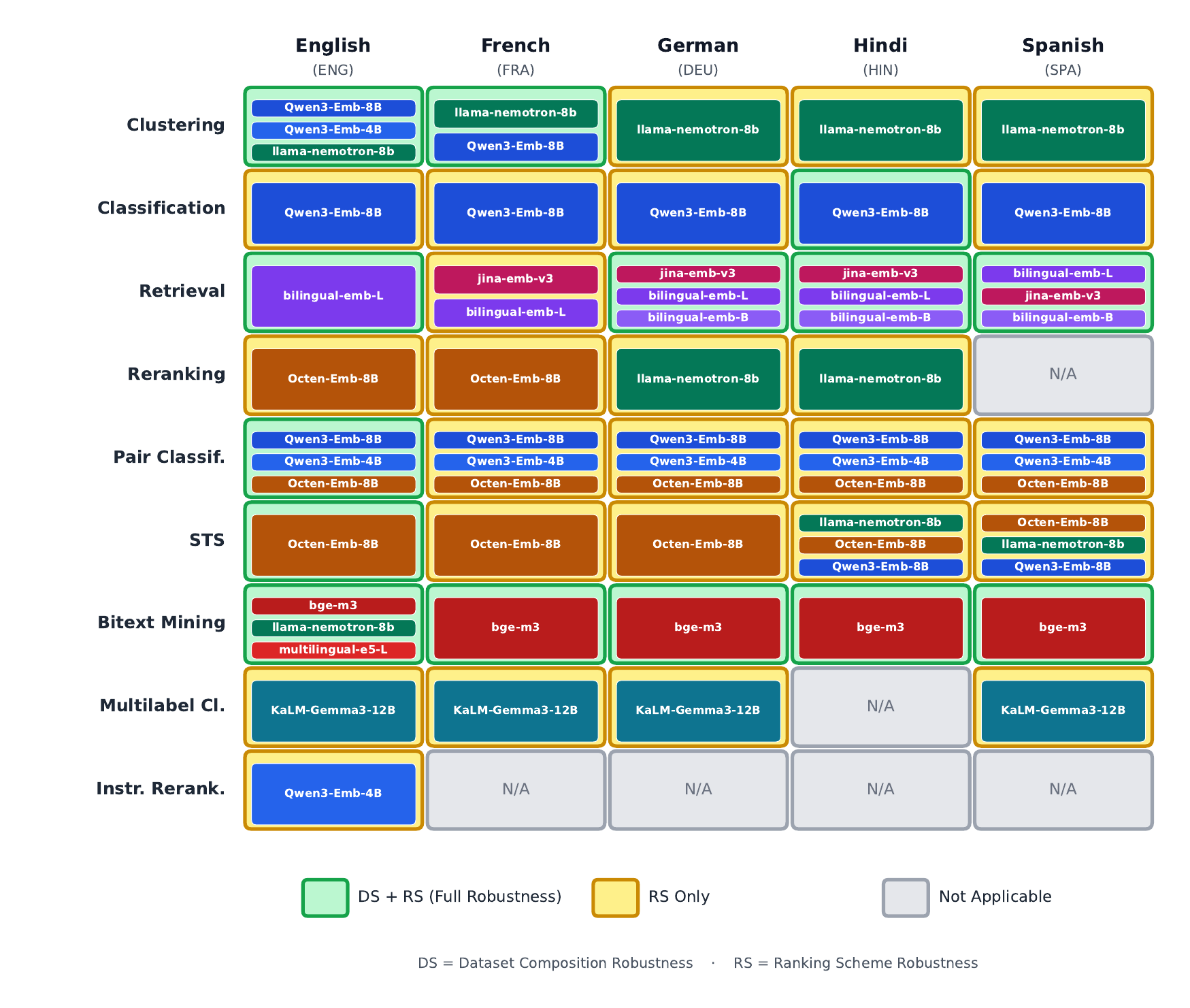}
    \caption{\small Most robust models in 9 tasks (rows) and 5 languages (columns). Bitext mining and retrieval achieve full robustness in most languages. Instruction reranking has limited cross-lingual coverage (English only).}
    \label{fig:task_summary}
\end{figure*}

Figure~\ref{fig:task_summary} presents a comprehensive overview of the most robust models for each task-language combination. Several patterns emerge from the analysis. First, model \textit{Qwen3-Embedding-8B} exhibits remarkable consistency across classification-oriented tasks with regard to the RS robustness, achieving top performance in classification and pair classification across all five languages. Second, for retrieval, \textit{bilingual-embedding-large} shows the strongest robustness, appearing as the sole fully robust model for English and consistently ranking among the top models for other languages. Third, \textit{bge-m3} dominates bitext mining with full robustness across all five languages, indicating highly stable rankings regardless of dataset composition or ranking scheme. Fourth, task coverage varies considerably: bitext mining achieves complete cross-lingual coverage with full robustness, while instruction reranking is limited to English only. Finally, \textit{llama-embed-nemotron-8b} emerges as a versatile model, appearing across multiple tasks like clustering, reranking, STS, and bitext mining. For a practitioner decision rule based on Figure~\ref{fig:task_summary}, see Appendix~\ref{appendix:reading_results}. Next, we present detailed results for clustering in French; results for English, German, Hindi, and Spanish are omitted for space and provided in the repository, with a discussion in Appendix~\ref{appendix:results}.

\noindent\textbf{Clustering -- French (FRA).} \textbf{\textit{Dataset compositions:}} The clustering task includes four datasets, which form two clusters of uncorrelated datasets: [MasakhaNEWSClusteringS2S, AlloProfClusteringS2S.v2, HALClusteringS2S.v2] and [SIB200ClusteringS2S]. The first contains three highly correlated datasets, whereas the second consists of a single dataset. This results in four dataset compositions: one using all datasets and three subsamples, each formed by randomly selecting one dataset from the first cluster (with fixed random seeds) and combining it with the second cluster.

\textbf{\textit{Model rankings:}} 
Figure~\ref{fig:fra_clustering} (see Appendix~\ref{appendix:fra_clustering_stability_sensitivity}) shows three model clusters based on rankings across all ranking schemes and dataset compositions, two of which further decompose into three subclusters each. The \textbf{best-performing cluster} consists of models spanning from \textit{llama-embed-nemotron-8b} to \textit{multilingual-e5-large-instruct}. Semantically, this cluster captures high-capacity multilingual LLM-based embedding models trained with contrastive and instruction-tuning objectives to produce robust, language-agnostic semantic representations. The first subcluster includes \textit{llama-embed-nemotron-8b} and \textit{Qwen3-Embedding-8B}, both 8B-parameter LLM-based models that produce dense 4,096-dimensional embeddings via hidden-state pooling strategies and are optimized through advanced contrastive learning on multilingual and multi-task synthetic datasets. The former belongs to NVIDIA’s Nemotron family and is based on the Llama-3.1-8B decoder-only transformer architecture; it is instruction-tuned using a contrastive learning objective on a custom dataset of 16M query–document pairs to excel in challenging multilingual settings, particularly for cross-lingual and low-resource language use cases~\cite{babakhin2025llama}. The second model belongs to the Qwen family, based on the Qwen3 LLM, combining a multi-stage training strategy with model merging to improve multilingual and multi-task generalization~\cite{zhang2025qwen3}. Both models exhibit low sensitivity across ranking schemes for all dataset compositions (Figure~\ref{fig:fra_clustering_st}, Appendix~\ref{appendix:fra_clustering_stability_sensitivity}) and low sensitivity to dataset composition across all ranking schemes (Figure~\ref{fig:fra_clustering_sn}, Appendix~\ref{appendix:fra_clustering_stability_sensitivity}), making them strong candidates for clustering French text. The second-best subcluster comprises \textit{Octen-Embedding-8B} and \textit{multilingual-e5-large-instruct}, i.e., a large 8B-parameter Qwen3-based model and a mid-sized 560M-parameter e5-based model, respectively. The former fine-tunes the \textit{Qwen3-Embedding-8B} model for semantic search and retrieval~\cite{octen-embedding-2025}. The latter is based on the multilingual XLM-RoBERTa-Large encoder-only transformer and is instruction-tuned on a large-scale multilingual dataset in 93 languages~\cite{wang2024multilingual}. It exhibits slightly higher sensitivity across ranking schemes for all dataset compositions, while both models show minimal sensitivity to dataset composition across nearly all schemes. The third subcluster comprises \textit{Qwen3-Embedding-4B} and \textit{GritLM-8x7B}. The former is a smaller-scale 4B-parameter variant of one of the top-performing models, \textit{Qwen3-Embedding-8B}, while the latter fine-tunes the Mixtral-8x7B LLM using generative/representational instruction tuning~\cite{muennighoff2024generative}.

\begin{figure}[!ht]
    \centering
    \includegraphics[width=0.35\linewidth]{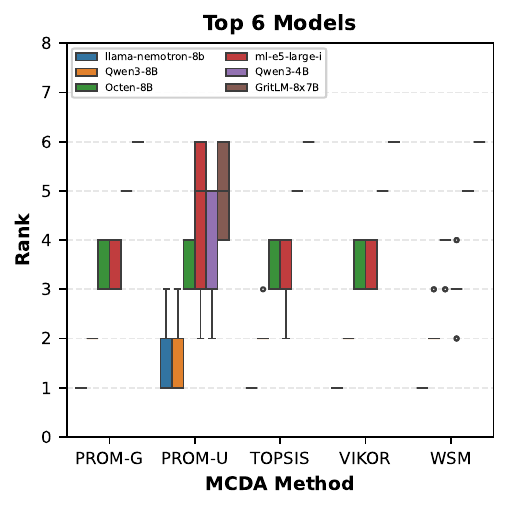}
    \caption{\small Ranking sensitivity of top-6 models across MCDM schemes for clustering in French. Boxplots show rank distributions across MCDM methods, each with three weighting schemes and three dataset compositions.}
    \label{fig:clustering_french_top6}
\end{figure}

\textbf{The second large cluster} groups mid-capacity LLM-derived embedding models, mostly built on Mistral-7B backbone. They provide solid semantic representations but show limited robustness across ranking schemes and greater sensitivity to dataset composition, especially with uncorrelated subsamples. This cluster spans from \textit{SFR-Embedding-Mistral} to \textit{SFR-Embedding-2\_R}. The four Mistral-7B-v0.1-based models (\textit{SFR-Embedding-Mistral}, \textit{e5-mistral-7b-instruct}, \textit{Linq-Embed-Mistral}, \textit{GritLM-7B}) achieve mid-range rankings, with lower ranking scheme change sensitivity under full dataset compositon than under subsamples; among them, \textit{Linq-Embed-Mistral} is more sensitive to dataset composition. \textit{Bilingual-embedding-large} and \textit{SFR-Embedding-2\_R} form a subcluster with lower rankings. \textbf{The third cluster} contains \textit{multilingual-e5-large}, which shows low rankings and high variability across schemes and dataset compositions. Figure~\ref{fig:clustering_french_top6} shows the top-6 models for French clustering (rank distribution, n=3 dataset compositions), where \textit{llama-nemotron-8b} and \textit{Qwen3-8B} consistently rank highest with narrow rank deviations, indicating strong robustness. For extended figures, see Appendix~\ref{appendix:fra_clustering_stability_sensitivity}. Table~\ref{fig:summary_clustering} (Appendix~\ref{appendix:results}) gives the most robust models for clustering in French.

\noindent\textbf{Clustering -- English (ENG).} \textbf{\textit{Dataset compositions:}} In English, the clustering task comprises nine datasets, which are partitioned into four clusters based on correlation analysis. The first includes five datasets ([WikiCitiesClustering, ArXivHierarchicalClusteringP2P, ArXivHierarchicalClusteringS2S, BiorxivClusteringP2P.v2, and MedrxivClusteringP2P.v2]). Each of the remaining contains a single dataset ([StackExchangeClustering.v2], [SIB200ClusteringS2S], [MasakhaNEWSClusteringS2S], and [BigPatentClustering.v2]), resulting in four different dataset compositions.

\textbf{\textit{Model rankings:}} The \textbf{best-performing cluster} includes \textit{Qwen3-Embedding-8B}, \textit{Qwen3-Embedding-4B}, and \textit{llama-embed-nemotron-8b}. They show low sensitivity to both ranking schemes (especially \textit{Qwen3-Embedding-8B}) and dataset composition, indicating robust rankings. Semantically, they represent large LLM-based embedding models using high-dimensional representations and contrastive instruction tuning to learn stable embedding spaces. \textbf{The second-best cluster} consists of large LLM-based models such as \textit{Octen-Embedding-8B}, \textit{SFR-Embedding-Mistral}, \textit{SFR-Embedding-2\_R}, \textit{GritLM-8x7B}, and \textit{multilingual-e5-large-instruct}. While some (e.g., \textit{Octen-Embedding-8B}, \textit{multilingual-e5-large-instruct}) achieve high rankings under certain schemes, they exhibit greater sensitivity across schemes than the top cluster. This suggests that, although effective, their performance depends more on the evaluation method and is less consistently aligned across different evaluation settings. \textbf{The third cluster} includes includes the Mistral-7B-based models \textit{Linq-Embed-Mistral}, \textit{e5-mistral-7b-instruct}, and \textit{GritLM-7B}, the smaller-scale \textit{Qwen3-Embedding-0.6B} model, the smaller-scale Gemma 3-based \textit{embeddinggemma-300m} and XLM-RoBERTa-based \textit{jina-embeddings-v3}. These models show mid-range rankings with generally low variability (except \textit{embeddinggemma-300m}). They are smaller-scale or mid-capacity encoders that trade performance for stability across ranking schemes.

\noindent\textbf{Clustering -- German (DEU), Spanish (SPA), Hindi (HIN):} Only a single dataset is available, so models are ranked directly by performance on that dataset. The top model is \textit{llama-embed-nemotron-8b}, followed by \textit{multilingual-e5-large-instruct} and Qwen3-based models (\textit{Qwen3-Embedding-8B}, \textit{Octen-Embedding-8B}, \textit{Qwen3-Embedding-4B}). Overall, the best-performing models rely on large LLM backbones (e.g., Llama-3.1-8B, Qwen3, Mixtral-8x7B, Mistral-7B). Results for Hindi and Spanish are similar due to the same single-dataset setting, with \textit{llama-embed-nemotron-8b} consistently ranking first, followed by \textit{multilingual-e5-large-instruct}, \textit{Qwen3-Embedding-8B}, and \textit{Octen-Embedding-8B}. The most robust models per language are given in Figure~\ref{fig:task_summary} and Table~\ref{fig:summary_clustering} (Appendix~\ref{appendix:fra_clustering_stability_sensitivity}).

\noindent\textbf{Analyses of the remaining eight tasks:} Appendix~\ref{appendix:results} provides a detailed discussion of the results for the remaining eight tasks (classification, retrieval, reranking, pair classification, STS, bitext mining, multilabel classification, and instruction reranking) across the five selected languages. Figure~\ref{fig:task_summary} presents the results of those analyses. In addition, Tables~\ref{fig:summary_classifiaction}–\ref{fig:summary_instruction_reranking} from Appendix~\ref{appendix:results} summarize the most robust models by task and by language, also including the robustness criteria assessed (DS and RS).

\begin{figure*}[!ht]
    \centering
    \includegraphics[width=0.8\linewidth]{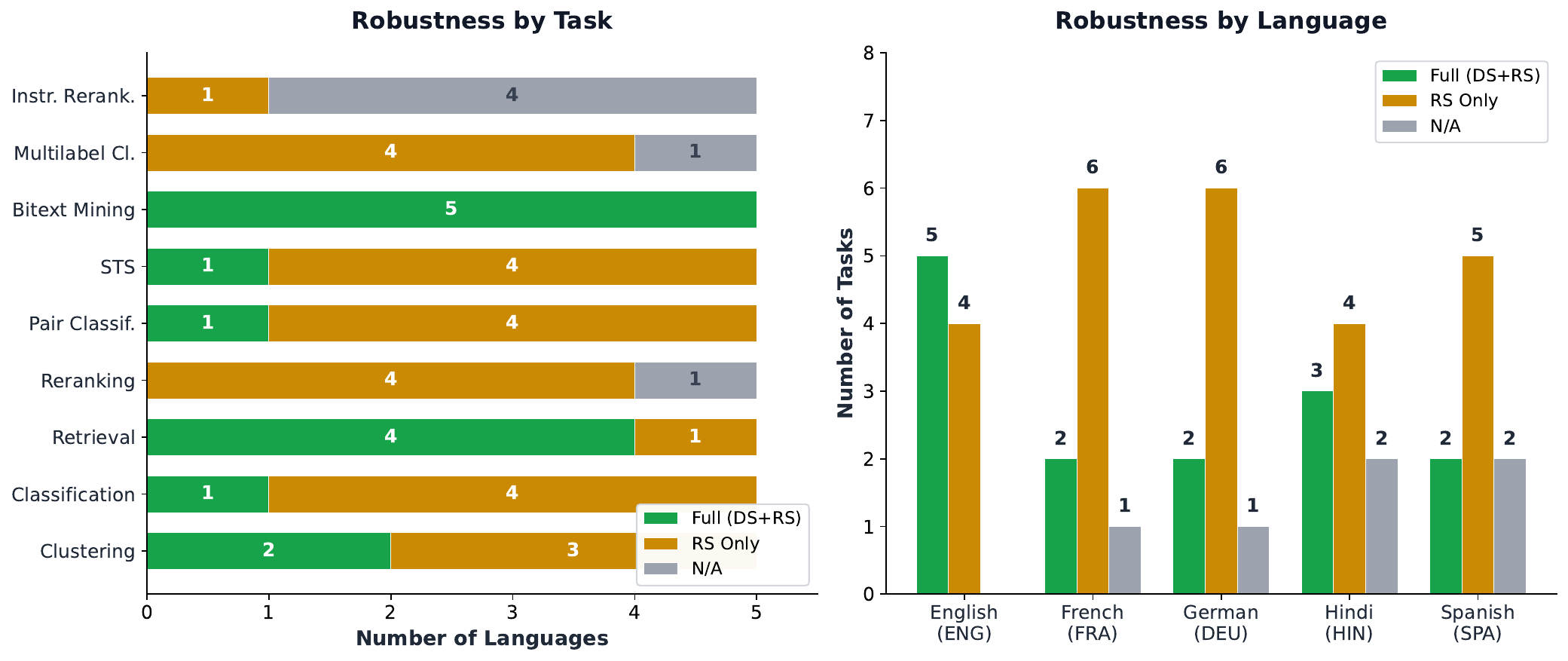}
    \caption{\small Robustness distribution across tasks (left) and languages (right). Stacked horizontal bars show the number of languages achieving full robustness (green), ranking scheme robustness only (yellow), or not applicable (gray) for each task. Grouped vertical bars compare robustness levels across languages. Full robustness (DS+RS) indicates stable model rankings under both dataset subsampling and ranking schemes.}
    \label{fig:robustness_summary}
\end{figure*}

\noindent\textbf{Robustness overview:} Figure~\ref{fig:robustness_summary} shows systematic differences in robustness across tasks and languages. From task perspective, a clear hierarchy emerges: bitext mining achieves full robustness in all five languages, making it the most reliable task for cross-task evaluation, followed by retrieval with four fully robust languages. In contrast, six of nine tasks achieve full robustness for at most one language, and instruction reranking lacks sufficient multilingual coverage. From the language perspective, English shows the strongest robustness (five fully robust tasks), benefiting from mature and diverse benchmark datasets, followed by Hindi with three. French and German, despite coverage across eight tasks, achieve full robustness in only two, with the remaining tasks showing RS-only robustness due to insufficient dataset diversity. Spanish follows a similar pattern, with two fully robust tasks and limited data for assessing DS robustness elsewhere. At present, estimating the required number of datasets is difficult, as robustness depends on both quantity and diversity. A practical heuristic is to include multiple dataset clusters with several datasets per cluster to enable meaningful substitutions. Determining the necessary dataset diversity for full robustness remains an open research direction. Overall, current multilingual benchmarks provide the most reliable comparisons for English, with more limited reliability for other languages depending on the task.

\noindent\textbf{Results summary:} Three distinct robustness profiles emerge: (i) \textbf{fully robust tasks} where rankings are stable across both dataset compositions and ranking schemes - bitext mining achieves this for all five languages, while retrieval achieves it for four; (ii) \textbf{ranking-stable tasks} where different ranking methodologies agree but dataset composition has not been assessed due to limited datasets available (a single dataset or set of uncorrelated datasets) - classification, pair classification, STS, reranking, and multilabel classification fall into this category for most languages; and (iii) \textbf{coverage-limited tasks} where insufficient benchmark data prevents robust evaluation - instruction reranking is evaluable only for English, and multilabel classification lacks Hindi coverage. From a model perspective, clear task-specific leaders emerge: \textit{bge-m3} for bitext mining, \textit{Qwen3-Embedding-8B} for classification tasks, \textit{bilingual-embedding-large} for retrieval, and \textit{Octen-Embedding-8B} for STS. These findings enable practitioners to identify which models satisfy both robustness criteria.

\noindent\textbf{Drivers of robustness.}
Across tasks and languages, two factors consistently predict
higher robustness in our analysis. \textbf{(1) Training objective alignment.}
Models trained with contrastive or dense retrieval objectives
produce embedding spaces optimized for similarity-preserving tasks, which
makes their rankings less sensitive to dataset composition.
\textit{Llama-embed-nemotron-8b} is the clearest example: trained on 16M query-document pairs with a contrastive objective on a diverse
multilingual corpus~\cite{babakhin2025llama}, it achieves stable rankings
across all nine tasks in all five languages. \textit{Bge-m3}, optimized via self-knowledge distillation for dense, sparse, and multi-vector retrieval~\cite{chen2024bge}, dominates bitext mining with
full robustness in all five languages for the same reason: its training
objective is tightly aligned with the evaluation criterion. In contrast, models adapted from generative LLMs via lightweight instruction
tuning (e.g., \textit{Qwen3-Embedding-8B}) produce strong but slightly more
scheme-sensitive rankings, likely because their representations retain residual generative structure that is not uniformly aligned across all nine tasks. \textbf{(2) Training data breadth.} Models trained on broad, multilingual, and multi-task synthetic corpora generalize more stably across the nine evaluated tasks than models fine-tuned on narrow, task-specific corpora. This, together with the zero-shot model filter, explains the retrieval exception: the top-performing retrieval models
(\textit{bilingual-embedding-large}, \textit{jina-embeddings-v3}) are
specialized encoder-based models optimized for retrieval similarity. These two factors (training objective alignment and data breadth)
offer a basis for predicting which models are likely to be robust
before running the full analysis. They motivate future work on benchmark
design: evaluation suites should include datasets that stress-test both
factors independently.

\noindent\textbf{Comparison with a baseline:} We compare our results with the MMTEB v2 leaderboard as a baseline (Appendix~\ref{sec:baseline}), observing variation in agreement across tasks, which motivates our robustness analysis. The highest and most stable correlations occur in bitext mining (0.85–0.98) and classification (0.93–0.97), while retrieval (0.50–0.92) and STS (0.50–0.97) show the lowest and most variable agreement. Clustering (0.45–0.72 for English), pair classification (0.65–0.77), and multilabel classification (down to 0.57) exhibit moderate to low stability. French, German, and Spanish show the most consistent agreement, Hindi moderate variability, and English the lowest correlations overall.

\subsection{Language-Task Agnostic Analysis (Task-Agnostic Within a Language)}

In the task-agnostic analysis, we assess model robustness across all nine tasks per language. We include all tasks, regardless of whether they show full or RS-only robustness, reflecting current benchmark availability; future work will restrict this to fully robust tasks and separate RS-only cases. Thus, current conclusions apply primarily to RS robustness, as not all tasks provide sufficient data to evaluate DS robustness. We set $\eta=10$, considering the union of top-10 models across ranking schemes when computing cross-task consistency (CT) (higher values indicate greater robustness). Results are presented as dendrograms (cross-consistency heatmaps), with rows as models and columns as ranking schemes and dataset compositions (full datasets first, followed by uncorrelated subsets; available in Appendix~\ref{appendix:task_agnostic} for the five languages). Heatmaps for other languages are available in the repository. We identify the most robust models by counting in how many tasks (out of nine) they appear as top performers and selecting those with the highest counts for visualization.

Figure~\ref{fig:task-agnostic} presents a high-level summary of the task-agnostic analysis (see Appendix~\ref{appendix:task_agnostic} for detailed heatmaps), identifying models that maintain the most robust performance across all nine tasks within each language. This addresses a key practical scenario: understanding the robustness of models across diverse downstream applications when task requirements are unknown or varied. The results reveal a clear hierarchy of cross-lingual generalization. \textit{Llama-embed-nemotron-8b} emerges as the only universally robust model, achieving task-agnostic status across all five languages, making it a strong default for multilingual deployments. \textit{Qwen3-Embedding-4B} serves as a competitive alternative for English, French, and Spanish. In contrast, \textit{Qwen3-Embedding-8B} and \textit{Octen-Embedding-8B} achieve the highest task-agnostic robustness only for Hindi. However, \textit{Qwen3-Embedding-8B} still appears among top performers in several tasks (4-5) across other languages, though with slightly lower cross-task consistency than the models highlighted here (6-7 tasks). Therefore, the figure does not indicate that \textit{Qwen-3-Embeding-8B} is not robust, but rather that its cross-task consistency is slightly lower than other models. These differences likely reflect training design. \textit{Llama-embed-nemotron-8b}, trained with contrastive/retrieval objectives on similarity and query–document data, produces more stable embeddings across tasks. By contrast, \textit{Qwen3-Embedding-8B}, trained with a causal language modeling objective and embedding indirectly derived from a generative model, yields embeddings that are less uniformly aligned across tasks. From a language perspective, German is the most constrained (one viable task-agnostic model), while Hindi shows the greatest diversity (three options).

\begin{figure}[!ht]
    \centering
    \includegraphics[width=0.6\linewidth]{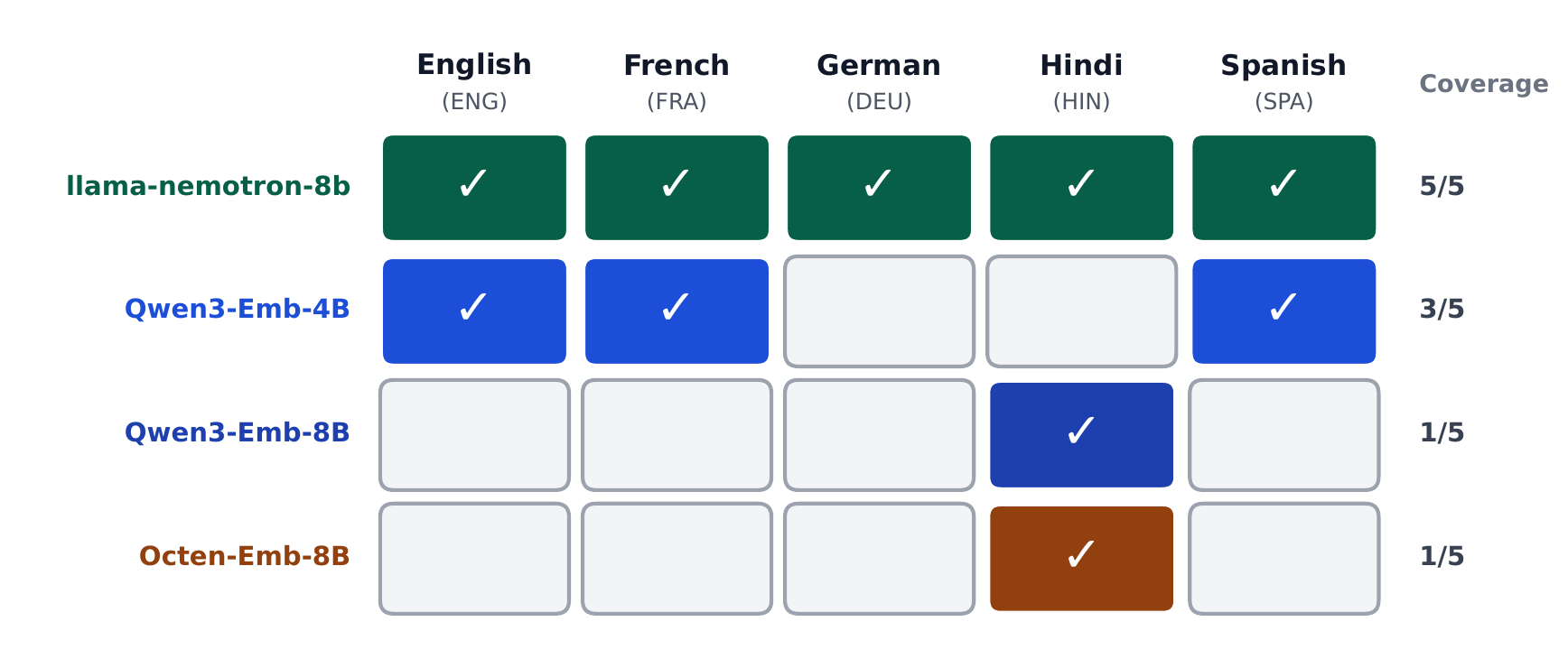}
    \caption{\small Task-agnostic recommendations across five languages (rows: models; columns: languages). Checkmarks denote robustness across all nine tasks; the last column shows language coverage. \textit{llama-embed-nemotron-8b} is robust in all languages (5/5), \textit{Qwen3-Embedding-4B} in three (ENG, FRA, SPA).}
    \label{fig:task-agnostic}
\end{figure}

To assess sensitivity to the choice of $\eta$, we report analyses for top-1 and top-5 settings (Appendix~\ref{app:sensitivty_top}). Results for top-5 and top-10 show consistent trends, with top-10 offering a more balanced view by considering multiple strong models rather than only few top ones. In contrast, the top-1 setting indicates task-agnostic behavior in only two of nine tasks, suggesting task-specific behavior rather than general robustness across tasks. For practical guidance on parameter $\eta$, see Appendix~\ref{appendix:hyperparameters}.

\textbf{Limitations.} 
Our work operates in the performance space and should be viewed as a benchmarking meta-study rather than a model selection approach, which is a distinct problem typically addressed through meta-learning relating dataset characteristics to model performance. We analyze how benchmark conclusions vary with evaluation design choices (dataset composition and score aggregation), providing sensitivity analysis that complements standard summaries (e.g., mean performance) by assessing their stability. Our approach inherits limitations of leaderboard-based meta-evaluations.
First, results reflect a single snapshot (Dec 26, 2025) and may change as benchmarks evolve, though our open methodology allows rerunning the analysis. Second, we rely on publicly reported scores without modeling per-dataset uncertainty, and some user-submitted metadata may contain inaccuracies; we rely on open science practices to improve quality over time. Third, our robustness notation focuses on ranking sensitivity rather than efficiency factors (e.g., latency, memory), highlighting the need to study robustness–efficiency trade-offs. Fourth, we analyze five languages in depth and release others as artifacts, which may require expert interpretation. Fifth, zero-shot filtering may miss leakage through prompt tuning or synthetic data generation~\cite{chung2025maintaining}; our method is agnostic to this choice (no filtering may change rankings but not the methodology). Sixth, rankings are ordinal and may obscure performance gaps; we provide underlying continuous MCDM scores to enable comparison.

\section{Conclusion}
\label{sec:conclusion}
We analyze the robustness of multilingual text embedding rankings within and across tasks using MTEB Multilingual v2. Combining correlation-aware dataset subsampling with multiple MCDM ranking schemes, we identify when rankings are stable and when they depend on dataset choice or the choice of evaluation scheme. Across five languages and nine tasks, only a small set of large LLM-based embedding models remain consistently top-ranked, while most models perform well only for specific tasks or dataset compositions. At the task-agnostic level, robustness is stricter: only a few models remain competitive across tasks, and \textit{llama-embed-nemotron-8b} is the sole model robust across all five languages. We further release robustness results for approximately 230 languages to support understanding of model robustness across different languages.

\section*{Software and Data}

The source code and data are available upon request from the authors.

\section*{Acknowledgement}
This work is funded by the Slovenian Research and Innovation Agency under program grant P2-0098, and project grants No. J2-70078 and No. GC-0001; and by the European Union under Grant Agreement 101211695 (HE MSCA-PF AutoLLMSelect) and Grant Agreement 101187010 (HE ERA Chair AutoLearn-SI).

\bibliographystyle{unsrt}

\newpage
\appendix
\onecolumn

\section{Dataset Distribution}
\label{appendix:dataset_distribution}
The number of datasets included in our analysis by language and task is reported in Table~\ref{fig:dataset_distribution}. The table lists the languages with the highest total number of datasets, while the complete table is available in the project repository. Of the 1,037 languages represented, 798 are associated with only a single dataset, most of which belong to the bitext mining task. The five languages with the highest total number of datasets are therefore selected for detailed analysis in Section~\ref{sec:results} and Appendix~\ref{appendix:results}.

\begin{table}[t]
  \caption{Distribution of datasets by task for ten languages with the largest total number of datasets.}
  \label{fig:dataset_distribution}
  \begin{center}
    \begin{footnotesize}
      \begin{sc}
        \begin{tabular}{lrrrrrrrrrr}
          \toprule
          Lang. & Total & Cls. & Clst. & Retrv. & Rernk. & STS & Pair Cls. & ML Cls. & Instr. Rernk. & Bitext Min.  \\
          \midrule
          ENG & 67 & 10 & 9 & 16 & 2 & 10 & 6 & 1 & 3 & 10 \\
          
          FRA & 25 & 4 & 4 & 3 & 1 & 2 & 4 & 1 & 0 & 6 \\
          
          DEU & 23 & 4 & 1 & 4 & 1 & 3 & 4 & 1 & 0 & 5 \\
          
          HIN & 19 & 4 & 1 & 5 & 1 & 2 & 1 & 0 & 0 & 6 \\
          
          SPA & 17 & 3 & 1 & 3 & 0 & 3 & 2 & 1 & 0 & 4 \\
          
          ITA & 16 & 4 & 1 & 2 & 1 & 2 & 1 & 1 & 0 & 4 \\
          
          RUS & 16 & 2 & 1 & 2 & 1 & 1 & 3 & 1 & 0 & 5 \\
          
          DAN & 15 & 3 & 2 & 3 & 1 & 0 & 0 & 1 & 0 & 5 \\

          SWE & 15 & 4 & 2 & 2 & 1 & 0 & 1 & 1 & 0 & 4 \\

          BEN & 14 & 2 & 1 & 3 & 1 & 1 & 0 & 0 & 0 & 6 \\
          \bottomrule
        \end{tabular}
      \end{sc}
    \end{footnotesize}
  \end{center}
\end{table}

\section{How to Read the Language-Specific, Task-Specific Results}
\label{appendix:reading_results}

In our language-specific, task-specific analysis, two outputs are produced per model for each language--task pair:
(i)~a distribution of ranks across the $K \times S_{q,\ell}$ combinations of ranking schemes and dataset compositions, and (ii)~two robustness
indicators, DS (dataset composition robustness) and RS (ranking scheme
robustness), reported in the summary tables in Appendix~\ref{appendix:results} and Figure~\ref{fig:task_summary} in the main text. A practitioner's decision rule is intentionally simple:

\begin{itemize}
    \item \textbf{DS\,=\,\checkmark\ and RS\,=\,\checkmark (full robustness, green cells in Figure~\ref{fig:task_summary}):} the model's ranking is stable under both changes in dataset composition and ranking scheme (strongest recommendation for that language-task pair).
    \item \textbf{RS\,=\,\checkmark\ only (yellow cells):} the ranking is consistent across ranking schemes but has not been tested across dataset compositions, because either only a single dataset or a single uncorrelated cluster is available (conditional recommendation, pending additional benchmark data for that language).
    \item \textbf{Gray cells}: insufficient benchmark data to evaluate the task for that language (no recommendation made).
\end{itemize}
    
This decision rule requires no knowledge of MCDM. The underlying MCDM ranking schemes and weighting strategies are used purely to stress-test whether a model's top position is stable; they are not themselves scores to be interpreted. Researchers who wish to report a single summary metric can continue to use standard mean performance across datasets, while consulting DS and RS indicators to assess the stability of that summary.

\section{Language-Specific, Task-Specific Multi-Ranking (Sensitivity)}
\label{appendix:fra_clustering_stability_sensitivity}

\begin{figure*}[ht]
  \begin{center}
    \centerline{\includegraphics[width=\textwidth]{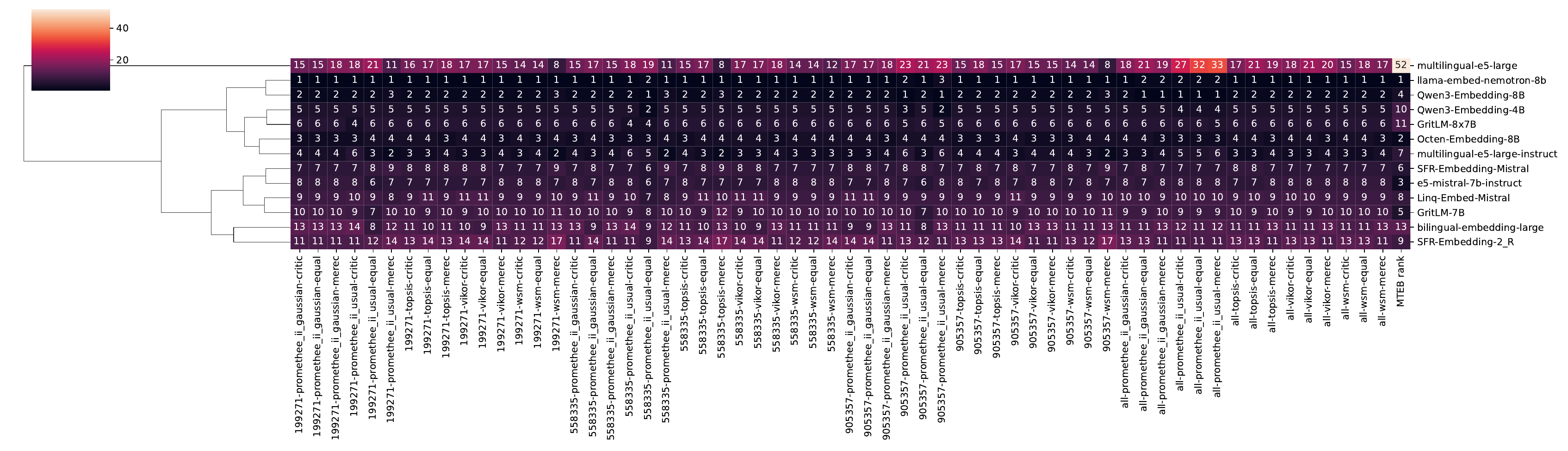}}
    \caption{
      Drendrogram for French language on the clustering task.
    }
    \label{fig:fra_clustering}
  \end{center}
\end{figure*}

The language-specific, task-specific multi-ranking results are visualized using dendrograms with parameter $\eta = 10$. Each dendrogram is constructed from the union of all models that appear in the top 10 ranked models under at least one of the ranking schemes considered in this study. For a language $\ell$, each row of the dendrogram corresponds to a model in the union $\mathcal{U}^{(\ell)}_{10}$, while each column corresponds to a specific ranking scheme and dataset composition. The dendrogram displays the ranks assigned to each model under the corresponding ranking scheme and provides a convenient clustering of models with similar rank distributions across ranking schemes. Dataset compositions are distinguished by the random seed used for dataset sampling. The random seed is given in the ranking scheme names, together with the aggregation and weighting method names. The final column corresponds to the ranking derived from MTEB’s task-language-specific aggregated score, available under the MTEB leaderboard's tab \textit{Performance per language} at the time of download. Because this score is available only for a subset of languages, it is included in our visualizations only for those task–language combinations for which it is available. Figure~\ref{fig:fra_clustering} illustrates the results for the clustering task in French.

\begin{figure}[ht]
  \begin{center}
    \centerline{\includegraphics[width=0.5\textwidth]{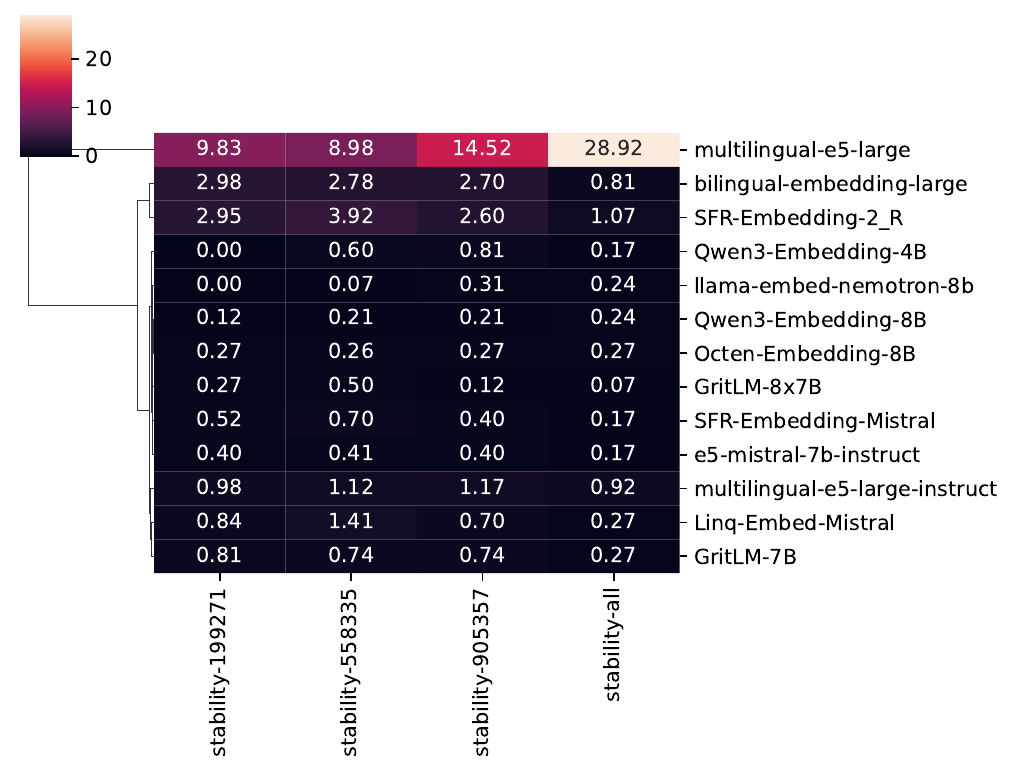}}
    \caption{
      Model sensitivity across ranking schemes for French language on the clustering task.
    }
    \label{fig:fra_clustering_st}
  \end{center}
\end{figure}

\begin{figure*}[ht]
  \begin{center}
    \centerline{\includegraphics[width=\textwidth]{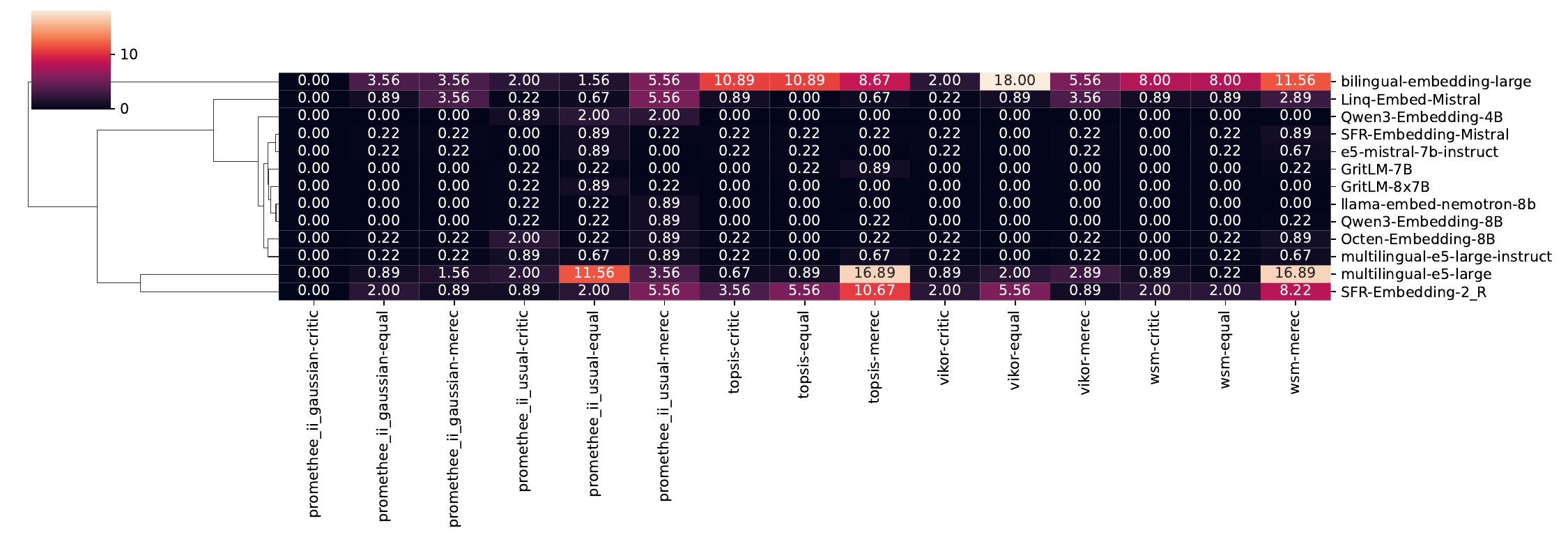}}
    \caption{
      Model sensitivity to dataset compositions by ranking scheme for French language on the clustering task.
    }
    \label{fig:fra_clustering_sn}
  \end{center}
\end{figure*}

Figure~\ref{fig:fra_clustering_st} presents model sensitivity across ranking schemes for the French language on the clustering task. Sensitivity is computed separately for each dataset composition, starting with the three uncorrelated compositions and ending with the composition that includes all datasets. Figure~\ref{fig:fra_clustering_sn} illustrates model sensitivity to dataset composition by ranking scheme for the same language and task. All three figures are referenced in the discussion of the French clustering results in Section~\ref{sec:results}.

\begin{figure*}[!ht]
  \begin{center}
    \centerline{\includegraphics[width=\textwidth]{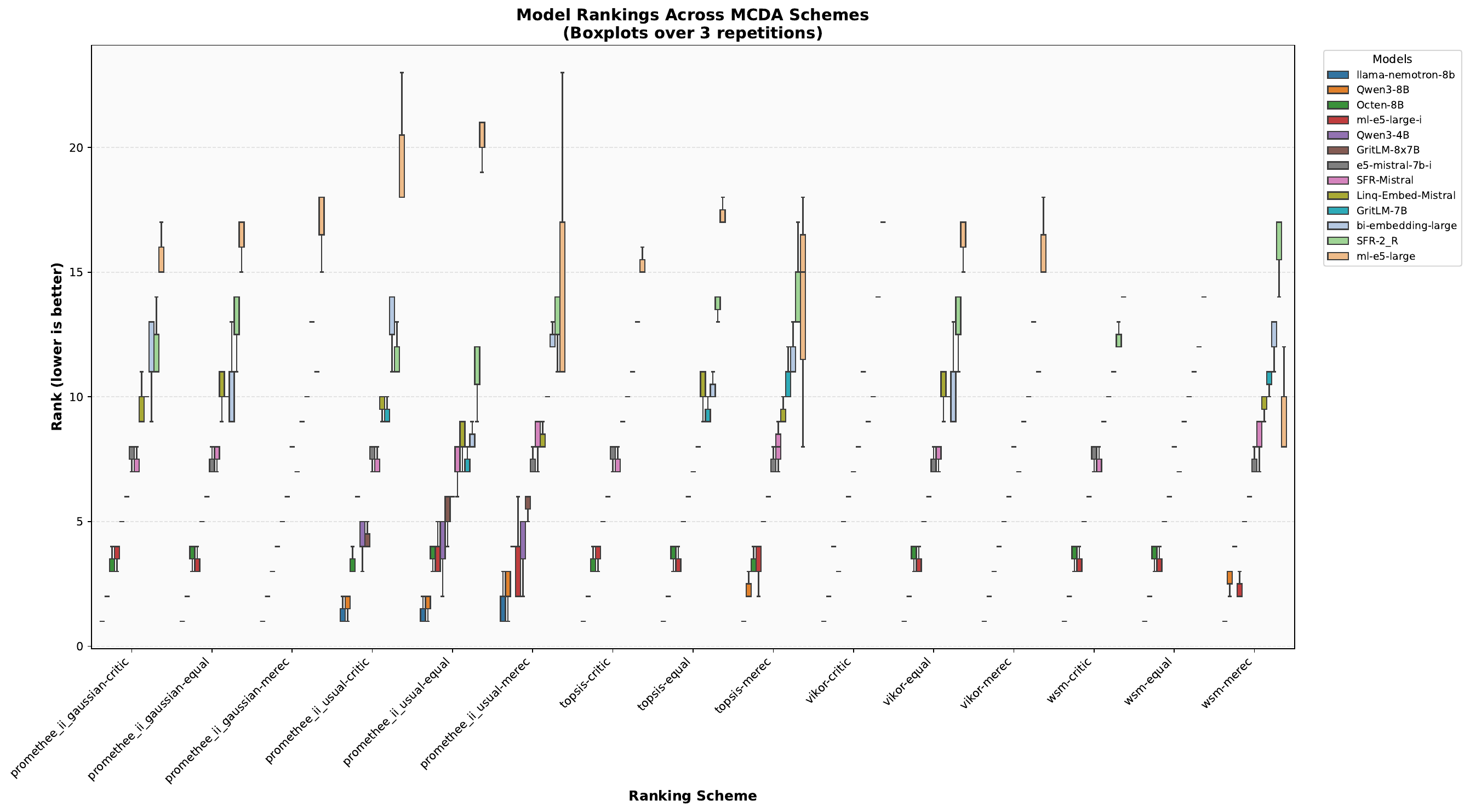}}
    \caption{
      Ranking sensitivity of the embedding models from Figure~\ref{fig:fra_clustering} across the ranking schemes for clustering in French.  Boxplots show rank distributions across different MCDM methods, each with three weighting schemes and three dataset compositions.
    }
    \label{fig:fra_clustering_shaded}
  \end{center}
\end{figure*}

\begin{figure*}[!ht]
  \begin{center}
    \centerline{\includegraphics[width=\textwidth]{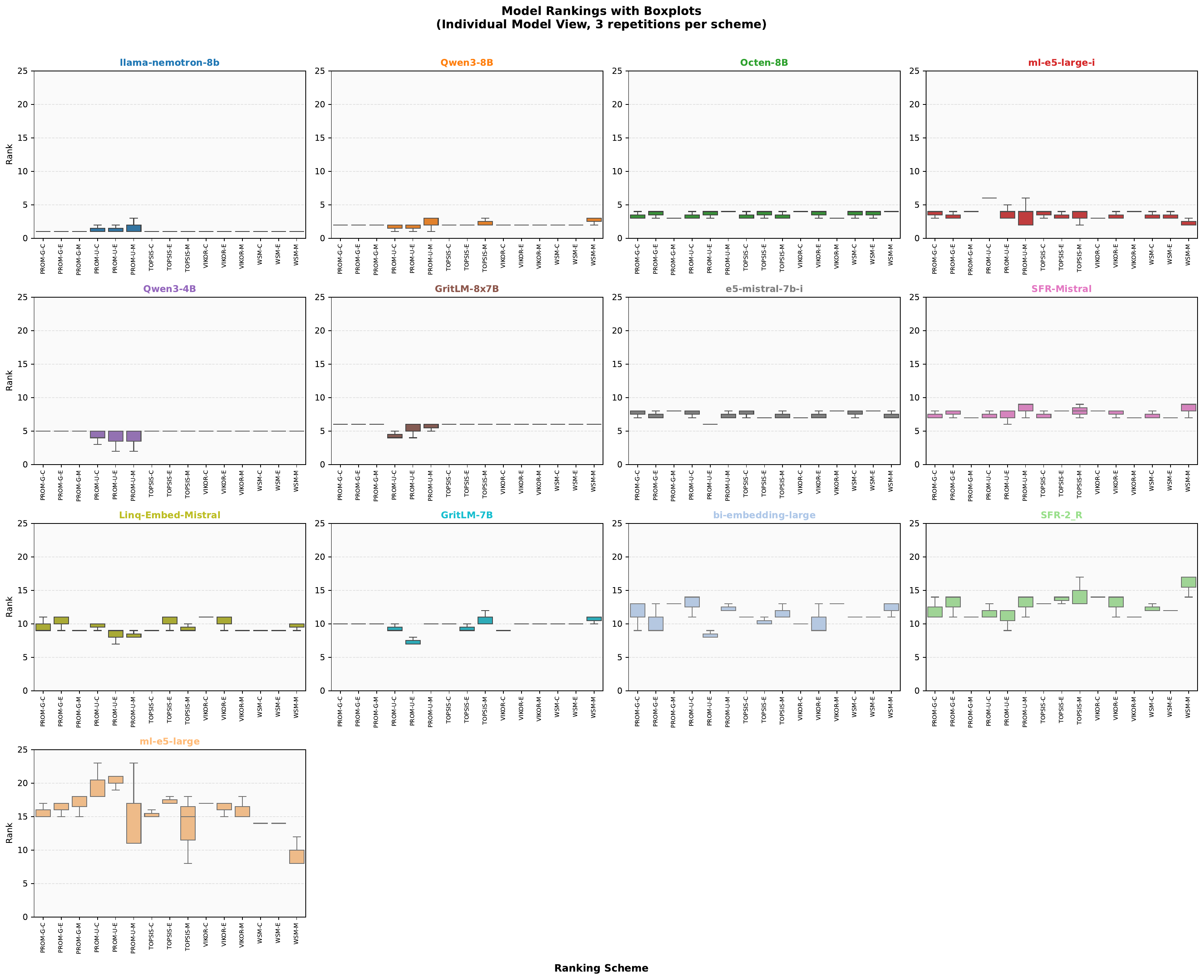}}
    \caption{
      Ranking sensitivity of each individual embedding model from Figure~\ref{fig:fra_clustering} across the ranking schemes for clustering in French. Boxplots show rank distributions across different MCDM methods, each with three weighting schemes and three dataset compositions.
    }
    \label{fig:fra_clustering_faceted}
  \end{center}
\end{figure*}

Figure~\ref{fig:fra_clustering_shaded} illustrates the ranking sensitivity of the embedding models from Figure~\ref{fig:fra_clustering} across the ranking schemes for clustering in French. Lines show models average rank, while the shaded regions indicate 95\% confidence intervals computed over the three random seeds (dataset compositions). Figure~\ref{fig:fra_clustering_faceted} extends this visualization to each individual model.

\begin{table}[]
  \caption{Most robust models on the clustering task by language. Robustness is assessed with respect to dataset compositions (DS) and ranking schemes (RS).}
  \label{fig:summary_clustering}
  \begin{center}
    \begin{small}
      \begin{sc}
        \begin{tabularx}{\columnwidth}{lccX}
          \toprule
          Lang. & DS & RS & Most Robust Models (Clustering) \\
          \midrule
          ENG & \checkmark & \checkmark & Qwen3-Embedding-8B, Qwen3-Embedding-4B, llama-embed-nemotron-8b \\
          FRA & \checkmark & \checkmark & llama-embed-nemotron-8b, Qwen3-Embedding-8B \\
          DEU &  & \checkmark & llama-embed-nemotron-8b \\
          HIN &  & \checkmark & llama-embed-nemotron-8b \\
          SPA &  & \checkmark & llama-embed-nemotron-8b \\
          \bottomrule
        \end{tabularx}
      \end{sc}
    \end{small}
  \end{center}
\end{table}

\section{Language-Specific, Task-Specific Multi-Ranking (Remaining Tasks)}
\label{appendix:results}

This appendix extends the analysis presented in Section~\ref{sec:results} to the remaining tasks for the five selected languages (French, English, German, Hindi, and Spanish). Analyses for all other languages can be conducted analogously using the visualizations available in the project repository.

\subsection{Classification}

\begin{table}[t]
  \caption{Most robust models on the classification task by language. Robustness is assessed with respect to dataset compositions (DS) and ranking schemes (RS).}
  \label{fig:summary_classifiaction}
  \begin{center}
    \begin{small}
      \begin{sc}
        \begin{tabularx}{\columnwidth}{lccX}
          \toprule
          Lang. & DS & RS & Most Robust Models (Classification) \\
          \midrule
          ENG &  & \checkmark & Qwen3-Embedding-8B \\
          FRA &  & \checkmark & Qwen3-Embedding-8B \\
          DEU &  & \checkmark & Qwen3-Embedding-8B \\
          HIN & \checkmark & \checkmark & Qwen3-Embedding-8B \\
          SPA &  & \checkmark & Qwen3-Embedding-8B \\
          \bottomrule
        \end{tabularx}
      \end{sc}
    \end{small}
  \end{center}
\end{table}

\begin{figure*}[ht]
  \begin{center}
    \centerline{\includegraphics[width=\textwidth]{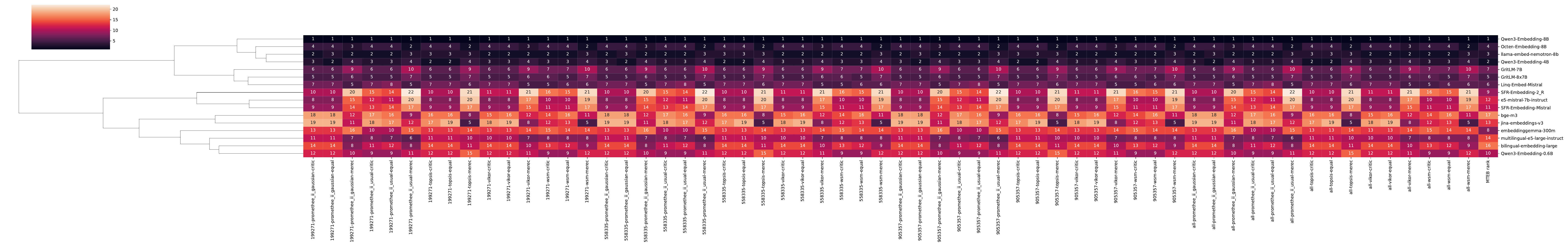}}
    \caption{
      Drendrogram for French language on the classification task.
    }
    \label{fig:fra_classification}
  \end{center}
\end{figure*}

In the classification task, four datasets involve the \textbf{French} language. All dataset pairs exhibit mutual correlations below the threshold of 0.9; consequently, each dataset forms an individual cluster, and all dataset compositions shown in Figure~\ref{fig:fra_classification} include all four datasets. The figure reveals five clusters with similar ranking patterns, of which the top cluster achieves the highest rankings and includes the models \textit{Qwen3-Embedding-8B}, \textit{Qwen3-Embedding-4B}, \textit{Octen-Embedding-8B}, and \textit{llama-embed-nemotron-8b}. Model \textit{Qwen3-Embedding-8B} is consistently ranked first across all ranking schemes, followed by its smaller-scale variant \textit{Qwen3-Embedding-4B} and \textit{llama-embed-nemotron-8b}. These two models alternate between second and third place, exhibiting slightly higher sensitivity across ranking schemes compared to the top-ranked model. Model \textit{Octen-Embedding-8B} is frequently ranked fourth and shows marginally higher ranking sensitivity than the preceding models. As these models also emerge among the top performers in the clustering task analyzed in detail in Section~\ref{sec:results}, we briefly note their large-scale LLM backbones with approximately 8B parameters: Llama-3.1-8B for \textit{llama-embed-nemotron-8b} and Qwen3 for the remaining three models~\cite{zhang2025qwen3,babakhin2025llama,octen-embedding-2025}. The second-best cluster comprises \textit{GritLM-7B}, \textit{GritLM-8x7B}, and \textit{Linq-Embed-Mistral}, which are likewise based on large-scale LLMs, namely Mistral-7B-v0.1 for \textit{GritLM-7B} and \textit{Linq-Embed-Mistral}, and Mixtral 8x7B for \textit{GritLM-8x7B}~\cite{choi2024linq,muennighoff2024generative}.

Although based on different core LLMs, the two GritLM models share the same generative and representational instruction-tuning procedure~\cite{muennighoff2024generative}. Within this cluster, \textit{GritLM-8x7B} exhibits the lowest variability across ranking schemes (i.e., the lowest ST score), whereas \textit{GritLM-7B} shows the highest. The third cluster comprises \textit{SFR-Embedding-2\_R}, \textit{SFR-Embedding-Mistral}, and \textit{e5-mistral-7b-instruct}. While \textit{e5-mistral-7b-instruct} is based on Mistral-7B-v0.1, \textit{SFR-Embedding-Mistral} builds upon both \textit{e5-mistral-7b-instruct} and Mistral-7B-v0.1, with \textit{SFR-Embedding-2\_R} belonging to the same model family. All three models exhibit high ST scores, indicating substantial variability in rankings across schemes. The fourth and fifth clusters span models from \textit{bge-m3} to \textit{Qwen3-Embedding-0.6B}, which achieve mid-range rankings and exhibit noticeable variability across ranking schemes. Overall, \textit{Qwen3-Embedding-8B} emerges as the most robust choice for this language–task combination, consistent with its performance in the clustering task (Section~\ref{sec:results}). In this setting, the sensitivity score across dataset compositions is consistently equal to zero.

The \textbf{English} language is associated with ten datasets, each forming an individual cluster. Consequently, only a single dataset composition exists, which includes all datasets. The best-performing cluster is dominated by Qwen3-based models, with \textit{Qwen3-Embedding-8B} consistently ranked first and \textit{Qwen3-Embedding-4B} consistently ranked second. The models \textit{Octen-Embedding-8B} and \textit{Qwen3-Embedding-0.6B} alternate between fourth and fifth positions. This cluster also includes the Llama-3.1-8B-based \textit{llama-embed-nemotron-8b}, which consistently attains the third rank. In contrast, the Mistral-7B-based models (\textit{Linq-Embed-Mistral} and \textit{SFR-Embedding-2\_R}) and the Gemma 3-based model \textit{embeddinggemma-300m} exhibit slightly lower and more variable rankings across ranking schemes. Overall, the dendrogram is dominated by large-scale LLM-based embedding models, with the \textit{Qwen3-Embedding} family emerging as the most robust choice for this language–task combination.

The \textbf{German} language is associated with four datasets, each forming an individual cluster, resulting in a single dataset composition. The best-performing cluster is again dominated by Qwen3-based models, with \textit{Qwen3-Embedding-8B} consistently ranked first across all ranking schemes. Its smaller-scale variant \textit{Qwen3-Embedding-4B} and the fine-tuned model \textit{Octen-Embedding-8B} alternate between second and third positions, followed by the Llama-3.1-8B-based \textit{llama-embed-nemotron-8b}. The second-best cluster comprises \textit{Linq-Embed-Mistral}, \textit{GritLM-7B}, and \textit{GritLM-8x7B}, representing large-scale models based on the Mistral-7B and Mixtral 8x7B backbones. As in the English case, the dendrogram is dominated by large-scale LLM-based embedding models, with the \textit{Qwen3-Embedding} family emerging as the most robust choice for this language–task combination.

The \textbf{Hindi} language is associated with four datasets, which are grouped into three clusters ([SentimentAnalysisHindi, MassiveIntentClassification], [IndicLangClassification], and [MultiHateClassification]). The best-performing cluster comprises the large-scale LLM-based models \textit{Qwen3-Embedding-8B} and \textit{llama-embed-nemotron-8b}, together with the smaller-scale \textit{Qwen3-Embedding-4B}. All three models exhibit stable rankings across ranking schemes and low sensitivity to dataset composition. The second-best cluster is also dominated by LLM-based models and further decomposes into two subclusters. The first includes the smaller-scale \textit{Qwen3-Embedding-0.6B} and the XLM-RoBERTa-based \textit{multilingual-e5-large-instruct}. The second contains the Mistral-7B-based models \textit{Linq-Embed-Mistral} and \textit{GritLM-7B}, along with the Mixtral-8x7B-based \textit{GritLM-8x7B}. Models in this cluster exhibit higher variability across ranking schemes than those in the best-performing cluster, with low to moderate sensitivity to dataset composition. The remaining large cluster is dominated by Mistral-7B-based models, including \textit{SFR-Embedding-Mistral}, \textit{SFR-Embedding-2\_R}, \textit{e5-mistral-7b-instruct}, and \textit{LLM2Vec-Mistral-7B-Instruct-v2-mntp-supervised}, which achieve mid-range rankings across ranking schemes. In contrast, \textit{Octen-Embedding-8B} attains low rankings and exhibits high sensitivity to dataset composition. Overall, \textit{Qwen3-Embedding-8B} and \textit{llama-embed-nemotron-8b} emerge as the most robust choices for this language–task combination.

The \textbf{Spanish} language is associated with three datasets, each forming an individual cluster due to mutual correlations below the specified threshold ([CataloniaTweetClassification], [MassiveIntentClassification], and [MultiHateClassification]). Consequently, only a single dataset composition is available, which includes all three datasets. The best-performing cluster comprises large-scale LLM-based models, with \textit{Qwen3-Embedding-8B} consistently ranked first across all ranking schemes, followed by \textit{llama-embed-nemotron-8b}, \textit{Octen-Embedding-8B}, and \textit{Qwen3-Embedding-4B}. The latter three exhibit slightly higher variability across ranking schemes. The second-best cluster includes the Mistral-7B-based models \textit{Linq-Embed-Mistral} and \textit{GritLM-7B}, together with the Mixtral-8x7B-based \textit{GritLM-8x7B}. These models achieve slightly lower rankings than those in the best-performing cluster but exhibit low variability across ranking schemes. The remaining three clusters, spanning models from \textit{SFR-Embedding-Mistral} to \textit{bge-m3}, attain mid-range rankings and generally show higher variability across ranking schemes than models in the first two clusters. Overall, \textit{Qwen3-Embedding-8B} emerges as the most stable choice for this language–task combination.

\subsection{Retrieval}
\begin{table}[t]
  \caption{Most robust models on the retrieval task by language. Robustness is assessed with respect to dataset compositions (DS) and ranking schemes (RS).}
  \label{fig:summary_retrieval}
  \begin{center}
    \begin{small}
      \begin{sc}
        \begin{tabularx}{\columnwidth}{lccX}
          \toprule
          Lang. & DS & RS & Most Robust Models (Retrieval) \\
          \midrule
          ENG & \checkmark & \checkmark & bilingual-embedding-large \\
          FRA &  & \checkmark & jina-embeddings-v3, bilingual-embedding-large \\
          DEU & \checkmark & \checkmark & jina-embeddings-v3, bilingual-embedding-large, bilingual-embedding-base \\
          HIN & \checkmark & \checkmark & jina-embeddings-v3, bilingual-embedding-large, bilingual-embedding-base \\
          SPA & \checkmark & \checkmark & bilingual-embedding-large, jina-embeddings-v3, bilingual-embedding-base \\
          \bottomrule
        \end{tabularx}
      \end{sc}
    \end{small}
  \end{center}
\end{table}

\begin{figure*}[ht]
  \begin{center}
    \centerline{\includegraphics[width=\textwidth]{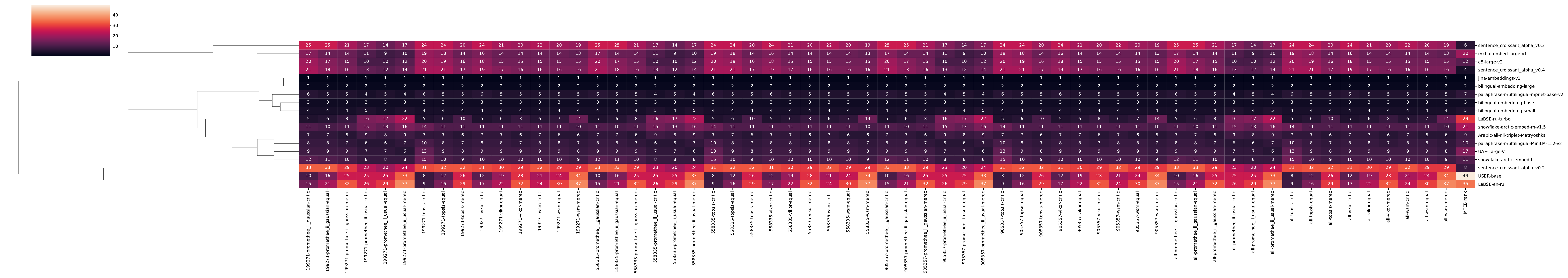}}
    \caption{
      Drendrogram for French language on the retrieval task.
    }
    \label{fig:fra_retrieval}
  \end{center}
\end{figure*}

In the retrieval task, three datasets are available for the \textbf{French} language, all of which exhibit pairwise correlations below the threshold of 0.9. Consequently, each dataset forms an individual cluster, and, as in the classification task, all dataset compositions include all three datasets. Figure~\ref{fig:fra_retrieval} presents the dendrogram of top-10 model rankings across the different ranking schemes, in which five major clusters are observed. The best-performing cluster is composed of two subclusters. The first includes \textit{jina-embeddings-v3}, which is consistently ranked first, and \textit{bilingual-embedding-large}, which is consistently ranked second. Both models are based on the XLM-RoBERTa architecture, the multilingual extension of RoBERTa optimized for 100 languages~\cite{conneau2020unsupervised}. While \textit{jina-embeddings-v3} is optimized for 30 languages, including French~\cite{sturua2024jina}, and supports substantially longer input sequences Model \textit{bilingual-embedding-large} is a bilingual (French–English) model optimized for a range of bilingual tasks. The second-best cluster comprises smaller-scale variants of the second-ranked model, namely \textit{bilingual-embedding-base} and \textit{bilingual-embedding-small}, together with \textit{paraphrase-multilingual-mpnet-base-v2}, which belongs to the Sentence Transformer family, is based on the MPNet architecture, and is optimized for semantic search and clustering~\cite{reimers2019sentence}. All five models exhibit low variability in rankings across ranking schemes. The third cluster includes \textit{Arabic-all-nli-triplet-Matryoshka}, \textit{paraphrase-multilingual-MiniLM-L12-v2}, \textit{UAE-Large-V1}, \textit{snowflake-arctic-embed-l}, \textit{LaBSE-ru-turbo}, and \textit{snowflake-arctic-embed-m-v1.5}. These models achieve mid-range rankings with slightly higher variability across ranking schemes than those in the first two clusters. The fourth cluster, spanning models from \textit{sentence\_croissant\_alpha\_v0.3} to \textit{sentence\_croissant\_alpha\_v0.4}, also exhibits mid-range rankings across all ranking schemes. The fifth cluster, which spans models from \textit{sentence\_croissant\_alpha\_v0.2} to \textit{LaBSE-en-ru}, attains the lowest rankings among all models shown and displays noticeably higher variability across ranking schemes compared to the other clusters. Overall, for the French retrieval task, the best-performing embedding models are predominantly smaller-scale models based on Transformer encoder architectures, optimized for semantic similarity, retrieval, and clustering tasks.

In the retrieval task, the \textbf{English} language is associated with 16 datasets, which are grouped into 13 clusters: two clusters contain two datasets each, while the remaining clusters consist of a single dataset. The resulting dendrogram reveals three large clusters composed of smaller subclusters. The best-performing subcluster comprises the bilingual (English–French) models \textit{bilingual-embedding-large} and \textit{bilingual-embedding-base}, as well as the multilingual model \textit{jina-embeddings-v3}, mirroring the results observed for the French language. While both \textit{bilingual-embedding} models exhibit stable rankings across ranking schemes, the rankings of \textit{jina-embeddings-v3} show higher variability. In addition, \textit{bilingual-embedding-large} demonstrates low sensitivity to dataset composition. The second subcluster within the same cluster includes \textit{bilingual-embedding-small} and \textit{e5-large-v2} from the e5 model family~\cite{wang2022text}, which exhibit slightly higher variability across ranking schemes than the top-performing models. The second-best cluster comprises the angle-optimized embedding model \textit{UAE-Large-V1}~\cite{li2023angle}, \textit{mxbai-embed-large-v1}~\cite{emb2024mxbai}, and the large and small variants of \textit{snowflake-arctic-embed}, which are optimized for retrieval tasks~\cite{merrick2024arctic}. Overall, \textit{bilingual-embedding-large} emerges as the most robust choice for English retrieval tasks.

The results for the \textbf{German} language on the retrieval task are highly similar. Four datasets are associated with German and are grouped into three clusters. The best-performing cluster in the dendrogram includes \textit{jina-embeddings-v3}, \textit{bilingual-embedding-large}, and \textit{bilingual-embedding-base}, all of which exhibit low sensitivity across ranking schemes and relatively low sensitivity to dataset composition. The second-best cluster comprises \textit{bilingual-embedding-small}, \textit{Arabic-all-nli-triplet-Matryoshka}, and \textit{paraphrase-multilingual-mpnet-base-v2}. These models achieve slightly lower rankings than those in the best-performing cluster but show very low ranking variability across schemes and low sensitivity to dataset composition. Overall, \textit{jina-embeddings-v3} and \textit{bilingual-embedding-large} demonstrate the most stable top rankings for German retrieval tasks. A closely analogous pattern is observed for the \textbf{Hindi} language, where four datasets are likewise grouped into three clusters.

The \textbf{Spanish} language is associated with three datasets, which are grouped into two clusters. In the resulting dendrogram, the best-performing cluster includes the same three models observed for the other languages, i.e., \textit{bilingual-embedding-large}, \textit{jina-embeddings-v3}, and \textit{bilingual-embedding-base}, all of which exhibit low variability across ranking schemes and relatively low sensitivity to dataset composition. The second-best cluster comprises \textit{bilingual-embedding-small}, \textit{Arabic-all-nli-triplet-Matryoshka}, and \textit{paraphrase-multilingual-mpnet-base-v2}. These models show no ranking variability when the dataset composition includes all datasets and slightly higher variability for the uncorrelated subsamples. With the exception of \textit{paraphrase-multilingual-mpnet-base-v2}, the remaining models in this cluster exhibit low sensitivity to the dataset composition. Overall, \textit{bilingual-embedding-large} and \textit{jina-embeddings-v3} again emerge as the top-performing models for the Spanish retrieval task.

\subsection{Reranking}

\begin{table}[t]
  \caption{Most robust models on the reranking task by language. Robustness is assessed with respect to dataset compositions (DS) and ranking schemes (RS).}
  \label{fig:summary_reranking}
  \begin{center}
    \begin{small}
      \begin{sc}
        \begin{tabularx}{\columnwidth}{lccX}
          \toprule
          Lang. & DS & RS & Most Robust Models (Reranking) \\
          \midrule
          ENG &  & \checkmark & Octen-Embedding-8B \\
          FRA &  & \checkmark & Octen-Embedding-8B \\
          DEU &  & \checkmark & llama-embed-nemotron-8b \\
          HIN &  & \checkmark & llama-embed-nemotron-8b \\
          SPA &  &  & Not applicable \\
          \bottomrule
        \end{tabularx}
      \end{sc}
    \end{small}
  \end{center}
\end{table}

\begin{figure}[ht]
  \begin{center}
    \centerline{\includegraphics[width=0.4\columnwidth]{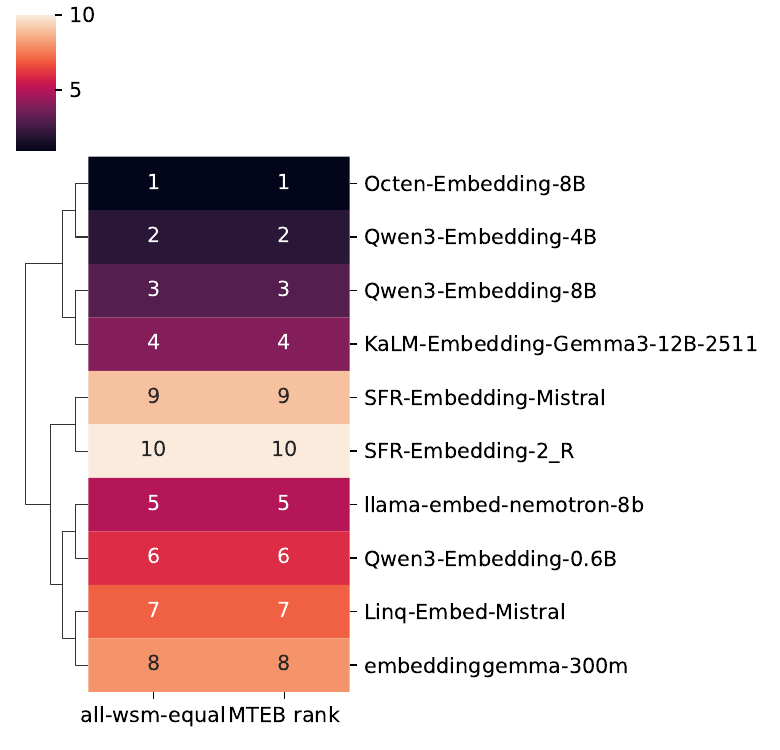}}
    \caption{
      Drendrogram for French language on the reranking task.
    }
    \label{fig:fra_reranking}
  \end{center}
\end{figure}

Figure~\ref{fig:fra_reranking} presents the dendrogram for the \textbf{French} language on the reranking task. As only a single dataset is available for French, model rankings are determined directly by performance on this dataset. The top 10 ranked models are predominantly large-scale LLM-based models, led by \textit{Octen-Embedding-8B}, followed by \textit{Qwen3-Embedding-4B} and \textit{Qwen3-Embedding-8B} in second and third place, respectively. The models \textit{KaLM-Embedding-Gemma3-12B-2511} and \textit{llama-embed-nemotron-8b} rank fourth and fifth. The remaining models include the smaller-scale \textit{Qwen3-Embedding-0.6B}, the Mistral-7B-based models \textit{Linq-Embed-Mistral}, \textit{SFR-Embedding-Mistral}, and \textit{SFR-Embedding-2\_R}, as well as the smaller-scale Gemma3-based model \textit{embeddinggemma-300m}. Overall, \textit{Octen-Embedding-8B} emerges as the most robust model for reranking in French.

For the \textbf{English} language, two datasets are available, each forming an individual cluster; consequently, resulting in a single dataset composition which includes all datasets. The best-performing cluster is led by the top-ranked \textit{Octen-Embedding-8B}, which exhibits stable rankings across all ranking schemes, followed by the second-ranked \textit{KaLM-Embedding-Gemma3-12B-2511}. The models \textit{NV-Embed-v1} and \textit{llama-embed-nemotron-8b} alternate between third and fourth place. All models in this cluster are large-scale and LLM-based, and with the exception of \textit{llama-embed-nemotron-8b}, they exhibit relatively stable rankings across schemes. These models are followed by the large-scale \textit{NV-Embed-v2}, based on the Mistral-7B-v0.1 LLM~\citep{lee2024nv}, and \textit{Qwen3-Embedding-8B}, both of which also show stable rankings across ranking schemes. The next performance cluster comprises additional LLM-based models, including \textit{Linq-Embed-Mistral}, the smaller-scale Qwen3-based models \textit{Qwen3-Embedding-4B} and \textit{Qwen3-Embedding-0.6B}, as well as \textit{GritLM-7B} and \textit{GritLM-8x7B}. While the first two models in this cluster exhibit stable rankings, the remaining models display higher variability across ranking schemes. Overall, \textit{Octen-Embedding-8B} emerges as the most robust choice for reranking in English.

For \textbf{German}, only a single dataset is available for the reranking task, and models are ranked directly based on their performance on this dataset. The top-ranked model is the LLM-based \textit{llama-embed-nemotron-8b}, followed by \textit{GritLM-7B} and the XLM-RoBERTa-based \textit{multilingual-e5-large-instruct}. The subsequent positions are occupied by large-scale LLM-based models, including \textit{Octen-Embedding-8B}, \textit{NV-Embed-v2}, \textit{KaLM-Embedding-Gemma3-12B-2511}, and \textit{Qwen3-Embedding-8B}. Overall, the reranking task in German is dominated by large-scale LLM-based models, with \textit{llama-embed-nemotron-8b} emerging as the top-performing model.

For \textbf{Hindi}, only a single dataset is available for the reranking task, and models are ranked directly based on their performance on this dataset. The top-ranked model is \textit{llama-embed-nemotron-8b}, followed by \textit{GritLM-7B}, \textit{multilingual-e5-large-instruct}, \textit{Octen-Embedding-8B}, \textit{NV-Embed-v2}, \textit{KaLM-Embedding-Gemma3-12B-2511}, and \textit{Qwen3-Embedding-8B}. These are followed by the Mistral-7B-based models \textit{e5-mistral-7b-instruct}, \textit{SFR-Embedding-Mistral}, and \textit{Linq-Embed-Mistral}. Overall, reranking performance in Hindi is dominated by large-scale LLM-based models, with \textit{llama-embed-nemotron-8b} emerging as the top-performing model.

There are no datasets in \textbf{Spanish} on the retrieval task.

\subsection{Pair classification}

\begin{table}[t]
  \caption{Most robust models on the pair classification task by language. Robustness is assessed with respect to dataset compositions (DS) and ranking schemes (RS).}
  \label{fig:summary_pair_classification}
  \begin{center}
    \begin{small}
      \begin{sc}
        \begin{tabularx}{\columnwidth}{lccX}
          \toprule
          Lang. & DS & RS & Most Robust Models (Pair Classification) \\
          \midrule
          ENG & \checkmark & \checkmark & Qwen3-Embedding-8B, Qwen3-Embedding-4B, Octen-Embedding-8B \\
          FRA &  & \checkmark & Qwen3-Embedding-8B, Qwen3-Embedding-4B, Octen-Embedding-8B \\
          DEU &  & \checkmark & Qwen3-Embedding-8B, Qwen3-Embedding-4B, Octen-Embedding-8B \\
          HIN &  & \checkmark & Qwen3-Embedding-8B, Qwen3-Embedding-4B, Octen-Embedding-8B \\
          SPA &  & \checkmark & Qwen3-Embedding-8B, Qwen3-Embedding-4B, Octen-Embedding-8B \\
          \bottomrule
        \end{tabularx}
      \end{sc}
    \end{small}
  \end{center}
\end{table}

\begin{figure*}[ht]
  \begin{center}
    \centerline{\includegraphics[width=\textwidth]{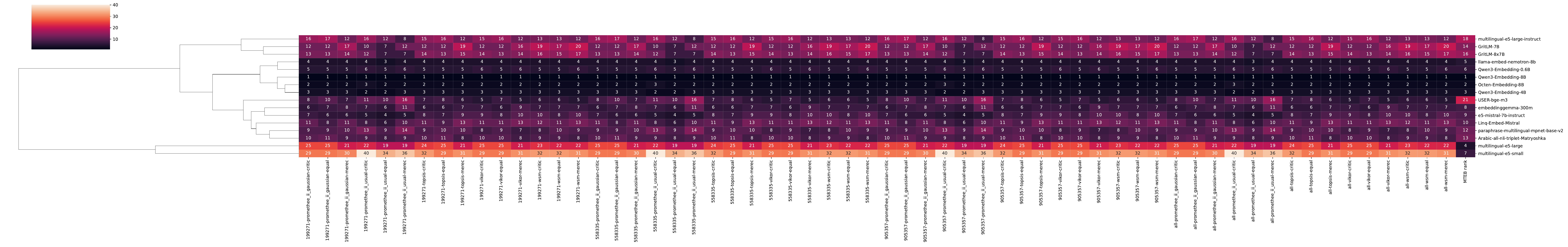}}
    \caption{
      Drendrogram for French language on the pair classification task.
    }
    \label{fig:fra_pair_classification}
  \end{center}
\end{figure*}

In the pair classification task, four datasets involve the \textbf{French} language, each forming an individual cluster due to pairwise correlations below the threshold of 0.9. Figure~\ref{fig:fra_pair_classification} presents the dendrogram for this language–task combination. The best-performing cluster spans models from \textit{llama-embed-nemotron-8b} to \textit{Qwen3-Embedding-4B} and decomposes into two subclusters. The first subcluster includes the \textit{Qwen3-Embedding} models with 8B and 4B parameters, as well as \textit{Octen-Embedding-8B}, which is fine-tuned from \textit{Qwen3-Embedding-8B}. The second subcluster also comprises LLM-based embedding models, namely the large-scale \textit{llama-embed-nemotron-8b}, based on the Llama-3.1-8B LLM, and the smaller-scale \textit{Qwen3-Embedding-0.6B}. All five models exhibit low ranking variability across ranking schemes. The second-best cluster spans models from \textit{USER-bge-m3} to \textit{Arabic-all-nli-triplet-Matryoshka}, achieving lower rankings than the best-performing cluster and exhibiting higher variability. The third cluster includes the XLM-RoBERTa-based \textit{multilingual-e5-large-instruct} and the LLM-based GritLM models (\textit{GritLM-7B} and \textit{GritLM-8x7B}), which show the highest ranking variability among the illustrated models. Finally, the fourth cluster comprises \textit{multilingual-e5-large} and its smaller-scale variant \textit{multilingual-e5-small}. Overall, for this language–task combination, \textit{Qwen3-Embedding-8B} and its fine-tuned variant \textit{Octen-Embedding-8B} emerge as the best-performing and most stable models across ranking schemes.

In this task, six datasets are associated with the \textbf{English} language and are grouped into five clusters. The resulting dendrogram reveals nine clusters in total. The best-performing cluster comprises large-scale LLM-based models, including \textit{Qwen3-Embedding-8B}, \textit{Octen-Embedding-8B}, \textit{Qwen3-Embedding-4B}, \textit{llama-embed-nemotron-8b}, and \textit{e5-mistral-7b-instruct}. However, these models exhibit relatively high sensitivity across ranking schemes, particularly under PROMETHEE II with the usual preference function, which ranks them substantially lower than other methods. Among them, \textit{Qwen3-Embedding-8B} achieves the highest rankings, followed by \textit{Octen-Embedding-8B} and \textit{Qwen3-Embedding-4B}. Two additional clusters achieve moderate rankings. The first includes \textit{Qwen3-Embedding-0.6B} and \textit{GritLM-7B}, while the second comprises \textit{embeddinggemma-300m}, \textit{USER-bge-m3}, \textit{bge-m3}, \textit{Linq-Embed-Mistral}, and \textit{SFR-Embedding-Mistral}, spanning models based on the Mistral-7B, Gemma 3, and bge-m3 architectures. The models \textit{paraphrase-multilingual-mpnet-base-v2} and \textit{Arabic-all-nli-triplet-Matryoshka} also attain mid-range rankings, whereas the remaining models exhibit lower rankings. Overall, \textit{Qwen3-Embedding-8B} emerges as the most robust choice for this language–task combination, followed by \textit{Octen-Embedding-8B} and \textit{Qwen3-Embedding-4B}.

For the \textbf{German} language, four datasets are available, each forming an individual cluster due to pairwise correlations below the specified threshold. Consequently, only a single dataset composition is considered. The dendrogram reveals three large clusters, of which the best-performing cluster further decomposes into two subclusters. In the first subcluster, the top-ranked model is \textit{Qwen3-Embedding-8B}, followed by \textit{Octen-Embedding-8B} and \textit{Qwen3-Embedding-4B}. The second subcluster contains the slightly lower-ranked models \textit{llama-embed-nemotron-8b} and \textit{Qwen3-Embedding-0.6B}. All five models exhibit low ranking sensitivity across ranking schemes. The remaining clusters consist of models with mid-range rankings and higher ranking variability across schemes. Overall, \textit{Qwen3-Embedding-8B} emerges as the most robust choice for pair classification in German, followed by \textit{Octen-Embedding-8B} and \textit{Qwen3-Embedding-4B}.

For pair classification in \textbf{Hindi}, only a single dataset is available; consequently, models are ranked solely based on their performance on this dataset. The top-ranked model is \textit{Qwen3-Embedding-8B}, followed by \textit{USER-bge-m3}, \textit{Qwen3-Embedding-4B}, and \textit{Octen-Embedding-8B}. Overall, Qwen3-based models demonstrate the strongest performance for pair classification in the Hindi language.

For pair classification in \textbf{Spanish}, two datasets are available, each forming an individual cluster; consequently, only a single dataset composition is considered. The dendrogram indicates that the best-performing cluster is dominated by large-scale LLM-based models. The top-ranked model is \textit{Qwen3-Embedding-8B}, followed by \textit{Octen-Embedding-8B} and \textit{Qwen3-Embedding-4B}, all of which are Qwen3-based models exhibiting stable rankings across ranking schemes. These models are followed by \textit{llama-embed-nemotron-8b}, \textit{Qwen3-Embedding-0.6B}, and \textit{embeddinggemma-300m}, which also demonstrate stable rankings across schemes. The remaining models in the dendrogram attain mid-range rankings and exhibit higher variability across ranking schemes. Overall, \textit{Qwen3-Embedding-8B} emerges as the most robust model for pair classification in Spanish, followed by \textit{Octen-Embedding-8B} and \textit{Qwen3-Embedding-4B}.

\subsection{Semantic Textual Similarity}
\begin{table}[t]
  \caption{Most robust models on the STS task by language. Robustness is assessed with respect to dataset compositions (DS) and ranking schemes (RS).}
  \label{fig:summary_sts}
  \begin{center}
    \begin{small}
      \begin{sc}
        \begin{tabularx}{\columnwidth}{lccX}
          \toprule
          Lang. & DS & RS & Most Robust Models (STS) \\
          \midrule
          ENG & \checkmark & \checkmark & Octen-Embedding-8B \\
          FRA &  & \checkmark & Octen-Embedding-8B \\
          DEU &  & \checkmark & Octen-Embedding-8B \\
          HIN &  & \checkmark & llama-embed-nemotron-8b, Octen-Embedding-8B, Qwen3-Embedding-8B \\
          SPA &  & \checkmark & Octen-Embedding-8B, llama-embed-nemotron-8b, Qwen3-Embedding-8B \\
          \bottomrule
        \end{tabularx}
      \end{sc}
    \end{small}
  \end{center}
\end{table}

\begin{figure*}[ht]
  \begin{center}
    \centerline{\includegraphics[width=\textwidth]{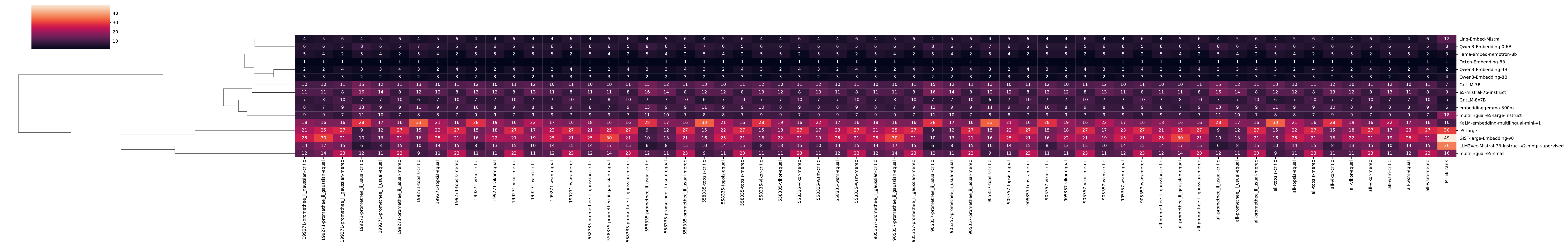}}
    \caption{
      Drendrogram for French language on the STS task.
    }
    \label{fig:fra_sts}
  \end{center}
\end{figure*}

In the STS task, two datasets are available for the \textbf{French} language, exhibiting pairwise correlations below the threshold of 0.9 and thus forming two separate clusters. Figure~\ref{fig:fra_sts} presents the dendrogram for this language–task combination, in which several clusters are clearly distinguishable. The best-performing cluster includes \textit{Octen-Embedding-8B}, which is consistently ranked first, together with the two \textit{Qwen3-Embedding} models with 8B and 4B parameters, both of which exhibit slightly higher variability across ranking schemes. Notably, \textit{Octen-Embedding-8B} is a fine-tuned variant of \textit{Qwen3-Embedding-8B}, optimized for semantic search and retrieval~\cite{octen-embedding-2025}. The second-best cluster comprises \textit{Linq-Embed-Mistral}, the smaller-scale \textit{Qwen3-Embedding-0.6B}, and \textit{llama-embed-nemotron-8b}. The large-scale models in this cluster are based on the e5-mistral-7b-instruct and Mistral-7B-v0.1 backbones for \textit{Linq-Embed-Mistral}, and on the Llama-3.1-8B LLM for \textit{llama-embed-nemotron-8b}. A cluster with mid-range performance includes the two GritLM variants (\textit{GritLM-8x7B} and \textit{GritLM-7B}), two models from the E5 family (\textit{multilingual-e5-large-instruct} and \textit{e5-mistral-7b-instruct}), and \textit{embeddinggemma-300m}. Among these, \textit{GritLM-8x7B}, based on the Mixtral-8x7B LLM, is the largest model (47B parameters), followed by \textit{GritLM-7B} and \textit{e5-mistral-7b-instruct}, both based on the 7B-parameter Mistral-7B-v0.1, the 300M-parameter \textit{embeddinggemma-300m} based on Gemma~3~\cite{vera2025embeddinggemma}, and \textit{multilingual-e5-large-instruct}, which is based on the XLM-RoBERTa encoder. The lowest-performing cluster, spanning models from \textit{KaLM-embedding-multilingual-mini-v1} to \textit{multilingual-e5-small}, exhibits both the lowest rankings and the highest variability across ranking schemes. Overall, for this language–task combination, the \textit{Qwen3-Embedding} models and their fine-tuned variant \textit{Octen-Embedding-8B} emerge as the most robust choices. More generally, the strongest performance is achieved by large-scale LLM-based text embedding models.

For \textbf{English}, ten datasets are available for the STS task and are grouped into five clusters. The first cluster comprises six datasets ([STS12–15, STSBenchmark, and SICK-R]), while each of the remaining clusters contains a single dataset. The dendrogram indicates that the best-performing cluster includes three Qwen3-based models (\textit{Octen-Embedding-8B}, \textit{Qwen3-Embedding-8B}, and \textit{Qwen3-Embedding-4B}), as well as the Llama-3.1-8B-based \textit{llama-embed-nemotron-8b}. Across ranking schemes, the three Qwen3-based models exhibit low sensitivity, whereas \textit{llama-embed-nemotron-8b} shows some sensitivity under uncorrelated subsamples. Across dataset compositions, all four models demonstrate low sensitivity to changes. The second-best cluster comprises the smaller-scale models \textit{multilingual-e5-large-instruct}, \textit{embeddinggemma-300m}, and \textit{Qwen3-Embedding-0.6B}, of which \textit{Qwen3-Embedding-0.6B} exhibits particularly high variability across ranking schemes. Nevertheless, all three models show low sensitivity to dataset composition. Subsequent clusters include smaller-scale multilingual models from the e5, gte, and bge-m3 families, followed by models based on the Mistral-7B and Mixtral-8x7B LLMs; these clusters exhibit higher ranking variability across schemes, and the latter also shows increased sensitivity to dataset composition. Overall, models from the best-performing cluster—most notably \textit{Octen-Embedding-8B}—emerge as the most robust choices for the STS task in English.

For \textbf{German}, three datasets are available for the STS task, each forming an individual cluster due to pairwise correlations below the specified threshold. Consequently, only a single dataset composition is available. The dendrogram indicates that the best-performing cluster is led by \textit{Octen-Embedding-8B}, which consistently attains the top rank, followed by \textit{Qwen3-Embedding-8B}, \textit{Qwen3-Embedding-4B}, and \textit{llama-embed-nemotron-8b}. The latter models exhibit greater variability across ranking schemes, particularly \textit{llama-embed-nemotron-8b}. The second-best cluster comprises \textit{Linq-Embed-Mistral}, \textit{Qwen3-Embedding-0.6B}, and \textit{multilingual-e5-large-instruct}, with \textit{Qwen3-Embedding-0.6B} showing slightly higher variability across ranking schemes. The subsequent cluster includes the LLM-based models \textit{GritLM-8x7B}, \textit{embeddinggemma-300m}, \textit{e5-mistral-7b-instruct}, and \textit{GritLM-7B}, all of which exhibit substantial ranking variability across schemes. Overall, \textit{Octen-Embedding-8B} emerges as the most robust model for the STS task in German.

For \textbf{Hindi}, two datasets are available for the STS task, each forming an individual cluster, resulting in a single dataset composition. The dendrogram indicates that the best-performing cluster comprises three large-scale LLM-based models: \textit{llama-embed-nemotron-8b}, \textit{Octen-Embedding-8B}, and \textit{Qwen3-Embedding-8B}. All three exhibit moderate variability across ranking schemes, with \textit{Qwen3-Embedding-8B} showing the lowest variability among them. These models are followed by \textit{Qwen3-Embedding-4B} and \textit{multilingual-e5-large-instruct}, which achieve slightly lower rankings, with the latter exhibiting higher variability across schemes. The second-best cluster includes \textit{bge-m3-custom-fr} and \textit{bge-m3}, both of which display pronounced ranking sensitivity. Subsequent clusters include \textit{embeddinggemma-300m} and \textit{gte-multilingual-base}, as well as a highly variable cluster comprising \textit{Qwen3-Embedding-0.6B}, \textit{Linq-Embed-Mistral}, \textit{e5-mistral-7b-instruct}, and \textit{GritLM-8x7B}. Overall, the most robust models for the STS task in Hindi are \textit{llama-embed-nemotron-8b}, \textit{Octen-Embedding-8B}, and \textit{Qwen3-Embedding-8B}.

For \textbf{Spanish}, three datasets are available for the STS task, each forming an individual cluster and resulting in a single dataset composition. The dendrogram indicates that the best-performing cluster comprises the LLM-based models \textit{Octen-Embedding-8B}, \textit{llama-embed-nemotron-8b}, \textit{Qwen3-Embedding-8B}, \textit{Linq-Embed-Mistral}, and \textit{Qwen3-Embedding-4B}. These models, however, exhibit notable variability across ranking schemes. They are followed by another highly variable cluster containing \textit{embeddinggemma-300m}, \textit{GritLM-8x7B}, \textit{GritLM-7B}, \textit{Qwen3-Embedding-0.6B}, \textit{multilingual-e5-large-instruct}, and \textit{e5-mistral-7b-instruct}. Overall, the most robust models for the STS task in Spanish are \textit{Octen-Embedding-8B}, \textit{llama-embed-nemotron-8b}, and \textit{Qwen3-Embedding-8B}.

\subsection{Bitext Mining}

\begin{table}[t]
  \caption{Most robust models on the bitext mining task by language. Robustness is assessed with respect to dataset compositions (DS) and ranking schemes (RS).}
  \label{fig:summary_bitext_mining}
  \begin{center}
    \begin{small}
      \begin{sc}
        \begin{tabularx}{\columnwidth}{lccX}
          \toprule
          Lang. & DS & RS & Most Robust Models (Bitext Mining) \\
          \midrule
          ENG & \checkmark & \checkmark & bge-m3, llama-embed-nemotron-8b, multilingual-e5-large-instruct \\
          FRA & \checkmark & \checkmark & bge-m3 \\
          DEU & \checkmark & \checkmark & bge-m3 \\
          HIN & \checkmark & \checkmark & bge-m3 \\
          SPA & \checkmark & \checkmark & bge-m3 \\
          \bottomrule
        \end{tabularx}
      \end{sc}
    \end{small}
  \end{center}
\end{table}

\begin{figure*}[ht]
  \begin{center}
    \centerline{\includegraphics[width=\textwidth]{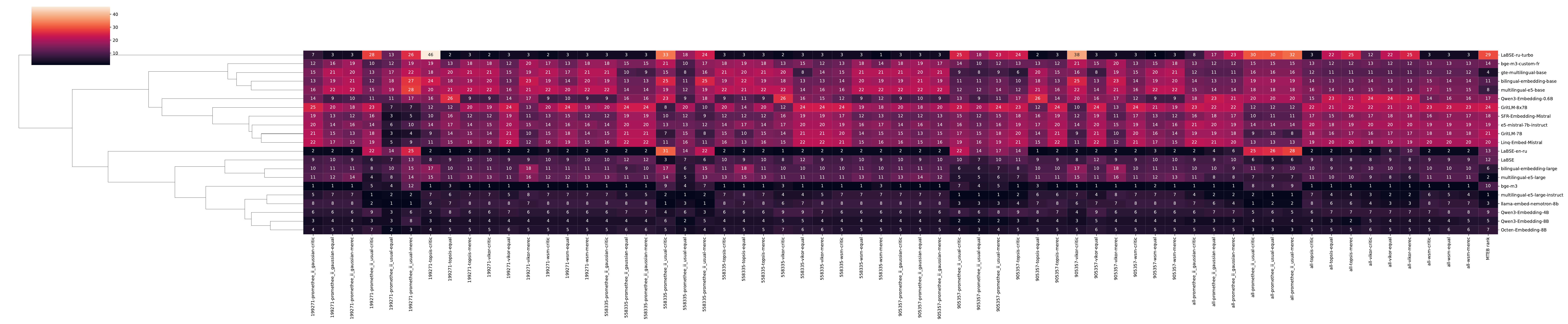}}
    \caption{
      Drendrogram for French language on the bitext mining task.
    }
    \label{fig:fra_bitext_mining}
  \end{center}
\end{figure*}

In the bitext mining task, six datasets involve the \textbf{French} language. One dataset, [BibleNLPBitextMining], exhibits low correlation (below 0.9) with the remaining datasets and therefore forms a separate cluster, while the other five datasets ([BUCC.v2, DiaBlaBitextMining, FloresBitextMining, NTREXBitextMining, Tatoeba]) form a single cluster. Figure~\ref{fig:fra_bitext_mining} presents the dendrogram of top-10 model rankings across ranking schemes and dataset compositions. The best-performing cluster spans models from \textit{bge-m3} to \textit{Octen-Embedding-8B}. The model \textit{bge-m3} is most frequently ranked first, exhibiting moderate variability across ranking schemes and low sensitivity to dataset composition. This model supports long input sequences and is optimized via a self-knowledge distillation approach to enable multiple retrieval paradigms, including dense, sparse, and multi-vector retrieval~\cite{chen2024bge}. It is followed by the Qwen3-based models \textit{Qwen3-Embedding-8B}, its fine-tuned variant \textit{Octen-Embedding-8B}~\cite{octen-embedding-2025}, and the smaller-scale \textit{Qwen3-Embedding-4B}, all of which exhibit the lowest ranking variability among the models shown. This cluster also includes the Llama-3.1-8B-based \textit{llama-embed-nemotron-8b} and the XLM-RoBERTa-based \textit{multilingual-e5-large-instruct}, both of which display higher ranking variability but low sensitivity to dataset composition. The second cluster comprises three BERT- and XLM-RoBERTa-based models: \textit{LaBSE}, \textit{bilingual-embedding-large}, and \textit{multilingual-e5-large}. \textit{LaBSE} is a BERT-based model fine-tuned for bilingual tasks across more than 100 languages~\cite{feng2022language}, while \textit{bilingual-embedding-large} and \textit{multilingual-e5-large} are XLM-RoBERTa-based models, the first specifically optimized for bilingual tasks involving French and English language, while the second optimized through weakly-supervised contrastive pre-training on more than 1B multilingual text pairs, followed by supervised fine-tuning~\cite{wang2024multilingual}. The third cluster spans models from \textit{bge-m3-custom-fr} to \textit{multilingual-e5-base}, representing base versions of several larger models. These models achieve mid-range rankings with high variability across ranking schemes, particularly for uncorrelated subsamples, and exhibit sensitivity to dataset composition. The fourth cluster consists of Mistral-7B-v0.1-based models (\textit{SFR-Embedding-Mistral}, \textit{e5-mistral-7b-instruct}, \textit{GritLM-7B}, and \textit{Linq-Embed-Mistral}) and the Mixtral-8x7B-based \textit{GritLM-8x7B}. Despite their scale, these models attain only mid-range rankings and show noticeable ranking variability and sensitivity to dataset composition. The remaining models exhibit very high variability, ranking highly under some schemes and poorly under others. Overall, models in the best-performing cluster represent reasonable choices for French bitext mining. However, no single model simultaneously achieves consistently high rankings, low variability across ranking schemes, and low sensitivity to dataset composition. While Qwen3-based models generally perform well, Mistral-7B-based models tend to achieve only mid-range performance on this task.

For \textbf{English}, ten datasets are available and grouped into two clusters, one containing nine datasets and the other a single dataset. The dendrogram indicates that one cluster clearly outperforms the others and includes the models \textit{bge-m3}, \textit{llama-embed-nemotron-8b}, \textit{multilingual-e5-large-instruct}, and the three Qwen3-based models \textit{Qwen3-Embedding-8B}, \textit{Octen-Embedding-8B}, and \textit{Qwen3-Embedding-4B}. The first two models exhibit high ranking sensitivity across ranking schemes for all dataset compositions, whereas the three Qwen3-based models show high sensitivity on uncorrelated subsamples but substantially decreased sensitivity when all datasets are included. This pattern indicates greater sensitivity to dataset composition for the latter models. The second-best cluster comprises models such as \textit{LaBSE}, \textit{bilingual-embedding-large}, and \textit{multilingual-e5-large}, which exhibit more stable rankings for the dataset composition including all datasets than for uncorrelated subsets, again indicating sensitivity to dataset composition. Similar behavior is observed for \textit{gte-multilingual-base}, \textit{Qwen3-Embedding-0.6B}, and \textit{bge-m3-custom-fr} within the same cluster, although these models generally attain slightly lower rankings. The third cluster consists of large-scale models based on the Mistral-7B and Mixtral-8x7B backbones, including \textit{e5-mistral-7b-instruct}, \textit{SFR-Embedding-Mistral}, \textit{GritLM-7B}, and \textit{GritLM-8x7B}, which achieve mid-range rankings for this language–task combination. Overall, models from the best-performing cluster represent strong candidates; however, their increased sensitivity across ranking schemes should be carefully considered.

For \textbf{German}, five datasets are available and grouped into two clusters, with the first cluster comprising four datasets and the second containing a single dataset. The resulting dendrogram reveals five major clusters, of which the top-performing cluster spans models from \textit{bge-m3} to \textit{llama-embed-nemotron-8b}. Although \textit{bge-m3} is ranked first by a large number of ranking schemes, it exhibits substantial variability across schemes. The models \textit{multilingual-e5-large-instruct}, \textit{llama-embed-nemotron-8b}, \textit{Qwen3-Embedding-8B}, and \textit{Octen-Embedding-8B} achieve slightly lower rankings but demonstrate low sensitivity across ranking schemes, particularly the latter two. The model \textit{multilingual-e5-large-instruct} also shows high sensitivity to dataset composition under certain ranking schemes. The models \textit{LaBSE}, \textit{bilingual-embedding-large}, and \textit{multilingual-e5-large} attain mid-range rankings with comparatively lower variability across ranking schemes. In contrast, \textit{LaBSE-en-ru} and \textit{LaBSE-ru-turbo} exhibit highly inconsistent rankings, achieving very high positions under some schemes and low positions under others. Overall, models in the best-performing cluster constitute strong candidates for this language–task combination, provided that ranking variability across schemes is carefully considered.

For \textbf{Hindi}, six datasets are available and grouped into two clusters, with the first cluster comprising five datasets and the second containing a single dataset. The resulting dendrogram reveals four model clusters. In the top-performing cluster, the model \textit{bge-m3} achieves the highest ranking, although with moderate variability across ranking schemes. The same cluster also includes \textit{Qwen3-Embedding-8B} and \textit{Octen-Embedding-8B}, which achieve slightly lower rankings but exhibit low variability across schemes. These are followed by \textit{llama-embed-nemotron-8b} and \textit{multilingual-e5-large-instruct}, which achieve comparable rankings but with higher variability, and by \textit{Qwen3-Embedding-4B}, which shows slightly lower rankings and increased variability. The models \textit{LaBSE}, \textit{bilingual-embedding-large}, and \textit{multilingual-e5-large} form a cluster with mid-range rankings and moderate variability. Overall, models in the best-performing cluster represent reasonable choices for this language–task combination, provided that ranking variability across schemes is taken into consideration.

For \textbf{Spanish}, four datasets are available and grouped into two clusters, with the first cluster comprising three datasets and the second containing a single dataset. The resulting dendrogram reveals four model clusters. In the best-performing cluster, the model \textit{bge-m3} attains the highest ranking; however, it exhibits substantial variability across ranking schemes. The same cluster also includes \textit{llama-embed-nemotron-8b}, \textit{multilingual-e5-large-instruct}, \textit{Qwen3-Embedding-8B}, and \textit{Octen-Embedding-8B}, which achieve slightly lower rankings but demonstrate reduced variability, particularly the latter two models. The second cluster comprises \textit{Qwen3-Embedding-4B}, \textit{LaBSE}, \textit{bilingual-embedding-large}, and \textit{multilingual-e5-large}, all of which obtain lower rankings than models in the first cluster. With the exception of \textit{multilingual-e5-large}, these models also exhibit lower variability across ranking schemes. Overall, while \textit{bge-m3} appears as the top-ranked model, its high ranking variability suggests that other models in the first cluster may also represent preferable choices in practice.

\subsection{Multilabel Classification}
\begin{table}[t]
  \caption{Most robust models on the multilabel classification task by language. Robustness is assessed with respect to dataset compositions (DS) and ranking schemes (RS).}
  \label{fig:summary_multilabel_classification}
  \begin{center}
    \begin{small}
      \begin{sc}
        \begin{tabularx}{\columnwidth}{lccX}
          \toprule
          Lang. & DS & RS & Most Robust Models (Multilabel Classification) \\
          \midrule
          ENG &  & \checkmark & KaLM-Embedding-Gemma3-12B-2511 \\
          FRA &  & \checkmark & KaLM-Embedding-Gemma3-12B-2511 \\
          DEU &  & \checkmark & KaLM-Embedding-Gemma3-12B-2511 \\
          HIN &  & & Not applicable \\
          SPA &  & \checkmark & KaLM-Embedding-Gemma3-12B-2511 \\
          \bottomrule
        \end{tabularx}
      \end{sc}
    \end{small}
  \end{center}
\end{table}

\begin{figure}[ht]
  \begin{center}
    \centerline{\includegraphics[width=0.4\columnwidth]{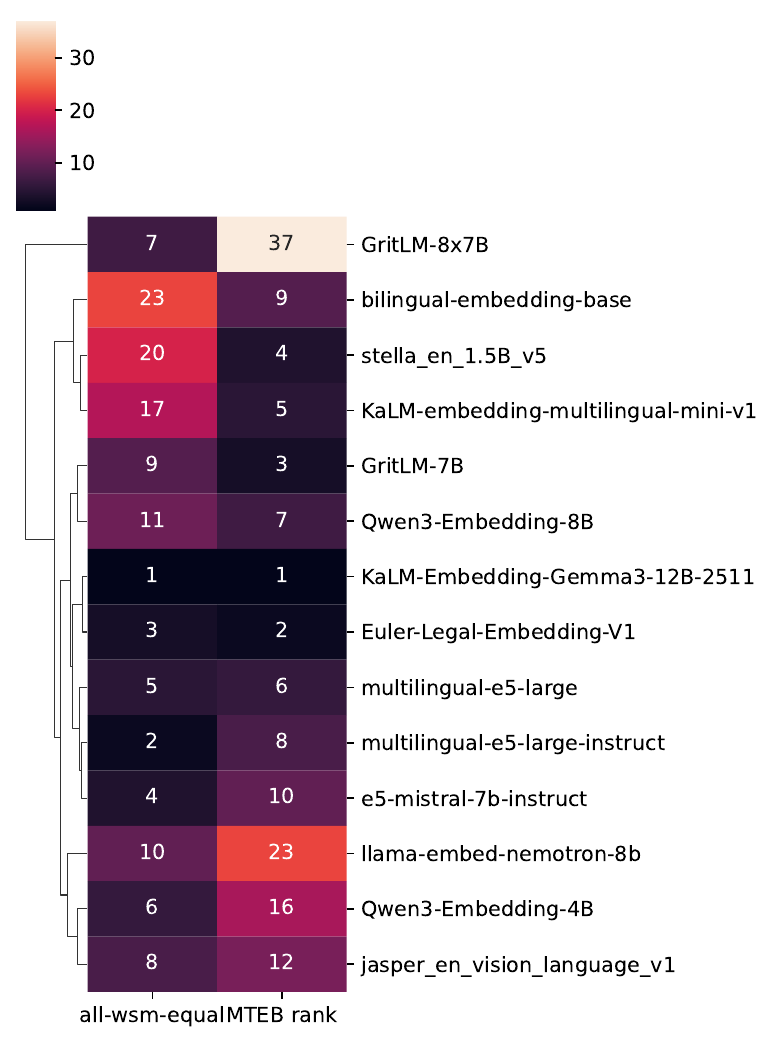}}
    \caption{
      Drendrogram for French language on the multilabel classification task.
    }
    \label{fig:fra_multilabel_classification}
  \end{center}
\end{figure}

In the multilabel classification task, only a single dataset is available for the \textbf{French} language. Consequently, model rankings are determined directly by their performance on this dataset and ordered in descending order. For simplicity, Figure~\ref{fig:fra_multilabel_classification} presents this ranking using the WSM ranking scheme with equal weighting, which yields an identical performance score and ranking. The top-ranked model is the large-scale \textit{KaLM-Embedding-Gemma3-12B-2511}, followed by the smaller-scale \textit{multilingual-e5-large-instruct}. The third- and fourth-ranked models are the large-scale Qwen3-based \textit{Euler-Legal-Embedding-V1} and the Mistral-7B-based \textit{e5-mistral-7b-instruct}, respectively, followed by \textit{multilingual-e5-large} and \textit{Qwen3-Embedding-4B}. These results indicate that, while the best-performing model is a large-scale model (approximately 12B parameters), strong performance is also achieved by smaller-scale models (approximately 0.6B parameters), highlighting a trade-off that may be advantageous for different deployment scenarios.

The results for the \textbf{English} language are similar. On the single dataset associated with this language, the top-ranked model is the large-scale \textit{KaLM-Embedding-Gemma3-12B-2511}, followed by the XLM-RoBERTa-based \textit{multilingual-e5-large-instruct}, the Qwen3-based \textit{Euler-Legal-Embedding-V1}, the Mistral-7B-based \textit{e5-mistral-7b-instruct}, and the XLM-RoBERTa-based \textit{multilingual-e5-large}. As in the French case, the best-performing models comprise a combination of large-scale LLM-based models, such as \textit{KaLM-Embedding-Gemma3-12B-2511}, \textit{Euler-Legal-Embedding-V1}, and \textit{e5-mistral-7b-instruct} and smaller-scale models, including \textit{multilingual-e5-large-instruct} and \textit{multilingual-e5-large}, suitable for different use-cases.

For \textbf{German}, only a single dataset is available for multilabel classification, and models are ranked directly based on their performance on this dataset. The top-ranked model is the large-scale \textit{KaLM-Embedding-Gemma3-12B-2511}, followed by the smaller-scale XLM-RoBERTa-based \textit{multilingual-e5-large-instruct}, the Qwen3-based \textit{Euler-Legal-Embedding-V1}, the Mistral-7B-based \textit{e5-mistral-7b-instruct}, and \textit{multilingual-e5-large}. The models \textit{Qwen3-Embedding-4B} and \textit{GritLM-8x7B} follow. Overall, the most effective models for multilabel classification in German comprise a combination of large-scale LLM-based models and smaller-scale models.

There are no datasets in \textbf{Hindi} for the multilabel classification task.

For \textbf{Spanish}, only a single dataset is available for the multiclass classification task, and models are ranked directly based on their performance on this dataset. The top-ranked model is \textit{KaLM-Embedding-Gemma3-12B-2511}, followed by \textit{multilingual-e5-large-instruct}, \textit{Euler-Legal-Embedding-V1}, \textit{e5-mistral-7b-instruct}, and \textit{multilingual-e5-large}. Overall, the most effective models for this language–task combination include both large-scale LLM-based models and smaller-scale models.

\subsection{Instruction Reranking}
\begin{table}[t]
  \caption{Most robust models on the instruction reranking task by language. Robustness is assessed with respect to dataset compositions (DS) and ranking schemes (RS).}
  \label{fig:summary_instruction_reranking}
  \begin{center}
    \begin{small}
      \begin{sc}
        \begin{tabularx}{\columnwidth}{lccX}
          \toprule
          Lang. & DS & RS & Most Robust Models (Instruction Reranking) \\
          \midrule
          ENG &  & \checkmark & Qwen3-Embedding-4B \\
          FRA &  &  & Not applicable \\
          DEU &  &  & Not applicable \\
          HIN &  &  & Not applicable \\
          SPA &  &  & Not applicable \\
          \bottomrule
        \end{tabularx}
      \end{sc}
    \end{small}
  \end{center}
\end{table}

In the instruction reranking task, datasets are available only for the \textbf{English} language; no datasets are found for \textbf{French}, \textbf{German}, \textbf{Hindi}, or \textbf{Spanish}. In English, there are three datasets, each forming an individual cluster, resulting in a single dataset composition that includes all datasets. The corresponding dendrogram reveals three main clusters. The best-performing cluster comprises three models, with \textit{Qwen3-Embedding-4B} consistently ranked first, followed by \textit{llama-embed-nemotron-8b} and \textit{Qwen3-Embedding-8B}. The top-ranked model exhibits highly stable rankings across schemes, while the latter two show slightly greater variability. The second cluster achieves marginally lower rankings and includes \textit{Octen-Embedding-8B} and \textit{Euler-Legal-Embedding-V1}, which attain the highest and most stable rankings within this cluster, alongside \textit{Qwen3-Embedding-0.6B}, \textit{embeddinggemma-300m}, and \textit{KaLM-Embedding-Gemma3-12B-2511}. The third cluster contains the remaining models and exhibits the lowest rankings across all ranking schemes. Overall, \textit{Qwen3-Embedding-4B} emerges as the most robust option for instruction reranking in English, followed by \textit{llama-embed-nemotron-8b} and \textit{Qwen3-Embedding-8B}.

\section{Language-Task Agnostic Analysis (Task-Agnostic Within a Language)}
\label{appendix:task_agnostic}

This Appendix presents analysis of the task-agnostic results for French, English, German, Hindi, and Spanish. 

\begin{figure*}[ht]

  \begin{center}
    \centerline{\includegraphics[width=\textwidth]{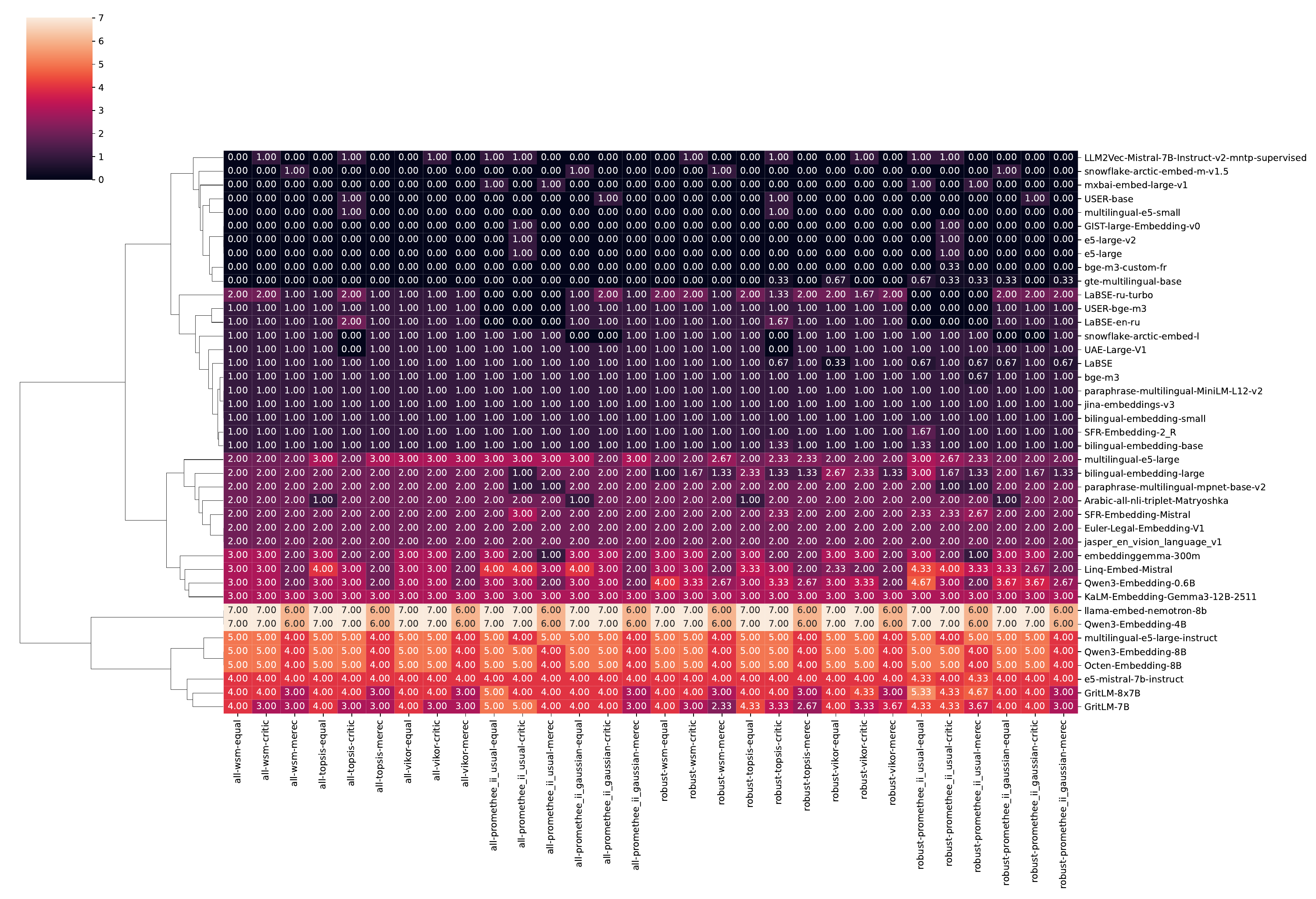}}
    \caption{
      Cross-task consistency by a ranking scheme for French language.
    }
    \label{fig:fra_consistency_scheme}
  \end{center}
\end{figure*}

Figure~\ref{fig:fra_consistency_scheme} presents the dendrogram obtained by clustering models based on their cross-task (CT) consistency scores across ranking schemes for the French language. The dendrogram reveals four major clusters, each further decomposing into multiple subclusters. The first cluster, which contains models with the highest cross-task consistency (with higher values indicating greater robustness), spans models from \textit{llama-embed-nemotron-8b} to \textit{GritLM-7B}. This cluster is predominantly composed of large-scale LLM-derived embedding models that fine-tune a core language model to learn general-purpose semantic spaces that transfer reliably across tasks. Their high cross-task consistency suggests that they capture task-invariant semantic spaces rather than overfitting to specific evaluation settings, making them robust task-agnostic encoders for French. The first subcluster includes \textit{llama-embed-nemotron-8b} and \textit{Qwen3-Embedding-4B}, which appear among the top-10 ranked models in 6–7 out of the nine tasks across all ranking schemes. This CT score pattern is consistent across both dataset compositions, including the full dataset composition and the robust, uncorrelated subsamples, identifying these models as the most robust task-agnostic encoders among those considered for French. The next subcluster comprises \textit{Qwen3-Embedding-8B}, \textit{Octen-Embedding-8B}, and \textit{multilingual-e5-large-instruct}. These models appear among the top 10 in 4–5 of the nine tasks, with consistent behavior across both full and robust dataset compositions. While the first two are Qwen3-based models, the latter is a smaller-scale (0.6B-parameter) model based on XLM-RoBERTa. The final subcluster includes two Mistral-7B-based models (\textit{e5-mistral-7b-instruct} and \textit{GritLM-7B}) and one Mixtral-8x7B-based model (\textit{GritLM-8x7B}), which appear among the top 10 in 3–5 of the nine tasks across ranking schemes. Among these, \textit{e5-mistral-7b-instruct} exhibits the highest robustness.

The second cluster spans from \textit{multilingual-e5-large} to \textit{KaLM-Embedding-Gemma3-12B-2511} and represents a mix of models of different scales and architectures. Semantically, this cluster reflects embedding models whose learned representation spaces are only weakly aligned with the evaluated tasks and whose performance is more sensitive to evaluation conditions. The subcluster containing models appearing in 2-4 tasks out of 9 contains two Gemma3-based models of different scales, one large-scale (12B parameters) \textit{KaLM-Embedding-Gemma3-12B-2511} and one smaller-scale (300M parameters) \textit{embeddinggemma-300m}. The large-scale model \textit{Linq-Embed-Mistral} based on Mistral-7B and a small-scale model \textit{Qwen3-Embedding-0.6B} based on Qwen3 also appear in this cluster. Model \textit{KaLM-Embedding-Gemma3-12B-2511} has consistent performance across all dataset compositions, as well as \textit{Linq-Embed-Mistral}, while \textit{multilingual-e5-large} is less consistent on the uncorrelated dataset compositions. 

The third cluster spans \textit{LaBSE-ru-turbo} to \textit{bilingual-embedding-base}, including models that appear as top-10 ranked in 1-2 tasks of 9, while the fourth cluster, spanning from \textit{LLM2Vec-Mistral-7B-Instruct-v2-mntp-supervised} to \textit{gte-multilingual-base}, includes models that appear in 0-1 tasks of 9. This cluster captures embedding models whose semantic spaces show weak alignment with the evaluated tasks and poor task-agnostic robustness, indicating limited general-purpose applicability.

\begin{figure*}[ht]
  \begin{center}
    \centerline{\includegraphics[width=\textwidth]{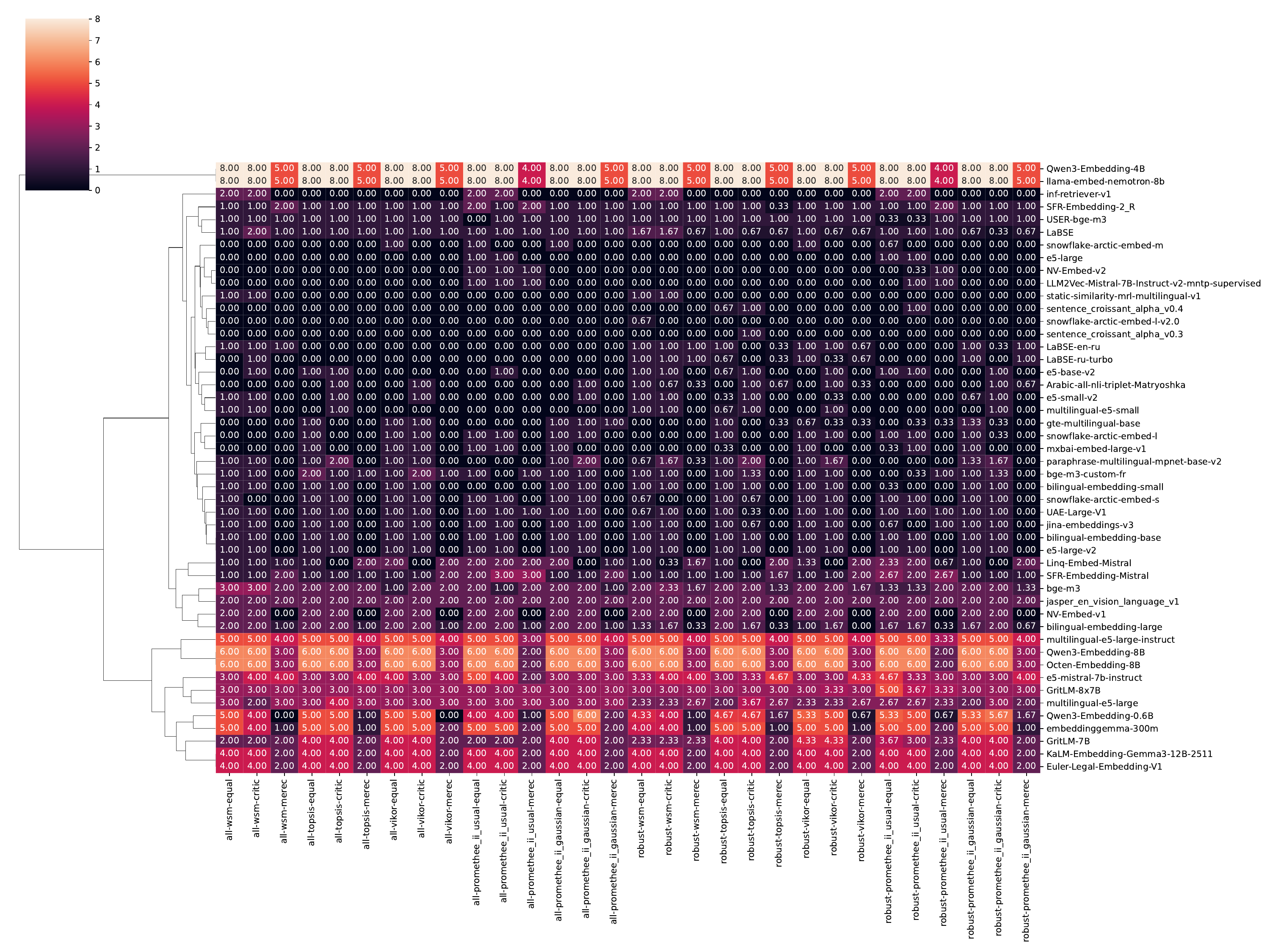}}
    \caption{
      Cross-task consistency by ranking scheme for English language.
    }
    \label{fig:eng_consistency_scheme}
  \end{center}
\end{figure*}

The task-agnostic results for the \textbf{English} language are shown in Figure~\ref{fig:eng_consistency_scheme}. The two most robust models are again \textit{llama-embed-nemotron-8b} and \textit{Qwen3-Embedding-4B}, which appear among the top-10 ranked models in eight of the nine tasks for most ranking schemes. This number decreases to five only for ranking schemes employing the MEREC weighting method, a pattern that is consistent across both the dataset composition including all datasets and the uncorrelated subsamples. The second-best cluster spans models from \textit{multilingual-e5-large-instruct} to \textit{Euler-Legal-Embedding-V1}. The models \textit{Qwen3-Embedding-8B} and \textit{Octen-Embedding-8B} appear among the top 10 in six tasks under most ranking schemes; however, this number drops to two or three under MEREC-based schemes, resulting in higher variability in their CT scores. In contrast, \textit{multilingual-e5-large-instruct} exhibits lower variability, typically appearing among the top 10 in four to five tasks across all ranking schemes. Substantial variability in task coverage, ranging from zero to five tasks, is observed for the smaller-scale models \textit{Qwen3-Embedding-0.6B} and \textit{embeddinggemma-300m}. More stable task coverage, predominantly between three and four tasks, is observed for the large-scale models \textit{e5-mistral-7b-instruct} and \textit{GritLM-8x7B}. The models \textit{GritLM-7B}, \textit{KaLM-Embedding-Gemma3-12B-2511}, and \textit{Euler-Legal-Embedding-V1} also exhibit high variability, with appearances ranging between two and four tasks. Overall, \textit{llama-embed-nemotron-8b} and \textit{Qwen3-Embedding-4B} emerge as the most robust task-agnostic models for English among those illustrated in the dendrogram.

\begin{figure*}[ht]
  \begin{center}
    \centerline{\includegraphics[width=\textwidth]{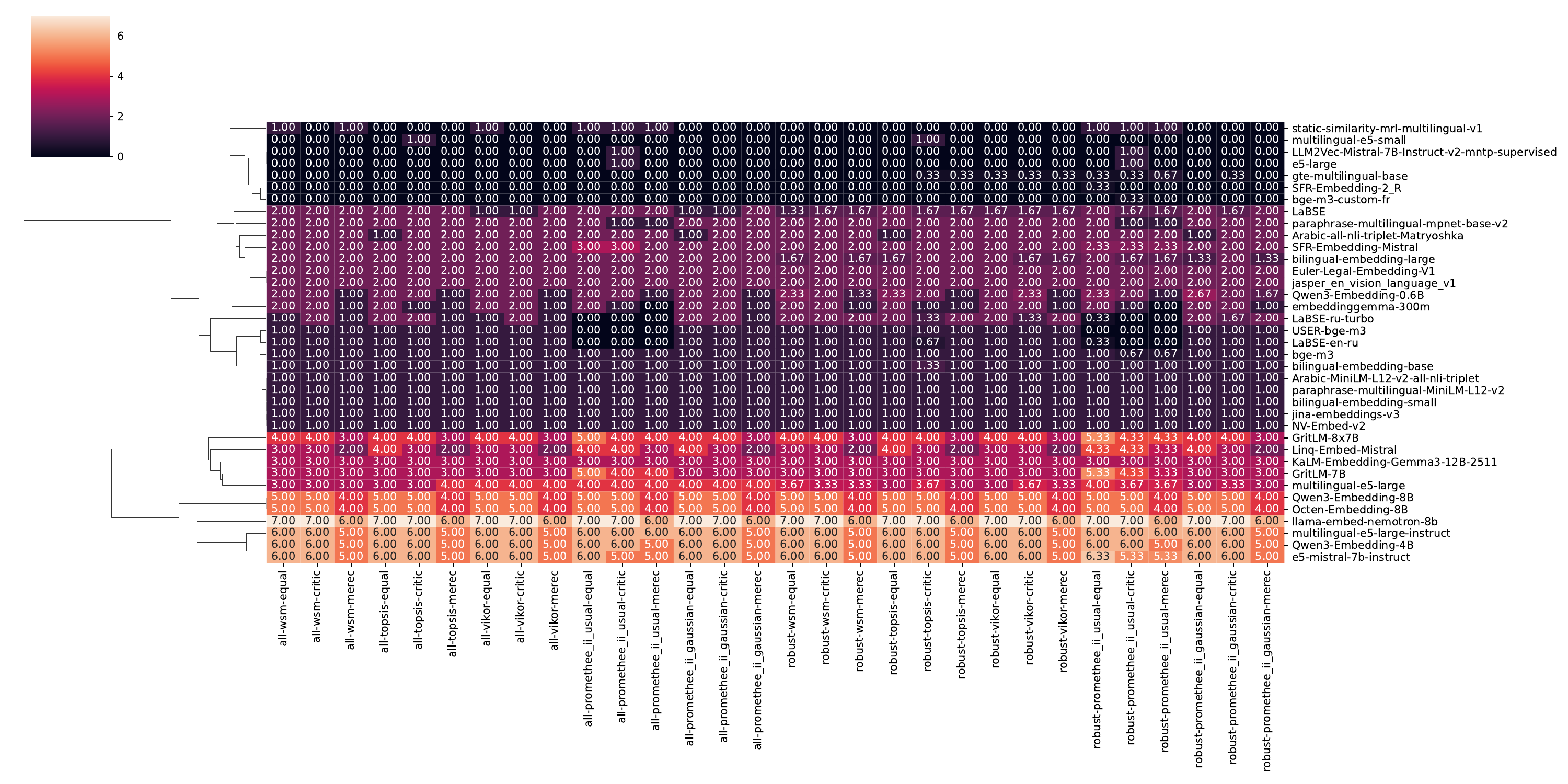}}
    \caption{
      Cross-task consistency by ranking scheme for German language.
    }
    \label{fig:deu_consistency_scheme}
  \end{center}
\end{figure*}

For the \textbf{German} language, task-agnostic results are shown in Figure~\ref{fig:deu_consistency_scheme}. The dendrogram reveals five major clusters of models. The model \textit{llama-embed-nemotron-8b} consistently appears among the top-10 ranked models in 6–7 of the nine tasks across ranking schemes. The models \textit{multilingual-e5-large-instruct}, \textit{Qwen3-Embedding-4B}, and \textit{e5-mistral-7b-instruct} follow, appearing in 5–6 tasks. The models \textit{Qwen3-Embedding-8B} and \textit{Octen-Embedding-8B} appear in 4–5 tasks, while \textit{GritLM-8x7B} appears almost consistently in four tasks. For all of these models, the number of tasks decreases under ranking schemes employing the MEREC weighting method. Large-scale models such as \textit{Linq-Embed-Mistral}, \textit{KaLM-Embedding-Gemma3-12B-2511}, and \textit{GritLM-7B}, together with \textit{multilingual-e5-large}, exhibit more stable CT scores of approximately three across all ranking schemes and dataset compositions. The remaining models in the dendrogram appear in two or fewer tasks across ranking schemes and dataset compositions. Overall, \textit{llama-embed-nemotron-8b} emerges as the most robust task-agnostic model for the German language.

For the \textbf{Hindi} language, the models \textit{Octen-Embedding-8B}, \textit{Qwen3-Embedding-8B}, and \textit{llama-embed-nemotron-8b} achieve the highest CT scores across ranking schemes and dataset compositions, appearing among the top-10 ranked models in six of the nine tasks for all ranking schemes except those employing the MEREC weighting method, where the number of tasks decreases to five. For \textit{Qwen3-Embedding-4B}, the number of tasks decreases to five and four, respectively, under these schemes. Similar behavior is observed for \textit{multilingual-e5-large-instruct}, which appears in three to four tasks across ranking schemes. The models \textit{paraphrase-multilingual-mpnet-base-v2} and \textit{Arabic-all-nli-triplet-Matryoshka} exhibit consistent task coverage, appearing in three tasks. In contrast, \textit{embeddinggemma-300m} and \textit{LaBSE} show greater variability, appearing in one to two tasks. Overall, \textit{Octen-Embedding-8B}, \textit{Qwen3-Embedding-8B}, and \textit{llama-embed-nemotron-8b} emerge as the most robust task-agnostic models for Hindi.

\begin{figure*}[ht]
  \begin{center}
    \centerline{\includegraphics[width=\textwidth]{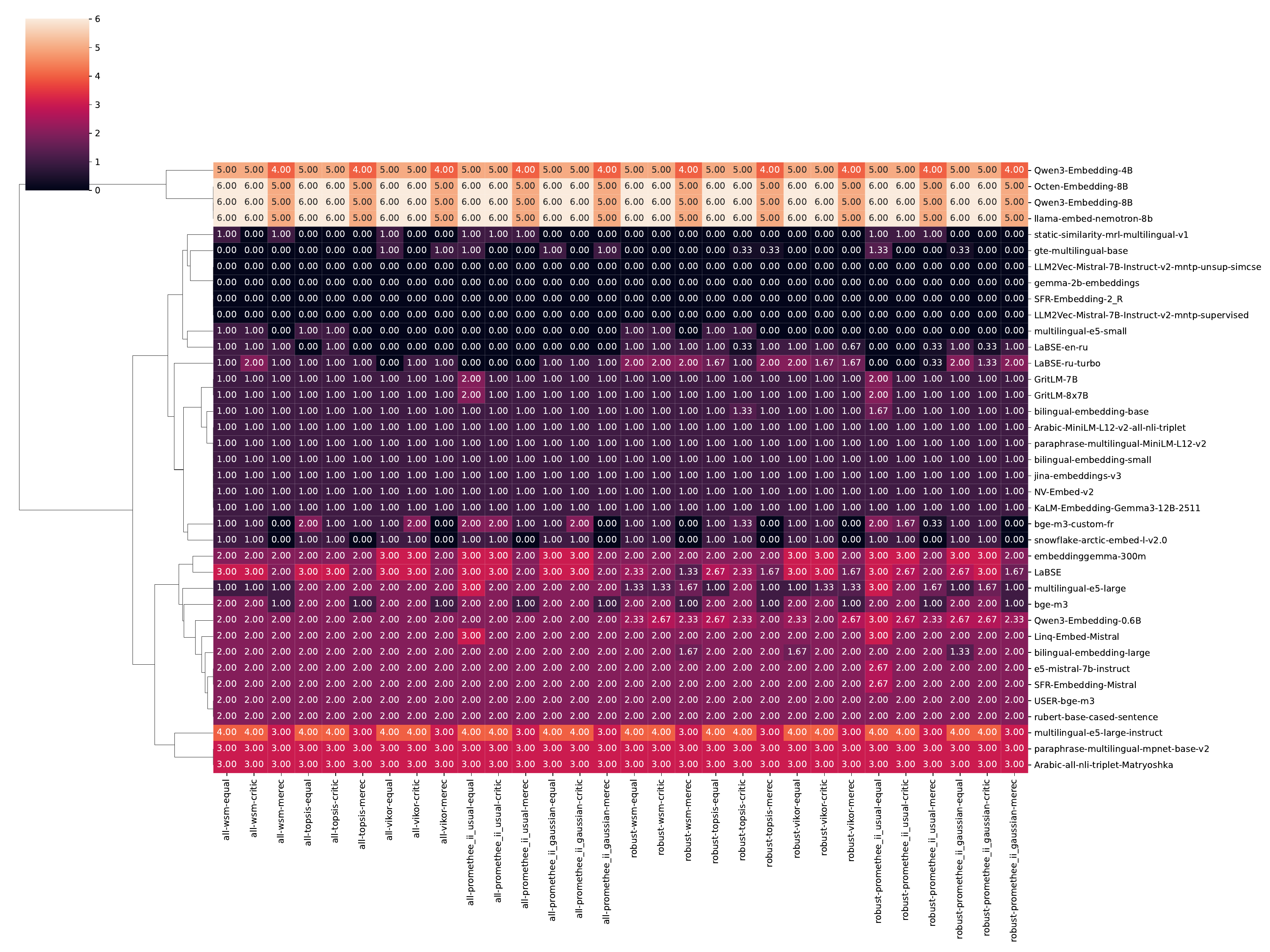}}
    \caption{
      Cross-task consistency by ranking scheme for Hindi language.
    }
    \label{fig:hin_consistency_scheme}
  \end{center}
\end{figure*}

\begin{figure*}[ht]
  \begin{center}
    \centerline{\includegraphics[width=\textwidth]{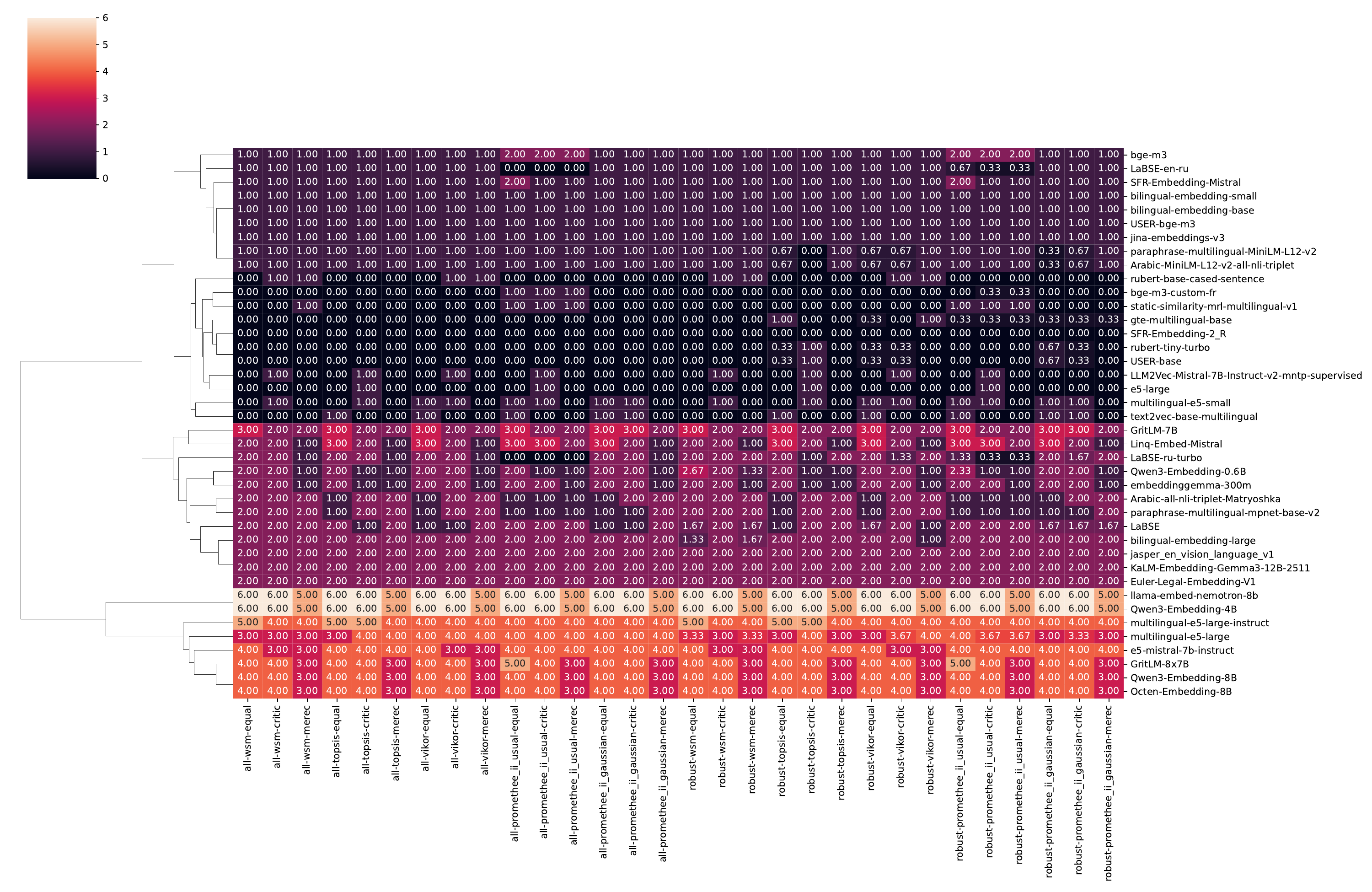}}
    \caption{
      Cross-task consistency by ranking scheme for Spanish language.
    }
    \label{fig:spa_consistency_scheme}
  \end{center}
\end{figure*}

Finally, for the \textbf{Spanish} language, the task-agnostic results are shown in Figure~\ref{fig:spa_consistency_scheme}. The models \textit{llama-embed-nemotron-8b} and \textit{Qwen3-Embedding-4B} achieve the highest CT scores across ranking schemes, appearing among the top-10 ranked models in six of the nine tasks for most schemes; under ranking schemes employing the MEREC weighting method, this number decreases to five. These models are followed by \textit{multilingual-e5-large-instruct}, which appears in four to five tasks across ranking schemes. The models \textit{multilingual-e5-large}, \textit{e5-mistral-7b-instruct}, \textit{GritLM-8x7B}, \textit{Qwen3-Embedding-8B}, and \textit{Octen-Embedding-8B} exhibit greater variability, appearing in three to four tasks. The models \textit{GritLM-7B} and \textit{Linq-Embed-Mistral} follow, with appearances ranging between two and three tasks, while all remaining models in the dendrogram appear in two or fewer tasks. Overall, \textit{llama-embed-nemotron-8b} and \textit{Qwen3-Embedding-4B} emerge as the most robust task-agnostic models for Spanish.

Table~\ref{table:summary_task_agnostic} summarizes the most robust task-agnostic models for the five analyzed languages (English, French, German, Hindi, and Spanish). The table is dominated by large-scale embedding models based on either the Qwen3 or Llama-3.1-8B core architectures, namely \textit{Qwen3-Embedding-4B}, \textit{llama-embed-nemotron-8b}, \textit{Qwen3-Embedding-8B}, and \textit{Octen-Embedding-8B}.

\begin{table}[t]
  \caption{Most robust task-agnostic models by language.}
  \label{table:summary_task_agnostic}
  \begin{center}
    \begin{small}
      \begin{sc}
        \begin{tabularx}{\columnwidth}{lX}
          \toprule
          Lang. & Most Robust Task-Agnostic Models  \\
          \midrule
          ENG & Qwen3-Embedding-4B, llama-embed-nemotron-8b \\
          FRA & Qwen3-Embedding-4B, llama-embed-nemotron-8b \\
          DEU & llama-embed-nemotron-8b \\
          HIN & Octen-Embedding-8B, Qwen3-Embedding-8B, llama-embed-nemotron-8b \\
          SPA & Qwen3-Embedding-4B, llama-embed-nemotron-8b \\
          \bottomrule
        \end{tabularx}
      \end{sc}
    \end{small}
  \end{center}
\end{table}

\section{Selected MCDM Aggregation Methods}
\label{appendix:rankingMethods}

This section gives additional details of the four selected MCDM aggregation methods in terms of their mathematical formulation.

The \textit{Weighted Sum Model (WSM)} is a simple MCDM method corresponding to a ranking scheme $s \in \mathcal{K}$. 
Given a performance matrix $P^{(q,\ell)} \in \mathbb{R}^{m_\ell \times n_{q,\ell}}$ for task $\mathcal{T}_q$ and language $\ell$, 
and a vector of criteria weights $w^{(q,\ell)} \in \mathbb{R}^{n_{q,\ell}}$, 
the score of model $i \in \mathcal{M}^{(\ell)}$ is computed with Eq.~\ref{eq:wsm}.

\begin{equation}
s^{(s,q,\ell)}_i = \sum_{j \in \mathcal{D}^{(q,\ell)}} w^{(q,\ell)}_j \, P^{(q,\ell)}_{i,j}.
\label{eq:wsm}
\end{equation}

If necessary, the performance values $P^{(q,\ell)}_{i,j}$ are normalized beforehand to account for 
maximization or minimization criteria\footnote{https://pymcdm.readthedocs.io/en/master/modules/american\_school.html\#wsm}. The final ranking is obtained by sorting the scores in descending order, where smaller ranks indicate better performance.

The \textit{Technique for Order Preference by Similarity to an Ideal Solution (TOPSIS)} 
is a ranking scheme $s \in \mathcal{K}$ that ranks models based on their Euclidean distance 
to two reference points: a positive ideal solution and a negative ideal solution. 
The method prefers models that are close to the positive ideal and far from the negative one.

Given a performance matrix $P^{(q,\ell)} \in \mathbb{R}^{m_\ell \times n_{q,\ell}}$ 
and a vector of criteria weights $w^{(q,\ell)} \in \mathbb{R}^{n_{q,\ell}}$, 
the method proceeds as follows:

\textbf{(1) Normalization.} The performance values are normalized as
\begin{equation}
\hat{P}^{(q,\ell)}_{i,j} = 
\frac{P^{(q,\ell)}_{i,j}}{\sqrt{\sum_{z \in \mathcal{M}^{(\ell)}} \left(P^{(q,\ell)}_{z,j}\right)^2}}.
\label{eq:topsis.normalization}
\end{equation}

\textbf{(2) Weighted normalization.} The weighted normalized matrix is
\begin{equation}
V^{(q,\ell)}_{i,j} = w^{(q,\ell)}_j \, \hat{P}^{(q,\ell)}_{i,j}.
\end{equation}

\textbf{(3) Positive and negative ideal solutions.} For each criterion $j \in \mathcal{D}^{(q,\ell)}$, 
the ideal values are defined as
\begin{equation}
v^{(q,\ell)+}_j =
\begin{cases}
\max\limits_{i \in \mathcal{M}^{(\ell)}} V^{(q,\ell)}_{i,j}, & \text{if } j \text{ is to be maximized}, \\
\min\limits_{i \in \mathcal{M}^{(\ell)}} V^{(q,\ell)}_{i,j}, & \text{if } j \text{ is to be minimized},
\end{cases}
\label{eq:topsis.positive.value}
\end{equation}

\begin{equation}
v^{(q,\ell)-}_j =
\begin{cases}
\min\limits_{i \in \mathcal{M}^{(\ell)}} V^{(q,\ell)}_{i,j}, & \text{if } j \text{ is to be maximized}, \\
\max\limits_{i \in \mathcal{M}^{(\ell)}} V^{(q,\ell)}_{i,j}, & \text{if } j \text{ is to be minimized}.
\end{cases}
\label{eq:topsis.negative.value}
\end{equation}

\textbf{(4) Distance to ideal solutions.} The Euclidean distance of model $i$ to the positive and negative ideal solutions is
\begin{equation}
d_i^{(q,\ell)+} = \sqrt{\sum_{j \in \mathcal{D}^{(q,\ell)}} \left(V^{(q,\ell)}_{i,j} - v^{(q,\ell)+}_j \right)^2},
\label{eq:topsis.distance.pos}
\end{equation}
\begin{equation}
d_i^{(q,\ell)-} = \sqrt{\sum_{j \in \mathcal{D}^{(q,\ell)}} \left(V^{(q,\ell)}_{i,j} - v^{(q,\ell)-}_j \right)^2}.
\label{eq:topsis.distance.neg}
\end{equation}

\textbf{(5) Relative closeness.} The score of model $i$ is computed as
\begin{equation}
s_i^{(s,q,\ell)} = \frac{d_i^{(q,\ell)-}}{d_i^{(q,\ell)+} + d_i^{(q,\ell)-}}.
\label{eq:topsis.score}
\end{equation}

\textbf{(6) Ranking.} Models are ranked in descending order of $s_i^{(s,q,\ell)}$, where smaller ranks indicate better performance.

The \textit{VlseKriterijumska Optimizacija I Kompromisno Resenje (VIKOR)} 
is a ranking scheme $s \in \mathcal{K}$ that selects a compromise solution by 
maximizing group utility and minimizing individual regret. Given a performance matrix $P^{(q,\ell)} \in \mathbb{R}^{m_\ell \times n_{q,\ell}}$ 
and a vector of criteria weights $w^{(q,\ell)} \in \mathbb{R}^{n_{q,\ell}}$, 
the method proceeds as follows:

\textbf{(1) Best and worst values.} For each criterion $j \in \mathcal{D}^{(q,\ell)}$, 
define the best and worst values as
\begin{equation}
v^{(q,\ell)*}_j =
\begin{cases}
\max\limits_{i \in \mathcal{M}^{(\ell)}} P^{(q,\ell)}_{i,j}, & \text{if } j \text{ is to be maximized}, \\
\min\limits_{i \in \mathcal{M}^{(\ell)}} P^{(q,\ell)}_{i,j}, & \text{if } j \text{ is to be minimized},
\end{cases}
\label{eq:vikor.best.value}
\end{equation}

\begin{equation}
v^{(q,\ell)-}_j =
\begin{cases}
\min\limits_{i \in \mathcal{M}^{(\ell)}} P^{(q,\ell)}_{i,j}, & \text{if } j \text{ is to be maximized}, \\
\max\limits_{i \in \mathcal{M}^{(\ell)}} P^{(q,\ell)}_{i,j}, & \text{if } j \text{ is to be minimized}.
\end{cases}
\label{eq:vikor.worst.value}
\end{equation}

\textbf{(2) Normalization.} The normalized values are computed as
\begin{equation}
\hat{P}^{(q,\ell)}_{i,j} = 
\frac{v^{(q,\ell)*}_j - P^{(q,\ell)}_{i,j}}{v^{(q,\ell)*}_j - v^{(q,\ell)-}_j}.
\label{eq:vikor.normalization}
\end{equation}

\textbf{(3) Utility and regret measures.} For each model $i \in \mathcal{M}^{(\ell)}$, define
\begin{equation}
S_i^{(q,\ell)} = \sum_{j \in \mathcal{D}^{(q,\ell)}} w^{(q,\ell)}_j \, \hat{P}^{(q,\ell)}_{i,j},
\label{eq:vikor.scores.s}
\end{equation}

\begin{equation}
R_i^{(q,\ell)} = \max_{j \in \mathcal{D}^{(q,\ell)}} \left( w^{(q,\ell)}_j \, \hat{P}^{(q,\ell)}_{i,j} \right).
\label{eq:vikor.scores.r}
\end{equation}

\textbf{(4) Aggregation.} Let
\[
S^{(q,\ell)*} = \min_{i \in \mathcal{M}^{(\ell)}} S_i^{(q,\ell)}, \quad
S^{(q,\ell)-} = \max_{i \in \mathcal{M}^{(\ell)}} S_i^{(q,\ell)},
\]
\[
R^{(q,\ell)*} = \min_{i \in \mathcal{M}^{(\ell)}} R_i^{(q,\ell)}, \quad
R^{(q,\ell)-} = \max_{i \in \mathcal{M}^{(\ell)}} R_i^{(q,\ell)}.
\]

The final score is computed as
\begin{equation}
s_i^{(s,q,\ell)} = 
v \, \frac{S_i^{(q,\ell)} - S^{(q,\ell)*}}{S^{(q,\ell)-} - S^{(q,\ell)*}} 
+ (1 - v) \, \frac{R_i^{(q,\ell)} - R^{(q,\ell)*}}{R^{(q,\ell)-} - R^{(q,\ell)*}},
\label{eq:vikor.scores.q}
\end{equation}
where $v \in [0,1]$ is a parameter reflecting the weight of group utility.

\textbf{(5) Ranking.} Models are ranked in ascending order of $s_i^{(s,q,\ell)}$, where smaller ranks indicate better performance.

The \textit{Preference Ranking Organization Method for Enrichment Evaluations (PROMETHEE II)} 
is a ranking scheme $s \in \mathcal{K}$ based on pairwise outranking relations, 
producing a complete ranking of alternatives. Given a performance matrix $P^{(q,\ell)} \in \mathbb{R}^{m_\ell \times n_{q,\ell}}$ 
and criteria weights $w^{(q,\ell)} \in \mathbb{R}^{n_{q,\ell}}$, 
the method additionally specifies a preference function 
$q^{(q,\ell)}_j(\cdot)$ for each criterion $j \in \mathcal{D}^{(q,\ell)}$, 
which maps the difference in performance between two models to a preference degree.

\textbf{(1) Pairwise preference.} For two models $i, z \in \mathcal{M}^{(\ell)}$, 
the preference of $i$ over $z$ for criterion $j$ is defined as
\begin{equation}
\operatorname{pref}^{(q,\ell)}_j(i,z) =
\begin{cases}
q^{(q,\ell)}_j\!\left(P^{(q,\ell)}_{i,j} - P^{(q,\ell)}_{z,j}\right), & \text{if } j \text{ is to be maximized}, \\
q^{(q,\ell)}_j\!\left(P^{(q,\ell)}_{z,j} - P^{(q,\ell)}_{i,j}\right), & \text{if } j \text{ is to be minimized}.
\end{cases}
\label{eq:promethee.preference}
\end{equation}

\textbf{(2) Aggregated preference index.} The overall preference of $i$ over $z$ is
\begin{equation}
\pi^{(q,\ell)}(i,z) = 
\sum_{j \in \mathcal{D}^{(q,\ell)}} w^{(q,\ell)}_j \, \operatorname{pref}^{(q,\ell)}_j(i,z).
\label{eq:promethee.sum}
\end{equation}

\textbf{(3) Preference flows.} The positive and negative preference flows for model $i$ are
\begin{equation}
\phi_i^{(q,\ell)+} = 
\frac{1}{m_\ell - 1} \sum_{\substack{z \in \mathcal{M}^{(\ell)} \\ z \neq i}} 
\pi^{(q,\ell)}(i,z),
\label{eq:promethee.positive.flow}
\end{equation}

\begin{equation}
\phi_i^{(q,\ell)-} = 
\frac{1}{m_\ell - 1} \sum_{\substack{z \in \mathcal{M}^{(\ell)} \\ z \neq i}} 
\pi^{(q,\ell)}(z,i).
\label{eq:promethee.negative.flow}
\end{equation}

\textbf{(4) Net flow and ranking.} The final score is given by
\begin{equation}
s_i^{(s,q,\ell)} = \phi_i^{(q,\ell)+} - \phi_i^{(q,\ell)-}.
\label{eq:promethee.net.flow}
\end{equation}

Models are ranked in descending order of $s_i^{(s,q,\ell)}$, i.e.,
\[
R^{(s)}\!\left(P^{(q,\ell)}\right)_i = \operatorname{rank}\!\left(s_i^{(s,q,\ell)}\right),
\]
where smaller ranks indicate better performance.

\section{Selected MCDM Weighting Methods}
\label{appendix:weightingMethods}

Each weighting function in the set $G = \{g^{(1)}, \ldots, g^{(v)}\}$ 
maps a performance matrix to a vector of criteria weights. 
Formally, for task $\mathcal{T}_q$ and language $\ell$,
\[
g^{(t)}: \mathbb{R}^{m_\ell \times n_{q,\ell}} \rightarrow \mathbb{R}^{n_{q,\ell}}, \quad t \in \{1,\ldots,v\},
\]
and
\[
w^{(q,\ell)} = g^{(t)}\!\left(P^{(q,\ell)}\right) = \left(w^{(q,\ell)}_1, \ldots, w^{(q,\ell)}_{n_{q,\ell}}\right).
\]

A simple weighting strategy is to assign equal importance to all criteria, i.e.,
\begin{equation}
w^{(q,\ell)}_j = \frac{1}{n_{q,\ell}}, \quad j \in \mathcal{D}^{(q,\ell)}.
\label{eq:equalWeights}
\end{equation}

The \textit{Method based on the Removal Effects of Criteria (MEREC)} 
is a weighting scheme that assigns higher importance to criteria whose removal 
induces larger changes in model performance. The method assumes strictly positive 
values in the performance matrix.

Given a performance matrix $P^{(q,\ell)} \in \mathbb{R}^{m_\ell \times n_{q,\ell}}$, 
the method proceeds as follows:

\textbf{(1) Normalization.} The normalized values are computed as
\begin{equation}
\hat{P}^{(q,\ell)}_{i,j} =
\begin{cases}
\frac{\min\limits_{z \in \mathcal{M}^{(\ell)}} P^{(q,\ell)}_{z,j}}{P^{(q,\ell)}_{i,j}}, & \text{if } j \text{ is to be maximized}, \\
\frac{P^{(q,\ell)}_{i,j}}{\max\limits_{z \in \mathcal{M}^{(\ell)}} P^{(q,\ell)}_{z,j}}, & \text{if } j \text{ is to be minimized}.
\end{cases}
\label{eq:merec.normalization}
\end{equation}

\textbf{(2) Full-criteria performance.} Using equal weights across all criteria,
\begin{equation}
S_i^{(q,\ell)} = 
\ln\!\left(1 + \frac{1}{n_{q,\ell}} 
\sum_{j \in \mathcal{D}^{(q,\ell)}} \left| \ln\!\left(\hat{P}^{(q,\ell)}_{i,j}\right) \right| \right).
\label{eq:merec.alternative.first}
\end{equation}

\textbf{(3) Performance with criterion removal.} For each criterion $j \in \mathcal{D}^{(q,\ell)}$,
\begin{equation}
S_{i,j}^{(q,\ell)} = 
\ln\!\left(1 + \frac{1}{n_{q,\ell}} 
\sum_{\substack{z \in \mathcal{D}^{(q,\ell)} \\ z \neq j}} 
\left| \ln\!\left(\hat{P}^{(q,\ell)}_{i,z}\right) \right| \right).
\label{eq:merec.alternative.second}
\end{equation}

\textbf{(4) Criterion effect.} The impact of removing criterion $j$ is
\begin{equation}
E_j^{(q,\ell)} = 
\sum_{i \in \mathcal{M}^{(\ell)}} 
\left| S_{i,j}^{(q,\ell)} - S_i^{(q,\ell)} \right|.
\label{eq:merec.criteria}
\end{equation}

\textbf{(5) Weight computation.} The final weights are obtained by normalization:
\begin{equation}
w^{(q,\ell)}_j = 
\frac{E_j^{(q,\ell)}}{\sum_{z \in \mathcal{D}^{(q,\ell)}} E_z^{(q,\ell)}}.
\label{eq:merec.sum}
\end{equation}

The \textit{CRiteria Importance Through Inter-criteria Correlation (CRITIC)} 
method is a weighting scheme that assigns higher importance to criteria with 
greater variability across models and stronger conflict (i.e., lower correlation) 
with other criteria.

Given a performance matrix $P^{(q,\ell)} \in \mathbb{R}^{m_\ell \times n_{q,\ell}}$, 
the method proceeds as follows:

\textbf{(1) Normalization.} The normalized values are computed as
\begin{equation}
\hat{P}^{(q,\ell)}_{i,j} =
\begin{cases}
\frac{P^{(q,\ell)}_{i,j} - \min\limits_{z \in \mathcal{M}^{(\ell)}} P^{(q,\ell)}_{z,j}}
{\max\limits_{z \in \mathcal{M}^{(\ell)}} P^{(q,\ell)}_{z,j} - \min\limits_{z \in \mathcal{M}^{(\ell)}} P^{(q,\ell)}_{z,j}}, 
& \text{if } j \text{ is to be maximized}, \\
\frac{\max\limits_{z \in \mathcal{M}^{(\ell)}} P^{(q,\ell)}_{z,j} - P^{(q,\ell)}_{i,j}}
{\max\limits_{z \in \mathcal{M}^{(\ell)}} P^{(q,\ell)}_{z,j} - \min\limits_{z \in \mathcal{M}^{(\ell)}} P^{(q,\ell)}_{z,j}}, 
& \text{if } j \text{ is to be minimized}.
\end{cases}
\label{eq:critic.normalization}
\end{equation}

\textbf{(2) Dispersion.} For each criterion $j \in \mathcal{D}^{(q,\ell)}$, 
compute the standard deviation
\begin{equation}
S_j^{(q,\ell)} = 
\sqrt{
\frac{1}{m_\ell - 1}
\sum_{i \in \mathcal{M}^{(\ell)}}
\left( \hat{P}^{(q,\ell)}_{i,j} - \bar{P}^{(q,\ell)}_j \right)^2
},
\label{eq:critic.dispersion}
\end{equation}
where
\[
\bar{P}^{(q,\ell)}_j = \frac{1}{m_\ell} \sum_{i \in \mathcal{M}^{(\ell)}} \hat{P}^{(q,\ell)}_{i,j}.
\]

\textbf{(3) Conflict.} The conflict degree of criterion $j$ is defined as
\begin{equation}
R_j^{(q,\ell)} = 
\sum_{\substack{z \in \mathcal{D}^{(q,\ell)} \\ z \neq j}} 
\left( 1 - \rho^{(q,\ell)}_{j,z} \right),
\label{eq:critic.conflict}
\end{equation}
where $\rho^{(q,\ell)}_{j,z}$ is the Pearson correlation coefficient between 
criteria $j$ and $z$.

\textbf{(4) Information measure.} The information content of criterion $j$ is
\begin{equation}
C_j^{(q,\ell)} = S_j^{(q,\ell)} \, R_j^{(q,\ell)}.
\end{equation}

\textbf{(5) Weight computation.} The final weights are obtained as
\begin{equation}
w^{(q,\ell)}_j = 
\frac{C_j^{(q,\ell)}}{\sum_{z \in \mathcal{D}^{(q,\ell)}} C_z^{(q,\ell)}}.
\label{eq:critic.sum}
\end{equation}

\section{MCDM Aggregation and Weighting Portfolio Diversity}
\label{app:portfolio}
The selected portfolio of MCDM methods, i.e., WSM, TOPSIS~\cite{chen1992fuzzy}, VIKOR~\cite{opricovic1998multicriteria}, and PROMETHEE II~\cite{brans1982ingenierie}, was chosen to represent the principal methodological families in MCDM, ensuring both conceptual diversity and complementary decision logic. 

WSM represents the class of \emph{value-based (multi-attribute utility) methods}, in which alternatives are evaluated through an explicit aggregation of weighted criteria into a single scalar utility. These methods are widely used due to their simplicity and interpretability. Closely related methods in this family include the \textit{Weighted Product Model (WPM)} and \textit{Multi-Attribute Utility Theory (MAUT)}, which differ mainly in the form of aggregation but follow the same underlying utility-based principle.

\textit{TOPSIS} belongs to the family of \emph{distance-based (reference point) methods}, which rank alternatives according to their geometric proximity to ideal and anti-ideal solutions in the criteria space~\cite{chen1992fuzzy}. These methods emphasize relative performance by considering both best and worst attainable outcomes. Similar approaches include \textit{EDAS}, \textit{CODAS}, \textit{Grey Relational Analysis (GRA)}, which adopt alternative distance or similarity measures but retain the same reference-based logic.

\textit{VIKOR} represents \emph{compromise-based methods}, which aim to identify solutions that balance collective utility and individual regret~\cite{opricovic1998multicriteria}. These methods explicitly model decision-making as a trade-off between conflicting criteria, producing rankings that reflect a compromise solution. Related methods include \textit{COPRAS}, \textit{CoCoSo}, which similarly combine multiple aggregation principles to capture compromise behavior.

Finally, \textit{PROMETHEE II} belongs to the \emph{outranking methods} family, which relies on pairwise comparisons and preference relations rather than global aggregation~\cite{brans1982ingenierie}. These methods allow for flexible modeling of preference structures through criterion-specific preference functions. The most closely related alternatives are the \textit{ELECTRE} family of methods, which share the same outranking philosophy.

Together, these four methods span the major paradigms of MCDM. Since many other techniques can be interpreted as variants within these families, the selected portfolio provides a representative and methodologically diverse basis for comparative evaluation without introducing redundancy.

\section{Sensitivity Analysis of Correlation Threshold $\tau$}
\label{sec:sensitivty_correlation}
This section presents a sensitivity analysis of the correlation threshold used to split datasets and construct benchmark compositions. For each task–language pair, we vary the threshold $\tau \in \{0.8, 0.85, 0.9\}$ and generate dataset clusters accordingly. The clustering solutions obtained for different thresholds are compared using the Adjusted Mutual Information (AMI) score, where values close to 0 indicate dissimilar clusters and 1 indicates identical partitions. For each threshold, we compute model rankings using all combinations of ranking schemes and weighting methods. To quantify the impact of the threshold choice, we compute the Spearman correlation between rankings obtained from different thresholds while keeping the ranking scheme and weighting method fixed. We then report the mean and standard deviation of these correlations across all ranking scheme and weighting method combinations.

Next, for each learning task, we present the results separately for each language. Each figure contains five panels, corresponding to the different languages. Within each panel, three points represent the pairwise comparisons between the tested correlation thresholds: $(0.9, 0.85)$, $(0.9, 0.8)$, and $(0.85, 0.8)$. The x-axis shows the AMI between the corresponding clustering outputs, while the y-axis reports the mean Spearman rank correlation between the resulting rankings. Error bars indicate the standard deviation across all ranking schemes and weighting methods for the given threshold pair.  Points within each panel are connected to highlight how changes in threshold parameters affect both clustering agreement and ranking stability.

\begin{figure*}[ht]
  \begin{center}
    \centerline{\includegraphics[width=\textwidth]{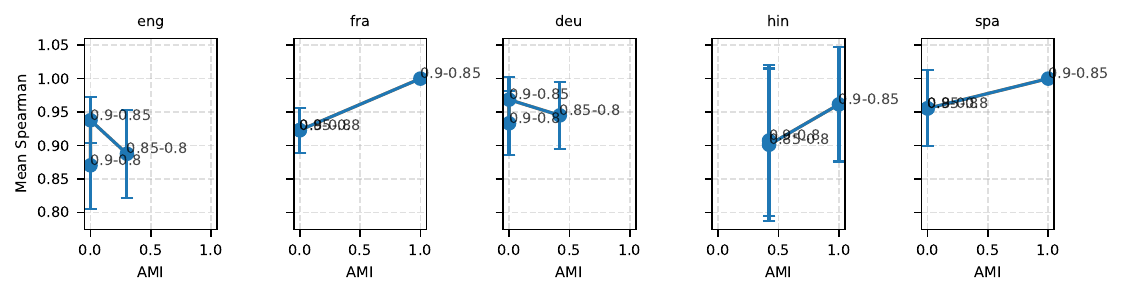}}
    \caption{
      Relationship between clustering similarity and ranking consistency across languages on the classification task.
    }
    \label{fig:clustering_sensitivity_classification}
  \end{center}
\end{figure*}

Figure~\ref{fig:clustering_sensitivity_classification} shows that despite substantial differences between clustering outputs, the resulting rankings remain highly consistent across all languages. In particular, even when the AMI is close to zero, indicating that the clusters are essentially unrelated, the mean Spearman rank correlation remains high (typically above 0.85 and often exceeding 0.9). This pattern holds across languages, with only moderate variation in sensitivity. Namely, while some languages (e.g., Hindi and Spanish) show a clearer increase in Spearman correlation with higher AMI, others (e.g., English and German) maintain strong ranking agreement regardless of clustering similarity. These results demonstrate that the ranking procedure is robust to threshold variations in clustering, suggesting that the underlying signal driving the rankings is largely independent of the specific clustering configuration.

\begin{figure*}[ht]
  \begin{center}
    \centerline{\includegraphics[width=\textwidth]{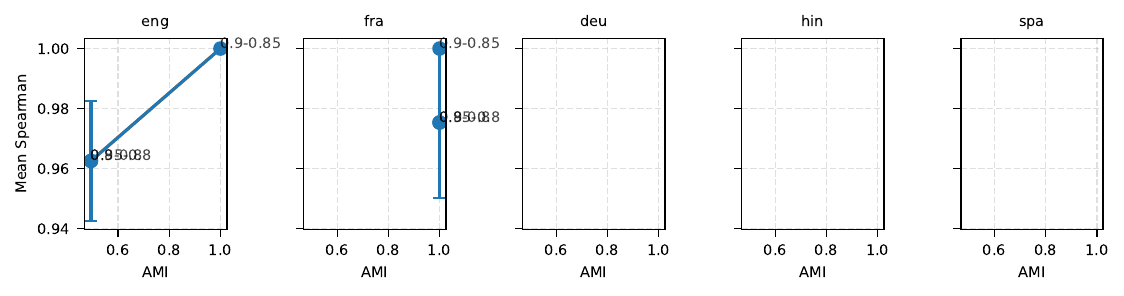}}
    \caption{
      Relationship between clustering similarity and ranking consistency across languages on the clustering task.
    }
    \label{fig:clustering_sensitivity_clustering}
  \end{center}
\end{figure*}

Figure~\ref{fig:clustering_sensitivity_clustering} shows that across the observed languages, the AMI values are predominantly in the moderate-to-high range ($\geq$ 0.6), indicating that the resulting clusters are already quite similar. Correspondingly, the mean Spearman rank correlation is uniformly very high, typically between 0.95 and 1.0. In cases such as English, a mild positive relationship between AMI and Spearman correlation can still be observed, with higher clustering similarity yielding near-perfect ranking agreement. For French, the clustering outputs are effectively identical, resulting in consistently near-perfect Spearman correlations, which indicates that the rankings are entirely stable across threshold variations. For languages where only a single dataset is available (German, Hindi, and Spanish), this analysis is not applicable, and thus no comparison is possible.

\begin{figure*}[ht]
  \begin{center}
    \centerline{\includegraphics[width=\textwidth]{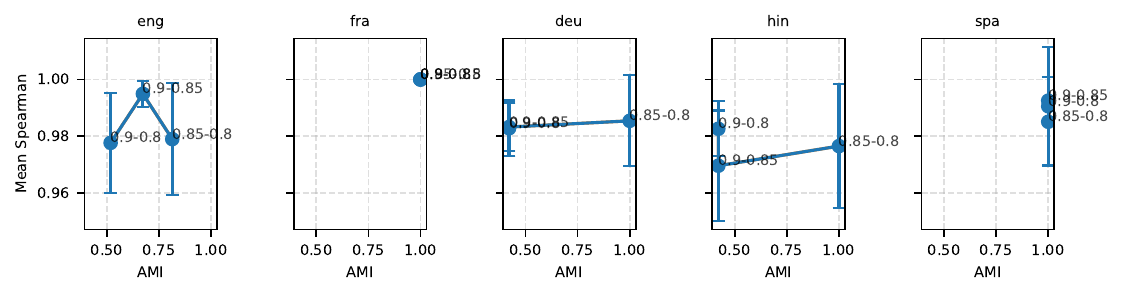}}
    \caption{
      Relationship between clustering similarity and ranking consistency across languages on the retrieval task.
    }
    \label{fig:clustering_sensitivity_retrieval}
  \end{center}
\end{figure*}

Figure~\ref{fig:clustering_sensitivity_retrieval} shows that across all languages, the rankings remain highly consistent despite moderate variation in clustering similarity, with AMI values ranging roughly from 0.5 to 1.0 and corresponding Spearman correlations consistently remaining very high (typically above 0.97). In English, minor non-monotonic fluctuations are observed, but ranking agreement remains near-perfect throughout. French exhibits identical clusters and consequently perfect ranking agreement. German and Spanish show slight positive trends, where increased clustering similarity leads to marginal improvements in Spearman correlation, though the effect is small due to already high baseline correlation. Hindi displays a somewhat clearer dependence on AMI, with modest gains in ranking agreement as clustering similarity increases. Overall, these results indicate that in the retrieval setting, rankings are exceptionally robust, with clustering differences having only a minimal impact.

\begin{figure*}[ht]
  \begin{center}
    \centerline{\includegraphics[width=\textwidth]{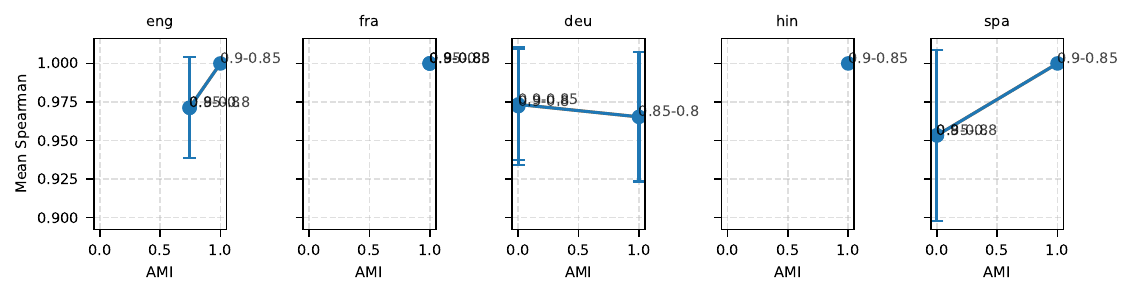}}
    \caption{
      Relationship between clustering similarity and ranking consistency across languages on the STS task.
    }
    \label{fig:clustering_sensitivity_sts}
  \end{center}
\end{figure*}

Figure~\ref{fig:clustering_sensitivity_sts} shows that across all languages, the rankings in the STS task remain highly consistent despite variation in clustering similarity, with Spearman correlations consistently high (generally above 0.95 and often approaching 1.0). English and Spanish show a clear positive relationship between AMI and Spearman correlation, where increased clustering similarity leads to near-perfect ranking agreement. French and Hindi exhibit identical clustering outputs, resulting in fully stable rankings. German presents a slightly different pattern, with high but mildly decreasing Spearman correlation as AMI increases, though the variation remains small and within a consistently high range. Overall, these results indicate that, similar to previous tasks, rankings in the STS task are highly robust to clustering differences.

\begin{figure*}[ht]
  \begin{center}
    \centerline{\includegraphics[width=\textwidth]{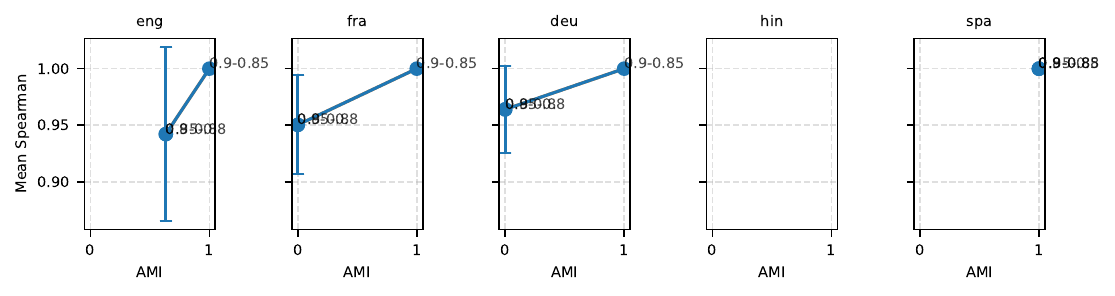}}
    \caption{
      Relationship between clustering similarity and ranking consistency across languages on the pair classification task.
    }
    \label{fig:clustering_sensitivity_pairclassification}
  \end{center}
\end{figure*}

Figure~\ref{fig:clustering_sensitivity_pairclassification} shows that across languages, the pair classification results demonstrate that rankings remain highly stable despite substantial differences in clustering. Even when the AMI is close to zero, indicating largely dissimilar clusters, the mean Spearman correlation remains high (approximately 0.94–0.97), confirming strong robustness of the ranking procedure. As clustering similarity increases, Spearman correlation consistently approaches perfect agreement (1.0), with only modest gains due to the already high baseline. English, French, and German all exhibit this pattern, with slightly increased variability at low AMI, while Spanish shows perfect agreement in the absence of clustering variation and Hindi has a single datasets on the task, making the analysis inapplicable. Overall, these results reinforce that even in the pair classification task, ranking outcomes are largely insensitive to clustering differences.

\begin{figure*}[ht]
  \begin{center}
    \centerline{\includegraphics[width=\textwidth]{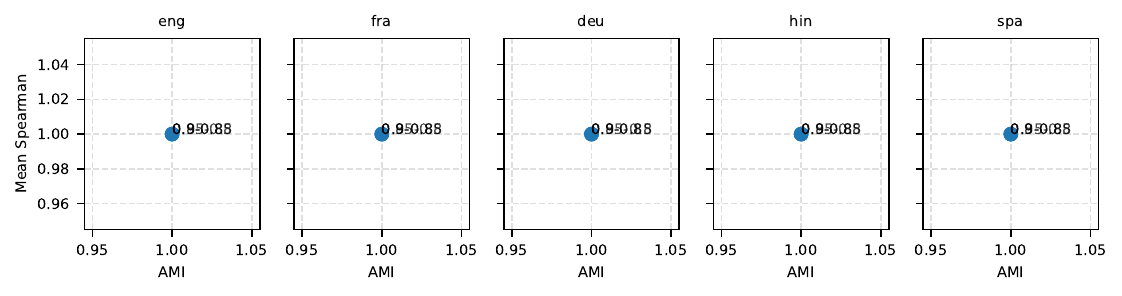}}
    \caption{
      Relationship between clustering similarity and ranking consistency across languages on the bitext mining task.
    }
    \label{fig:clustering_sensitivity_bitextmining}
  \end{center}
\end{figure*}

Figure~\ref{fig:clustering_sensitivity_bitextmining} shows that across all five languages, the bitext mining task exhibits complete stability in both clustering and ranking, with AMI consistently equal to 1 and mean Spearman correlation correspondingly at 1.0. No variation is observed across threshold comparisons, indicating that the clustering outputs are identical and the induced rankings are perfectly preserved. This uniform pattern holds for all languages (English, French, German, Hindi, and Spanish), with overlapping points reflecting the absence of any measurable difference between configurations. Overall, these results demonstrate that in the bitext mining task, the ranking procedure is entirely insensitive to clustering choices.

In the reranking task, the English language shows complete stability in both clustering and ranking, AMI consistently equal to 1 and mean Spearman correlation correspondingly at 1.0. Since the remaining language have only one dataset under this task, the analysis is inapplicable to their case. In the instruction reranking task the conclusions are similar. There are three datasets for the English language and none for the remaining four languages. The English language again shows complete stability in both clustering and ranking, with AMI consistently equal to 1 and mean Spearman correlation also at 1.0. All five languages have a single dataset under the Multiclass Classification task, making this analysis inapplicable for all of them.

\section{Sensitivity Analysis of Top-$\eta$ Sets of Models (Task-Agnostic Within a Language)}
\label{app:sensitivty_top}
This section compares the task-agnostic results for different value of the parameter $\eta \in \{1, 5, 10\}$ across each of the five languages. While the results for $\eta$=10 were already discussed in details in Appendix~\ref{appendix:task_agnostic}, this section compares those results with the results achieved for $\eta$=5 and $\eta$=1. The results are again visualized though dendrograms, referenced in the paragraphs that follow.

\begin{figure*}[ht]

  \begin{center}
    \centerline{\includegraphics[width=\textwidth]{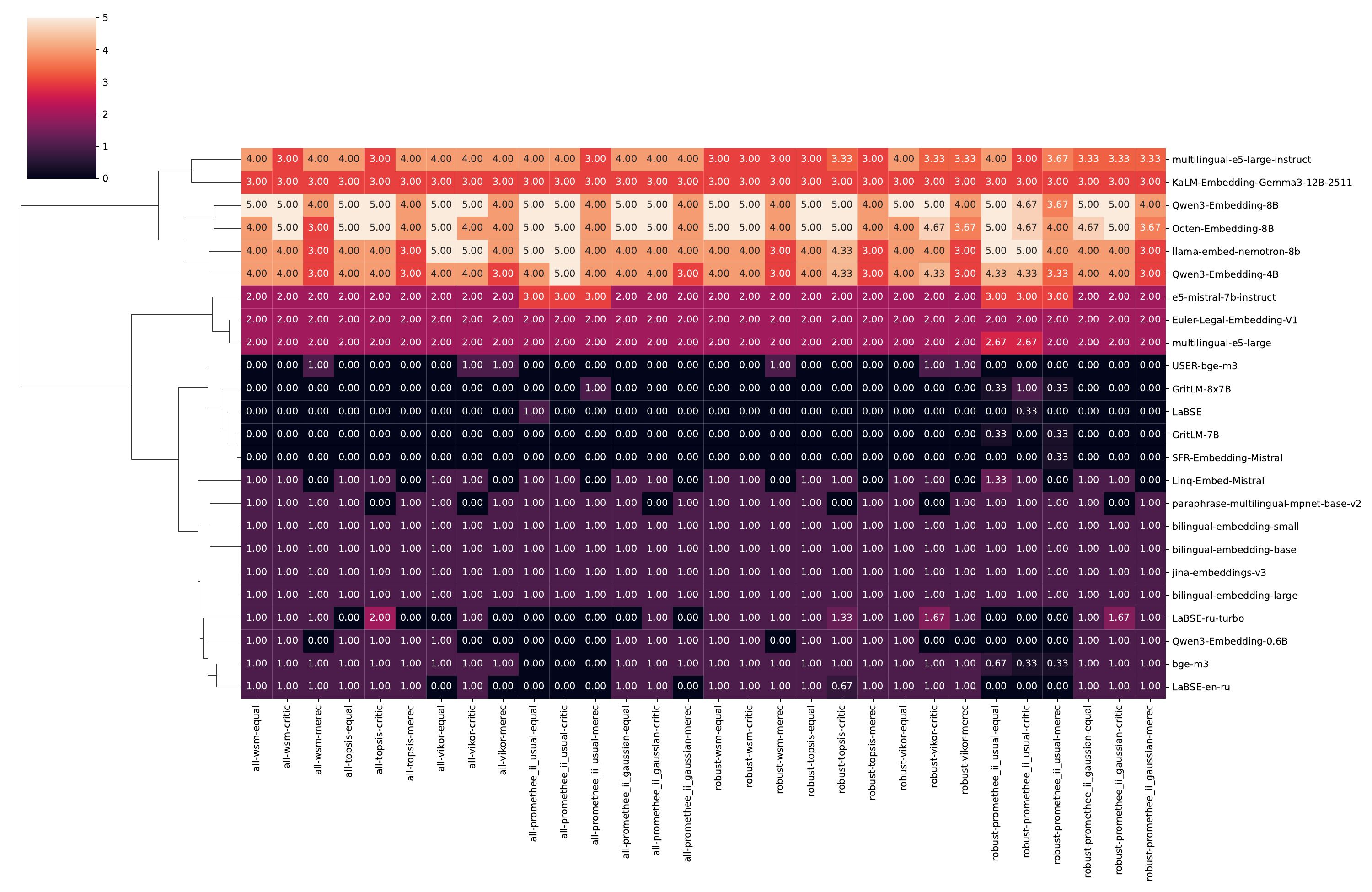}}
    \caption{
      Cross-task consistency by a ranking scheme for French language for $\eta$=5.
    }
    \label{fig:fra_consistency_scheme_top_5}
  \end{center}
\end{figure*}

\begin{figure*}[ht]

  \begin{center}
    \centerline{\includegraphics[width=\textwidth]{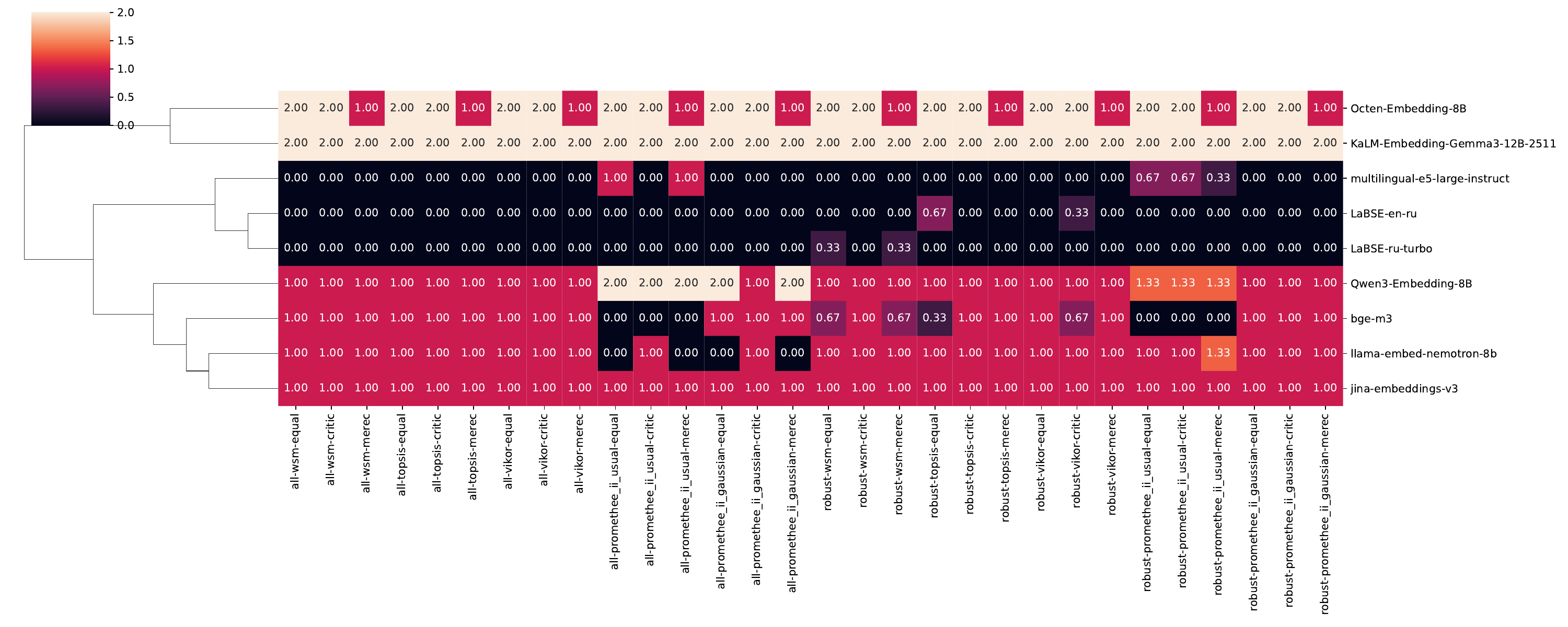}}
    \caption{
      Cross-task consistency by a ranking scheme for French language for $\eta$=1.
    }
    \label{fig:fra_consistency_scheme_top_1}
  \end{center}
\end{figure*}

Across the three figures referring to the French language, i.e., Figure~\ref{fig:fra_consistency_scheme} for $\eta$=10,  Figure~\ref{fig:fra_consistency_scheme_top_5} for $\eta$=5, and Figure~\ref{fig:fra_consistency_scheme_top_1} for $\eta$=1, a consistent set of top-performing models is observed, exhibiting strong stability across the ranking schemes.

\begin{figure*}[ht]

  \begin{center}
    \centerline{\includegraphics[width=\textwidth]{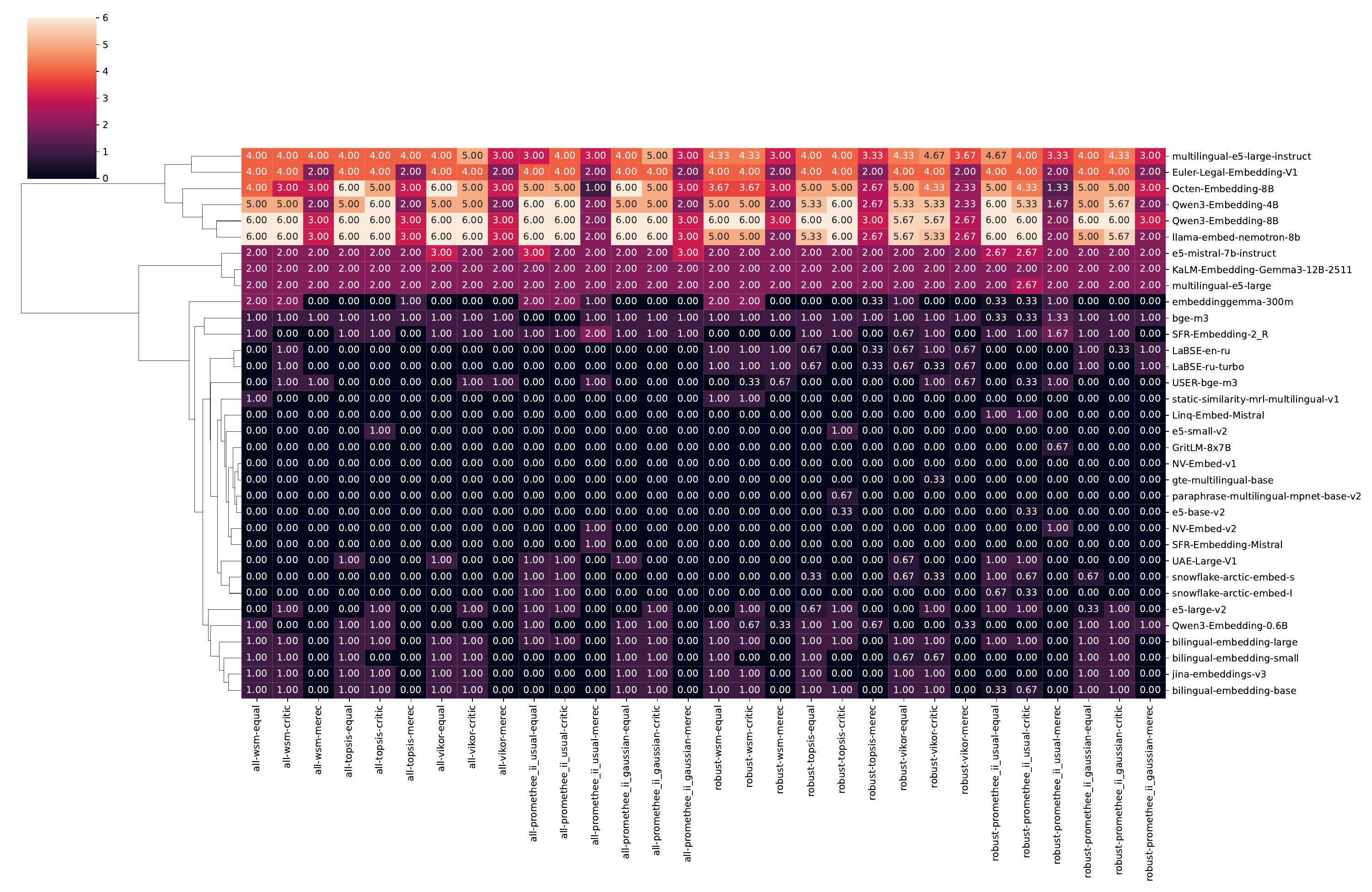}}
    \caption{
      Cross-task consistency by a ranking scheme for English language for $\eta$=5.
    }
    \label{fig:eng_consistency_scheme_top_5}
  \end{center}
\end{figure*}

\begin{figure*}[ht]

  \begin{center}
    \centerline{\includegraphics[width=\textwidth]{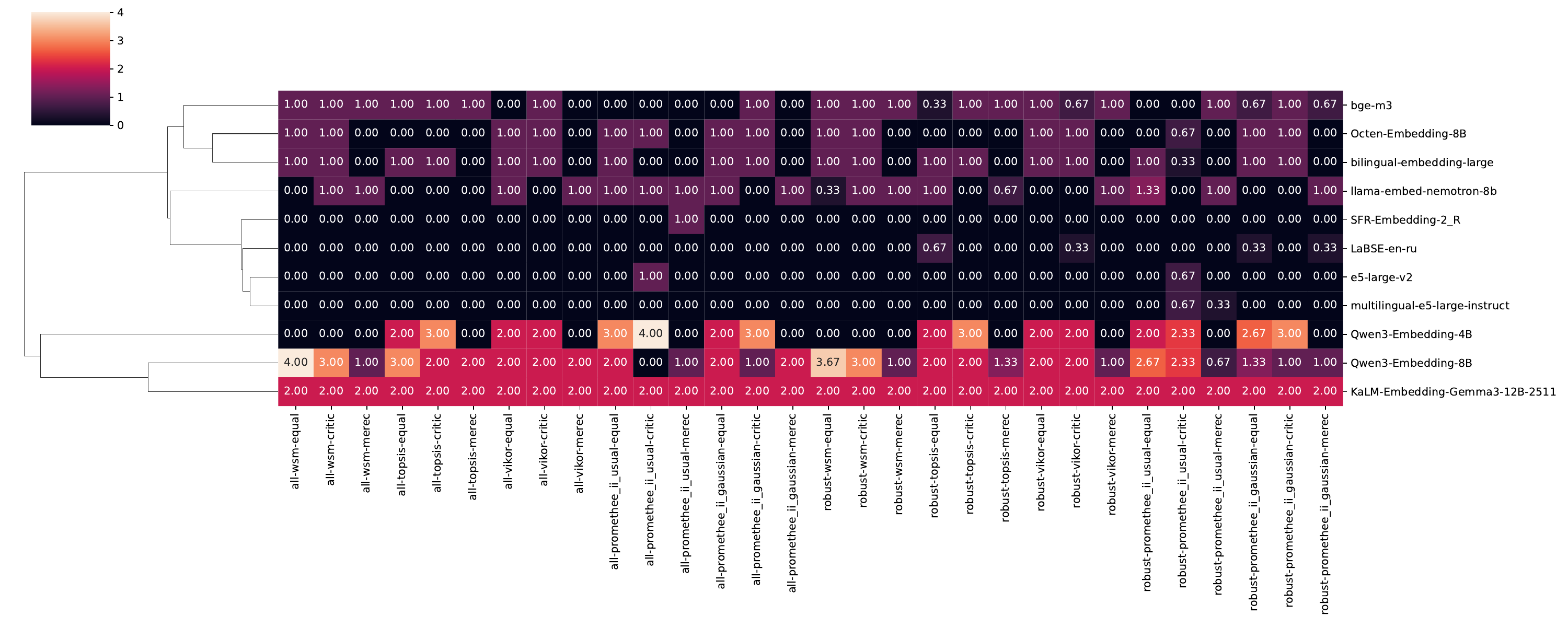}}
    \caption{
      Cross-task consistency by a ranking scheme for English language for $\eta$=1.
    }
    \label{fig:eng_consistency_scheme_top_1}
  \end{center}
\end{figure*}

For the English language, the three figures, i.e., Figure~\ref{fig:eng_consistency_scheme} for $\eta$=10,  Figure~\ref{fig:eng_consistency_scheme_top_5} for $\eta$=5, and Figure~\ref{fig:eng_consistency_scheme_top_1} for $\eta$=1, again show that a consistent pattern emerges regarding the stability and dominance of top-performing LLMs across different ranking schemes. 

\begin{figure*}[ht]

  \begin{center}
    \centerline{\includegraphics[width=\textwidth]{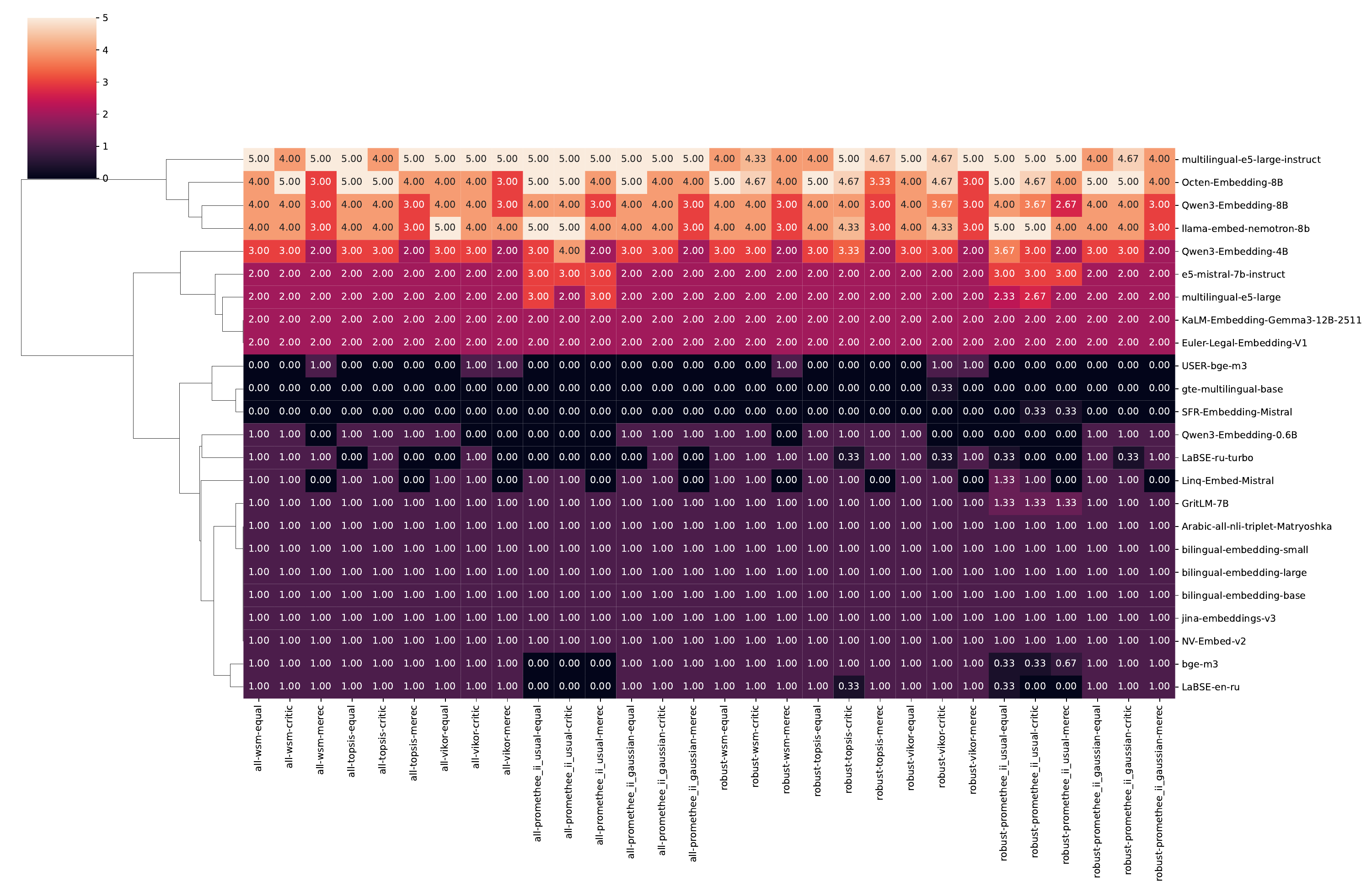}}
    \caption{
      Cross-task consistency by a ranking scheme for German language for $\eta$=5.
    }
    \label{fig:deu_consistency_scheme_top_5}
  \end{center}
\end{figure*}

\begin{figure*}[ht]

  \begin{center}
    \centerline{\includegraphics[width=\textwidth]{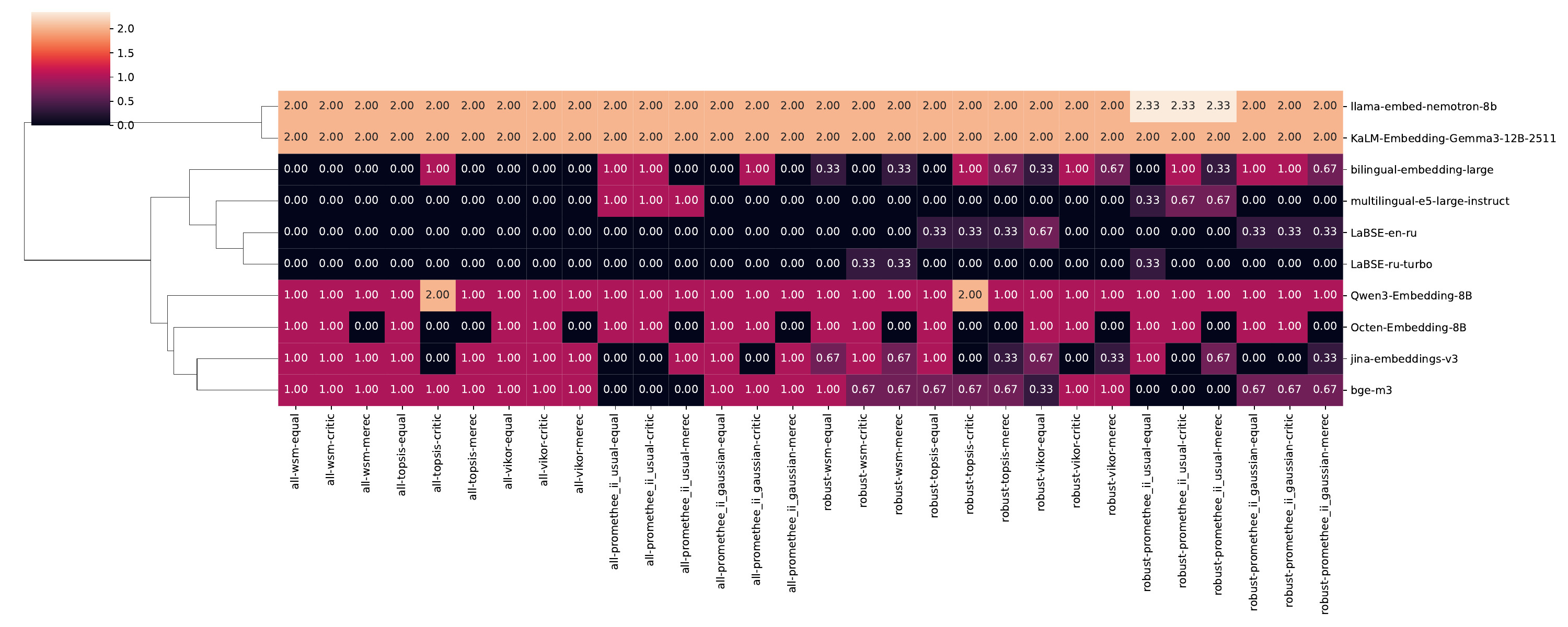}}
    \caption{
      Cross-task consistency by a ranking scheme for German language for $\eta$=1.
    }
    \label{fig:deu_consistency_scheme_top_1}
  \end{center}
\end{figure*}

Similarly, across the three German-language figures, i.e., Figure~\ref{fig:deu_consistency_scheme} for $\eta$=10,  Figure~\ref{fig:deu_consistency_scheme_top_5} for $\eta$=5, and Figure~\ref{fig:deu_consistency_scheme_top_1} for $\eta$=1, a clear and highly consistent set of top-performing models emerges, with strong stability across ranking schemes and only modest variation as the $\eta$ parameter cutoff becomes stricter. 

\begin{figure*}[ht]

  \begin{center}
    \centerline{\includegraphics[width=\textwidth]{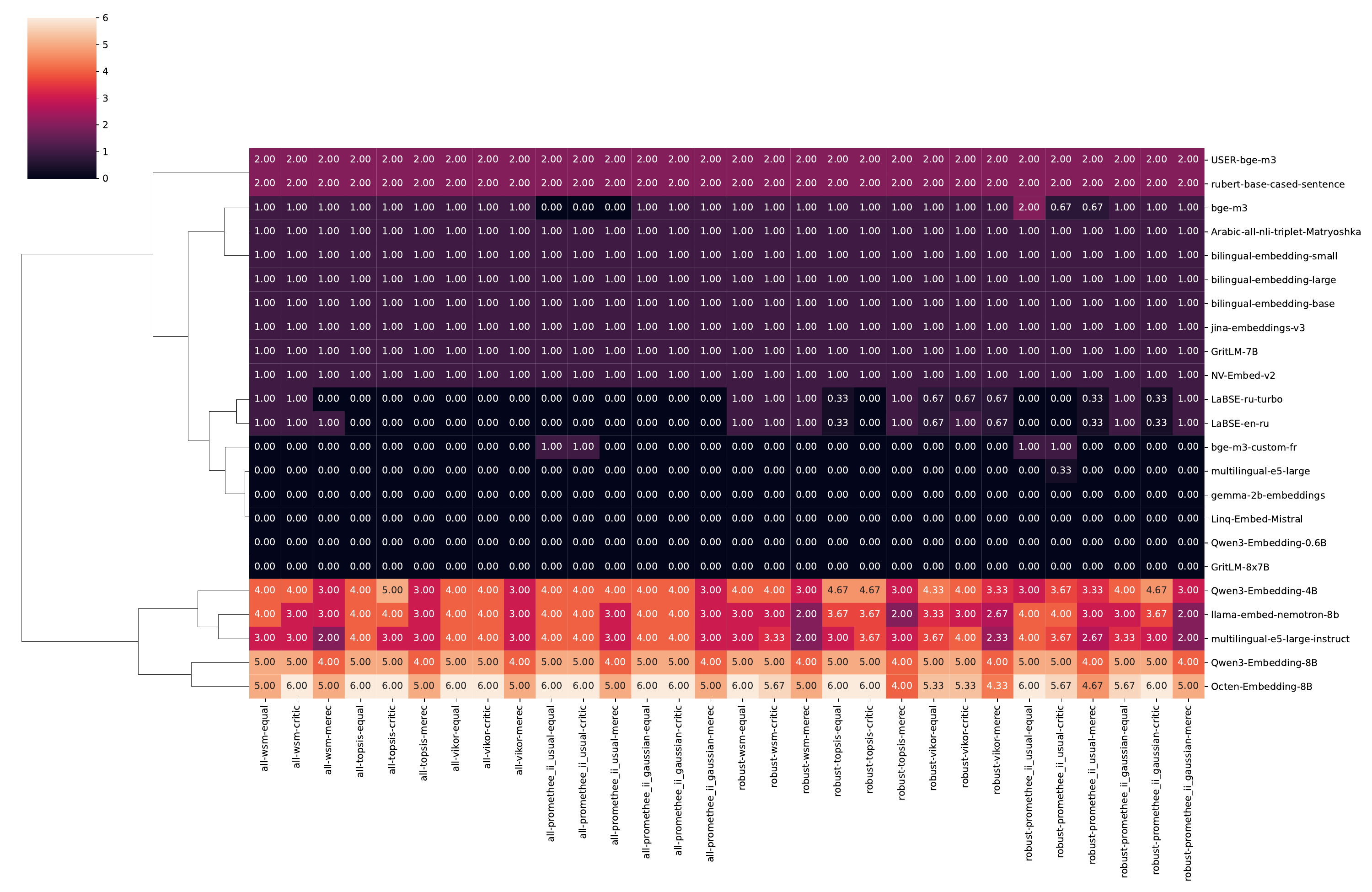}}
    \caption{
      Cross-task consistency by a ranking scheme for Hindi language for $\eta$=5.
    }
    \label{fig:hin_consistency_scheme_top_5}
  \end{center}
\end{figure*}

\begin{figure*}[ht]

  \begin{center}
    \centerline{\includegraphics[width=\textwidth]{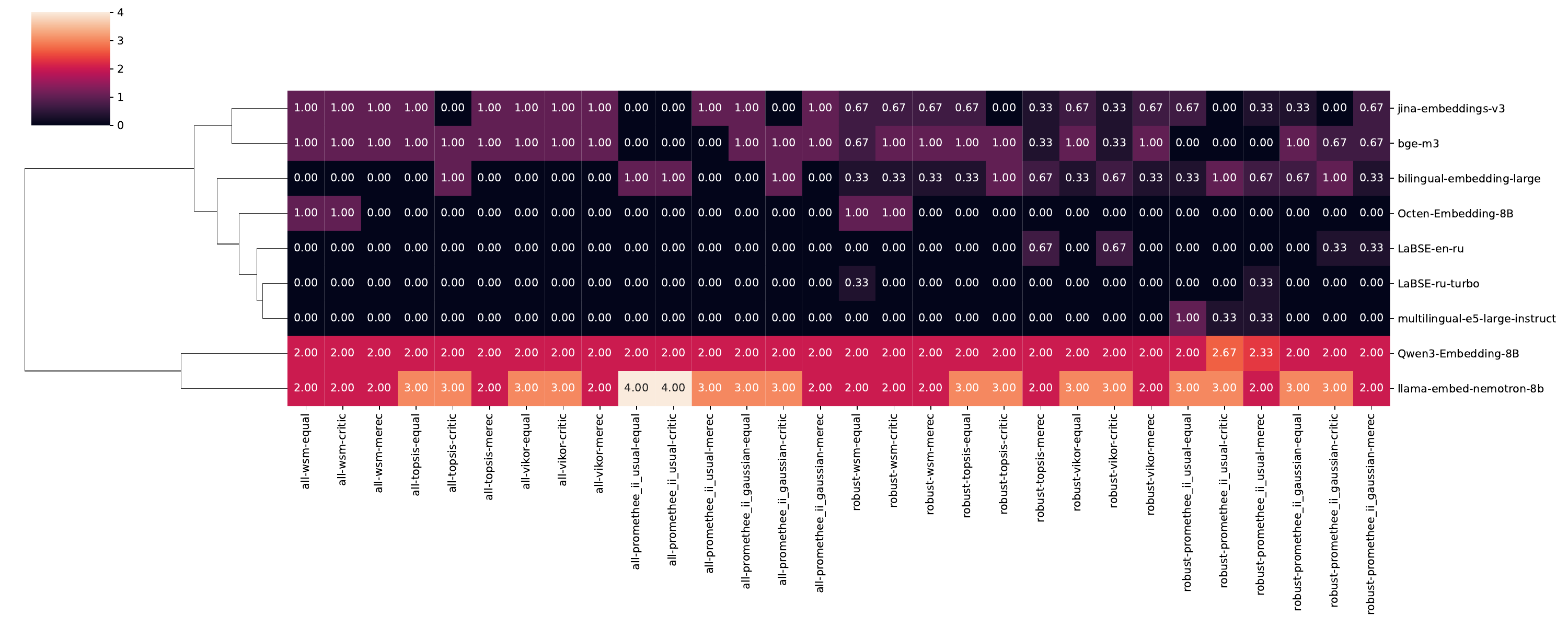}}
    \caption{
      Cross-task consistency by a ranking scheme for Hindi language for $\eta$=1.
    }
    \label{fig:hin_consistency_scheme_top_1}
  \end{center}
\end{figure*}

The three Hindi-language figures, i.e., Figure~\ref{fig:hin_consistency_scheme} for $\eta$=10,  Figure~\ref{fig:hin_consistency_scheme_top_5} for $\eta$=5, and Figure~\ref{fig:hin_consistency_scheme_top_1} for $\eta$=1, show a clear and highly consistent set of top-performing models, with strong stability across ranking schemes and only moderate variation as the $\eta$ parameter cutoff becomes stricter. 

\begin{figure*}[ht]

  \begin{center}
    \centerline{\includegraphics[width=\textwidth]{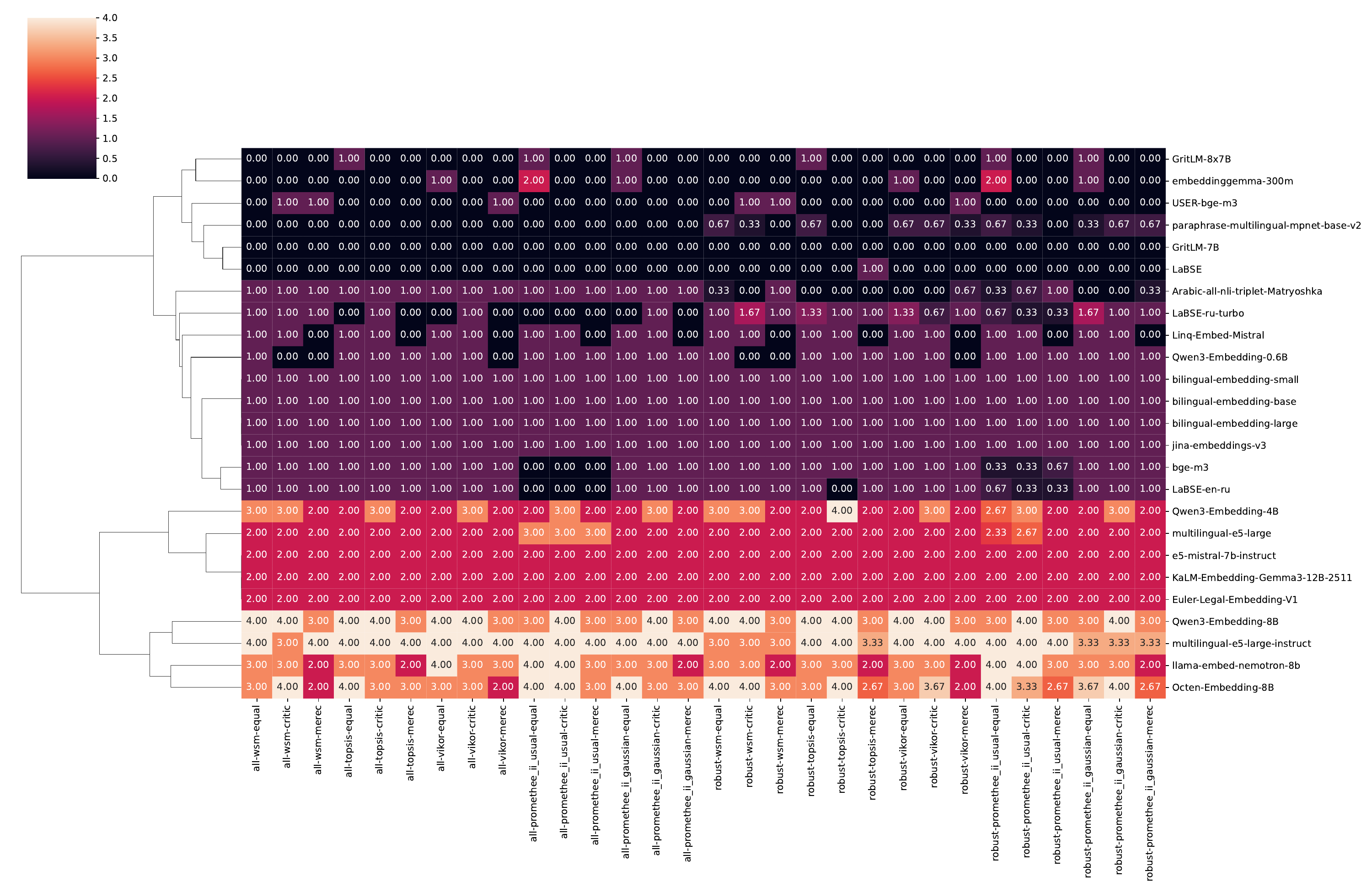}}
    \caption{
      Cross-task consistency by a ranking scheme for Spanish language for $\eta$=5.
    }
    \label{fig:spa_consistency_scheme_top_5}
  \end{center}
\end{figure*}

\begin{figure*}[ht]

  \begin{center}
    \centerline{\includegraphics[width=\textwidth]{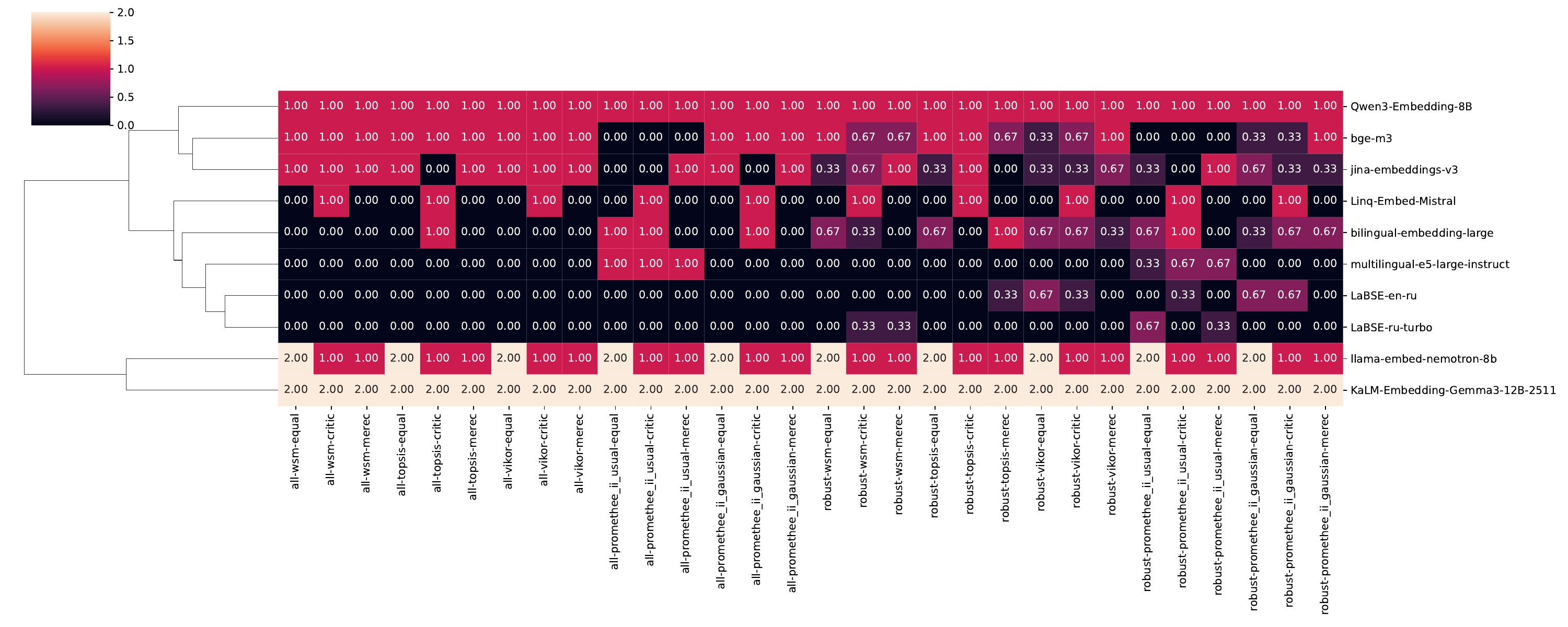}}
    \caption{
      Cross-task consistency by a ranking scheme for Spanish language for $\eta$=1.
    }
    \label{fig:spa_consistency_scheme_top_1}
  \end{center}
\end{figure*}

Finally, the three Spanish-language figures, i.e., Figure~\ref{fig:spa_consistency_scheme} for $\eta$=10,  Figure~\ref{fig:spa_consistency_scheme_top_5} for $\eta$=5, and Figure~\ref{fig:spa_consistency_scheme_top_1} for $\eta$=1, the results again reveal a clear core of high-performing and stable models, although with slightly more variation across ranking schemes compared to some of the previously described languages. 

\section{Comparison with MTEB Rank}
\label{sec:baseline}
We compare our results with the MMTEB v2 leaderboard as a baseline. For each language–task pair, we first obtain baseline rankings based on the reported performance scores. These baseline rankings are then compared with the rankings produced by each combination of ranking scheme, weighting method, and dataset composition using the Spearman rank correlation. To summarize the results, we aggregate the Spearman correlations in two ways. First, for each dataset composition, we aggregate correlations across the 15 combinations of ranking schemes and weighting methods. Second, for each ranking scheme–weighting method pair, we analyze the influence of the three dataset compositions by aggregating the corresponding correlations. This allows us to assess how consistent the obtained rankings are with respect to both the ranking schemes and dataset composition.

\subsection{Aggregation by Dataset Composition}

\begin{figure*}[ht]
  \begin{center}
    \centerline{\includegraphics[width=\textwidth]{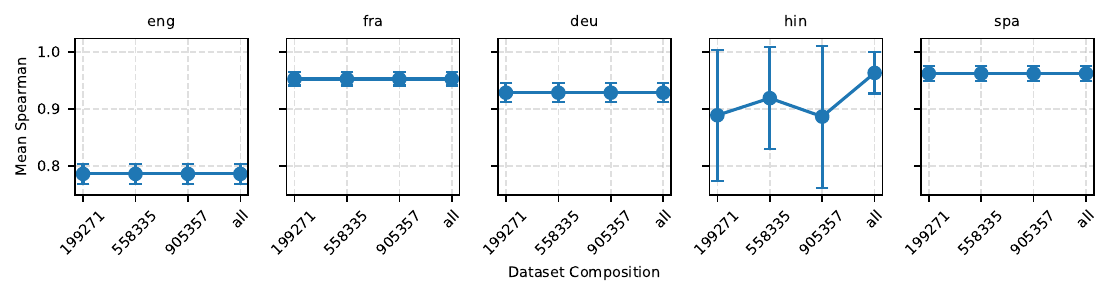}}
    \caption{
      Spearman correlation between the MTEB model rank and the rankings produced by different ranking schemes, averaged by dataset composition, on the classification task.
    }
    \label{fig:baseline_rank_comparison_classification}
  \end{center}
\end{figure*}

Across the five language-specific plots for the classification task presented in Figure~\ref{fig:baseline_rank_comparison_classification}, the average Spearman correlation between the baseline MTEB ranking and the ranking schemes proposed in this paper remains consistently high, but with notable differences in both level and variability. English shows the lowest correlation (around 0.78–0.79) with very small variation across dataset compositions, indicating stable but comparatively weaker alignment with the baseline. French and Spanish exhibit the highest and most stable correlations (around 0.95–0.97), with almost no sensitivity to dataset composition or ranking scheme. German follows a similar pattern but at a slightly lower correlation level (around 0.92–0.93), still with minimal variability. In contrast, Hindi displays the greatest variability, with correlations ranging roughly from around 0.88 to around 0.97, with a large standard deviation, indicating substantial sensitivity to dataset composition and greater uncertainty in ranking consistency. Overall, while most languages (French, Spanish, German) demonstrate both high and stable correlation with the baseline MTEB ranking regardless of dataset composition, English shows consistently lower correlation, and Hindi stands out as the most variable and least stable across dataset compositions.

\begin{figure*}[ht]
  \begin{center}
    \centerline{\includegraphics[width=\textwidth]{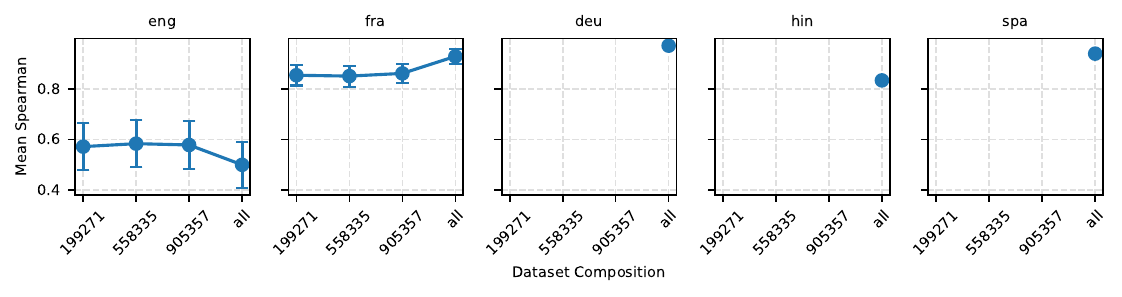}}
    \caption{
      Spearman correlation between the MTEB model rank and the rankings produced by different ranking schemes, averaged by dataset composition, on the clustering task.
    }
    \label{fig:baseline_rank_comparison_clustering}
  \end{center}
\end{figure*}

\begin{figure*}[ht]
  \begin{center}
    \centerline{\includegraphics[width=\textwidth]{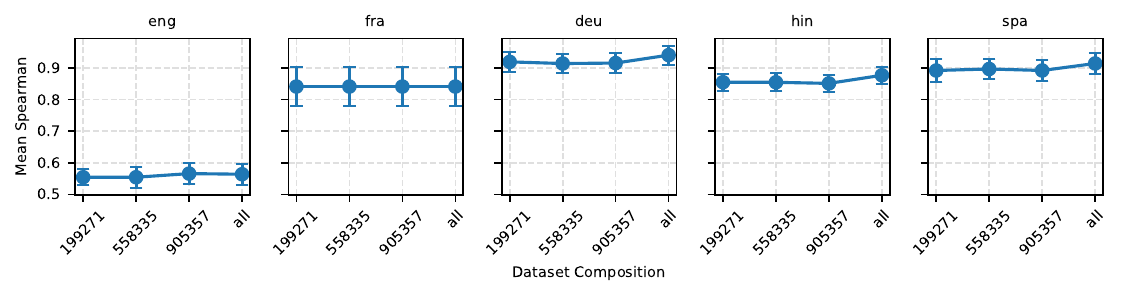}}
    \caption{
      Spearman correlation between the MTEB model rank and the rankings produced by different ranking schemes, averaged by dataset composition, on the retrieval task.
    }
    \label{fig:baseline_rank_comparison_retrieval}
  \end{center}
\end{figure*}

\begin{figure*}[ht]
  \begin{center}
    \centerline{\includegraphics[width=\textwidth]{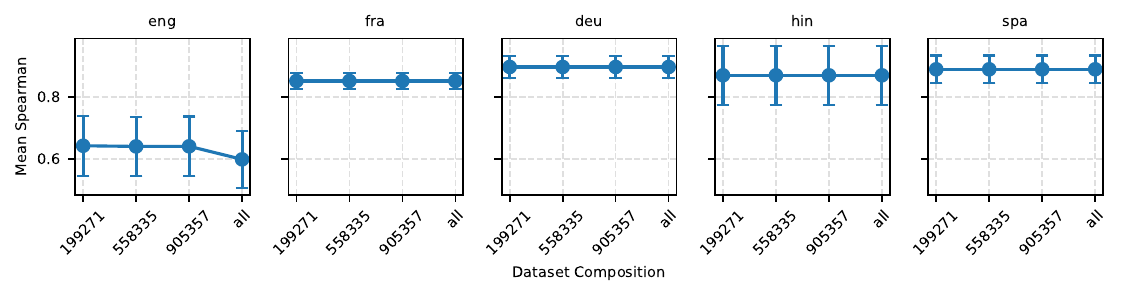}}
    \caption{
      Spearman correlation between the MTEB model rank and the rankings produced by different ranking schemes, averaged by dataset composition, on the STS task.
    }
    \label{fig:baseline_rank_comparison_sts}
  \end{center}
\end{figure*}

Across the clustering (Figure~\ref{fig:baseline_rank_comparison_clustering}), retrieval (Figure~\ref{fig:baseline_rank_comparison_retrieval}), and STS (Figure~\ref{fig:baseline_rank_comparison_sts}) tasks, a consistent pattern emerges in how closely the proposed ranking schemes align with the baseline MTEB ranking, though the absolute level of agreement differs substantially by task. In the clustering task, correlations are moderate to high but vary by language. English shows relatively low agreement (0.60–0.65) with a slight drop for the dataset composition including all datasets, while French exhibits higher and very stable correlation (around 0.85–0.90). German, Spanish, and Hindi have a single dataset in this task, resulting in a single dataset composition including that one dataset. While German and Spanish rankings show high correlation with the MTEB baseline, this correlation is lower (around 0.8) for Hindi. In the retrieval setting, German and Spanish reach the highest and most stable correlations (around 0.90–0.95), French and Hindi slightly lower (around 0.83–0.88), and English again much lower (around 0.55–0.58) with minimal sensitivity to dataset composition. In contrast, the STS task shows the most heterogeneous behavior. English remains low and even decreases for the dataset composition including all datasets (around 0.50–0.60), while French achieves higher correlation (up to 0.85). German, Hindi, and Spanish reach similarly high correlations (around 0.85–0.90), though with greater standard deviation, particularly for the Hindi language.

\begin{figure*}[ht]
  \begin{center}
    \centerline{\includegraphics[width=\textwidth]{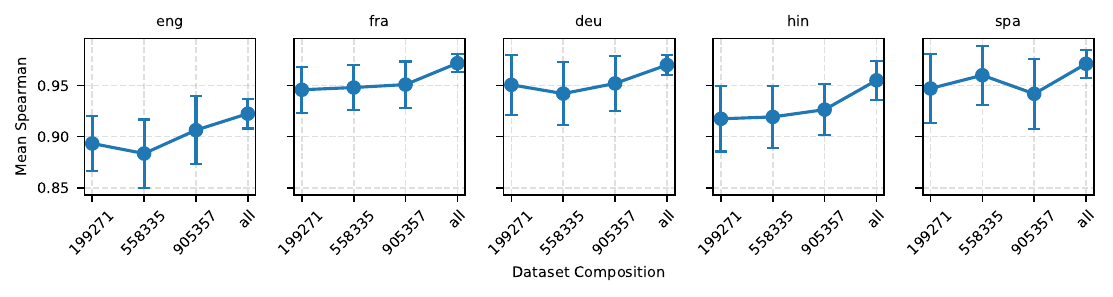}}
    \caption{
      Spearman correlation between the MTEB model rank and the rankings produced by different ranking schemes, averaged by dataset composition, on the bitext mining task.
    }
    \label{fig:baseline_rank_comparison_bitextmining}
  \end{center}
\end{figure*}

\begin{figure*}[ht]
  \begin{center}
    \centerline{\includegraphics[width=\textwidth]{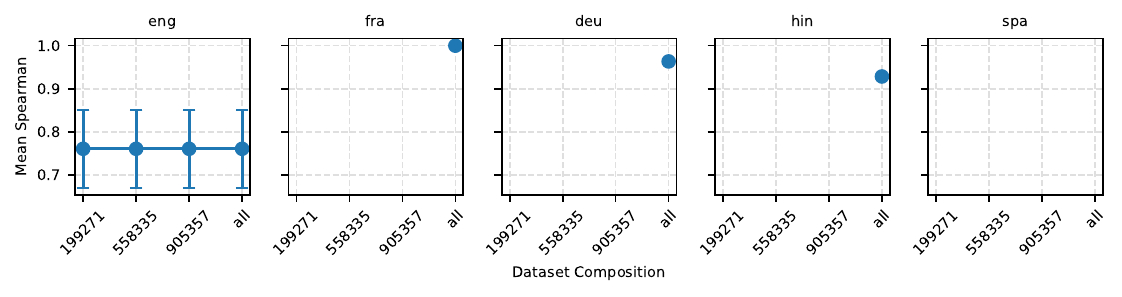}}
    \caption{
      Spearman correlation between the MTEB model rank and the rankings produced by different ranking schemes, averaged by dataset composition, on the reranking task.
    }
    \label{fig:baseline_rank_comparison_reranking}
  \end{center}
\end{figure*}

\begin{figure*}[ht]
  \begin{center}
    \centerline{\includegraphics[width=\textwidth]{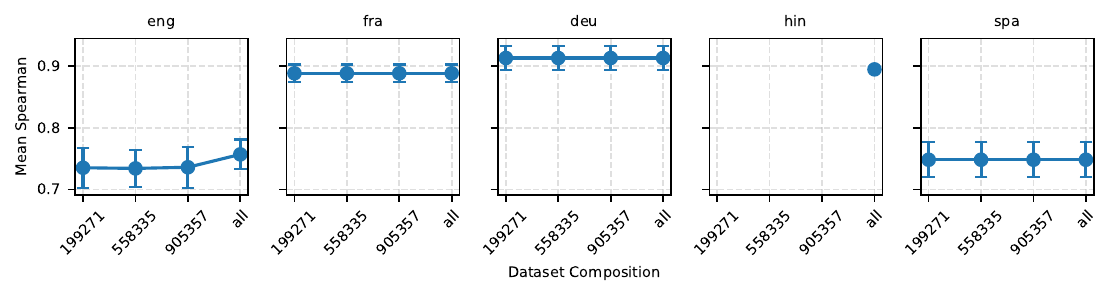}}
    \caption{
     Spearman correlation between the MTEB model rank and the rankings produced by different ranking schemes, averaged by dataset composition, on the pair classification task.
    }
    \label{fig:baseline_rank_comparison_pairclassification}
  \end{center}
\end{figure*}

\begin{figure*}[ht]
  \begin{center}
    \centerline{\includegraphics[width=\textwidth]{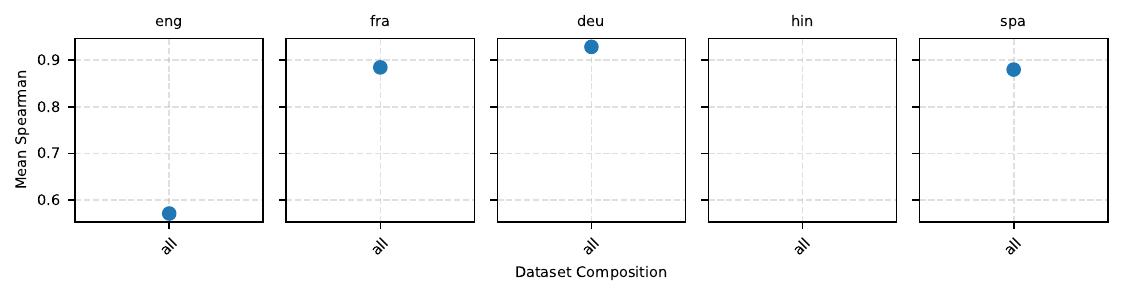}}
    \caption{
      Spearman correlation between the MTEB model rank and the rankings produced by different ranking schemes, averaged by dataset composition, on the multilabel classification task.
    }
    \label{fig:baseline_rank_comparison_multilabelclassification}
  \end{center}
\end{figure*}

\begin{figure*}[ht]
  \begin{center}
    \centerline{\includegraphics[width=\textwidth]{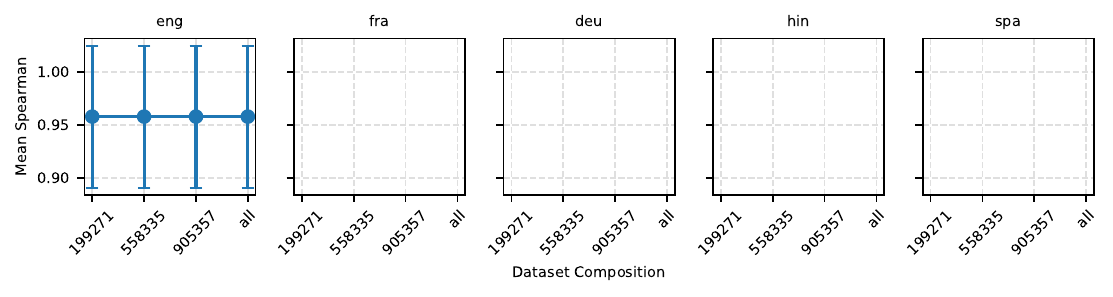}}
    \caption{
      Spearman correlation between the MTEB model rank and the rankings produced by different ranking schemes, averaged by dataset composition, on the instruction reranking task.
    }
    \label{fig:baseline_rank_comparison_instructionreranking}
  \end{center}
\end{figure*}

Across the remaining tasks, i.e., bitext mining (Figure~\ref{fig:baseline_rank_comparison_bitextmining}), reranking (Figure~\ref{fig:baseline_rank_comparison_reranking}), pair classification (Figure~\ref{fig:baseline_rank_comparison_pairclassification}), multilabel classification (Figure~\ref{fig:baseline_rank_comparison_multilabelclassification}), and instruction reranking (Figure~\ref{fig:baseline_rank_comparison_instructionreranking}), the Spearman correlation with the baseline varies notably by task. In bitext mining, agreement is consistently strong across all languages (around 0.90–0.97), with only minor variation across dataset compositions. The reranking task shows more mixed behavior. German and French have a single dataset, and the rankings maintain high correlation (around 0.88–0.92), Hindi (again with a single dataset) is moderately high, Spanish does not have datasets in this task so the analysis is inapplicable, and English is the weakest (around 0.76). In pair classification, results are heterogeneous, with German achieving very high correlation (around 0.91), French also high (around 0.88), while English is substantially lower (around 0.57), as well as Spanish (around 0.75), indicating strong language-dependent divergence from the baseline. Hindi has a single dataset in this task, showing correlation around 0.89. In the multilabel classification task all languages have a single datasets, except Hindi which has none. German has the highest correlation (around 0.93), Franch and Spanish around 0.88, while English the lowest (around 0.57). Finally, in instruction reranking only the English language has three datasets, while the remaning have no datasets. The correlation for English is around 0.96, with high standard deviation but minimal sensitivity to dataset composition.

Across all nine tasks, correlation with the MTEB baseline is generally high but clearly task-dependent, with the strongest and most consistent correlations observed in classification, multilabel classification, bitext mining, and retrieval, moderate agreement in clustering and reranking, and the weakest and most variable alignment in STS and pair classification. Across the five languages, French, German, and Spanish consistently show the highest and most stable correlation with the baseline, Hindi exhibits moderate but more variable behavior, and English consistently demonstrates the lowest correlation across nearly all tasks.

\subsection{Aggregation by Ranking Scheme}

\begin{figure*}[ht]
  \begin{center}
    \centerline{\includegraphics[width=0.5\textwidth]{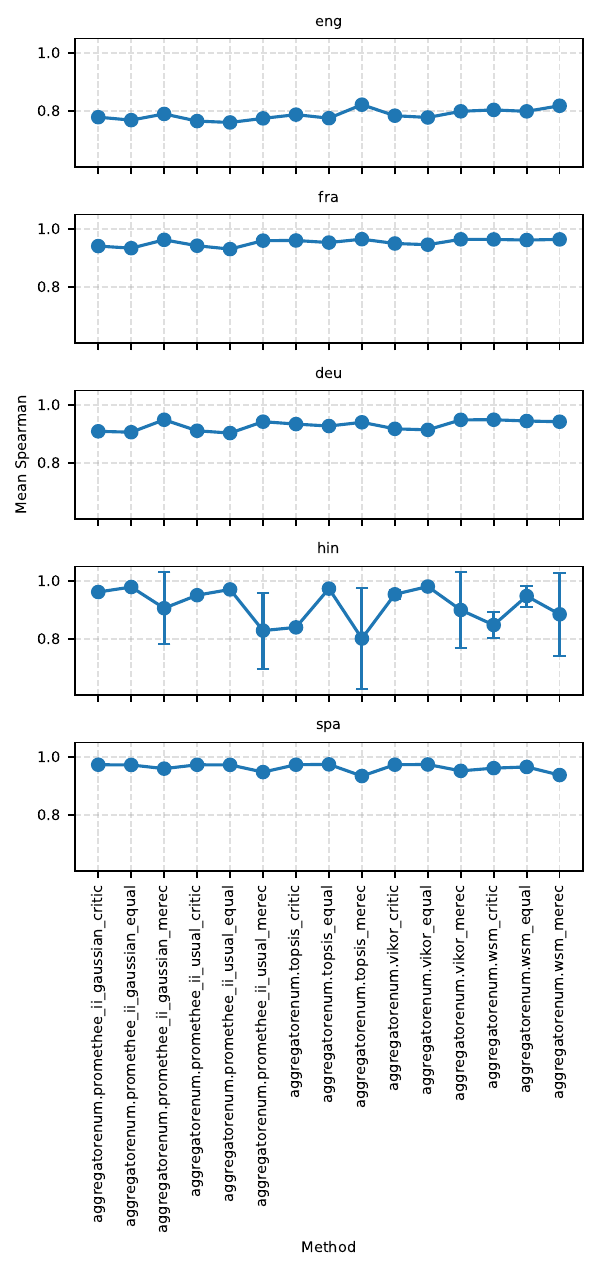}}
    \caption{
      Spearman correlation between the MTEB model rank and the rankings produced by different ranking schemes, averaged by ranking scheme, on the classification task.
    }
    \label{fig:baseline_rank_comparison_method_classification}
  \end{center}
\end{figure*}

In the classification task (Figure~\ref{fig:baseline_rank_comparison_method_classification}), the Spearman correlation with the MTEB baseline remains moderate to high across all ranking schemes for most languages, with French, German, and Spanish showing near-uniform values (0.93–0.97) regardless of the ranking scheme. English exhibits lower but still stable correlation (0.76–0.82), with only minor fluctuations across ranking schemes. In contrast, Hindi displays the greatest variability, with correlations ranging widely (0.80–0.98) and larger standard deviations, indicating sensitivity to the choice of ranking scheme.

\begin{figure*}[ht]
  \begin{center}
    \centerline{\includegraphics[width=0.5\textwidth]{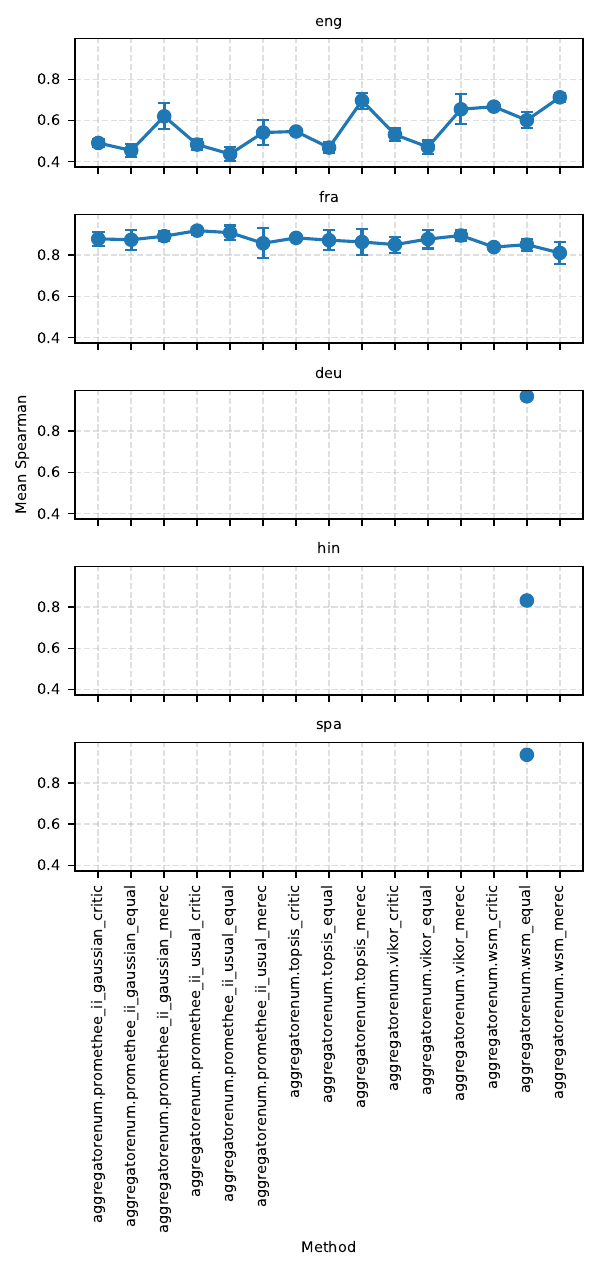}}
    \caption{
      Spearman correlation between the MTEB model rank and the rankings produced by different ranking schemes, averaged by ranking scheme, on the clustering task. Single point means that there is a single dataset for that language, and all ranking schemes produced the same results as the one reported.
    }
    \label{fig:baseline_rank_comparison_method_clustering}
  \end{center}
\end{figure*}

In the clustering task (Figure~\ref{fig:baseline_rank_comparison_method_clustering}), the Spearman correlation with the MTEB baseline shows greater sensitivity to the choice of aggregation method, particularly for English, where values vary widely (0.45–0.72) across ranking schemes. French remains relatively stable with consistent correlations (0.82–0.90), indicating limited impact of the ranking scheme, while German, Hindi, and Spanish each have a single dataset, so ranking correlation only for that single dataset is shown (for consistency reported under WSM method with equal weights). While German has the highest correlation of nearly 0.97, Spanish has lower of nearly 0.94, and Hindi the lowest of 0.83. 

\begin{figure*}[ht]
  \begin{center}
    \centerline{\includegraphics[width=0.5\textwidth]{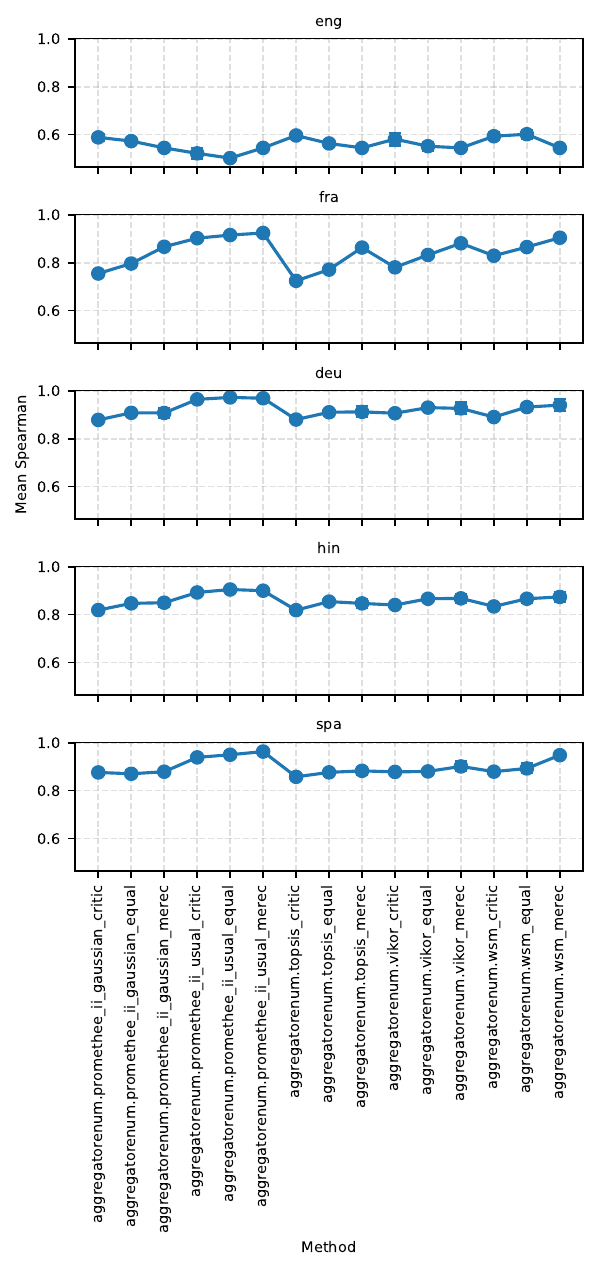}}
    \caption{
      Spearman correlation between the MTEB model rank and the rankings produced by different ranking schemes, averaged by ranking scheme, on the retrieval task.
    }
    \label{fig:baseline_rank_comparison_method_retrieval}
  \end{center}
\end{figure*}

In the retrieval task (Figure~\ref{fig:baseline_rank_comparison_method_retrieval}), the Spearman correlation with the MTEB baseline varies more noticeably across ranking schemes, with clear scheme-dependent fluctuations in several languages. English shows consistently low correlation (0.50–0.60) with modest variation, while French exhibits a wider range (0.72–0.92), including a pronounced decrease for certain schemes (e.g., TOPSIS with CRITIC weighting), indicating sensitivity to ranking scheme. German, Hindi, and Spanish maintain relativly high correlations (0.85–0.97), though all show some scheme-dependent variation, particularly around the same TOPSIS-based scheme. Overall, the choice of ranking scheme has a more visible impact on the correlation with the MTEB baseline in the retrieval task.

\begin{figure*}[ht]
  \begin{center}
    \centerline{\includegraphics[width=0.5\textwidth]{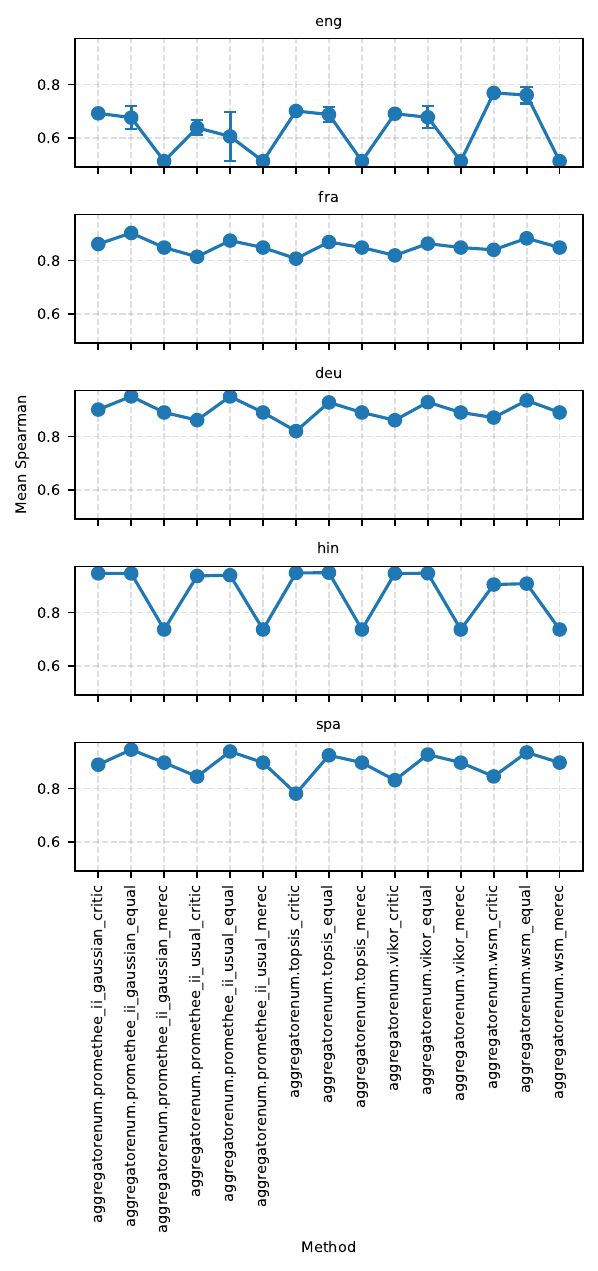}}
    \caption{
      Spearman correlation between the MTEB model rank and the rankings produced by different ranking schemes, averaged by ranking scheme, on the STS task.
    }
    \label{fig:baseline_rank_comparison_method_sts}
  \end{center}
\end{figure*}

In the STS task (Figure~\ref{fig:baseline_rank_comparison_method_sts}), the Spearman correlation with the MTEB baseline shows substantial variability across the ranking schemes for all languages, indicating a strong sensitivity to the choice of the ranking scheme. English exhibits the lowest and most unstable correlation (0.50–0.80), with frequent sharp drops for certain schemes, while French remains comparatively stable but still fluctuates (0.80–0.90). German maintains consistently high correlations (0.85–0.95) with moderate variation, whereas Hindi and Spanish show pronounced oscillations (0.73–0.97), reflecting scheme-dependent behavior. Overall, the ranking scheme plays a significant role in STS, affecting both the level and stability of correlation with the MTEB baseline across all five languages.

\begin{figure*}[ht]
  \begin{center}
    \centerline{\includegraphics[width=0.5\textwidth]{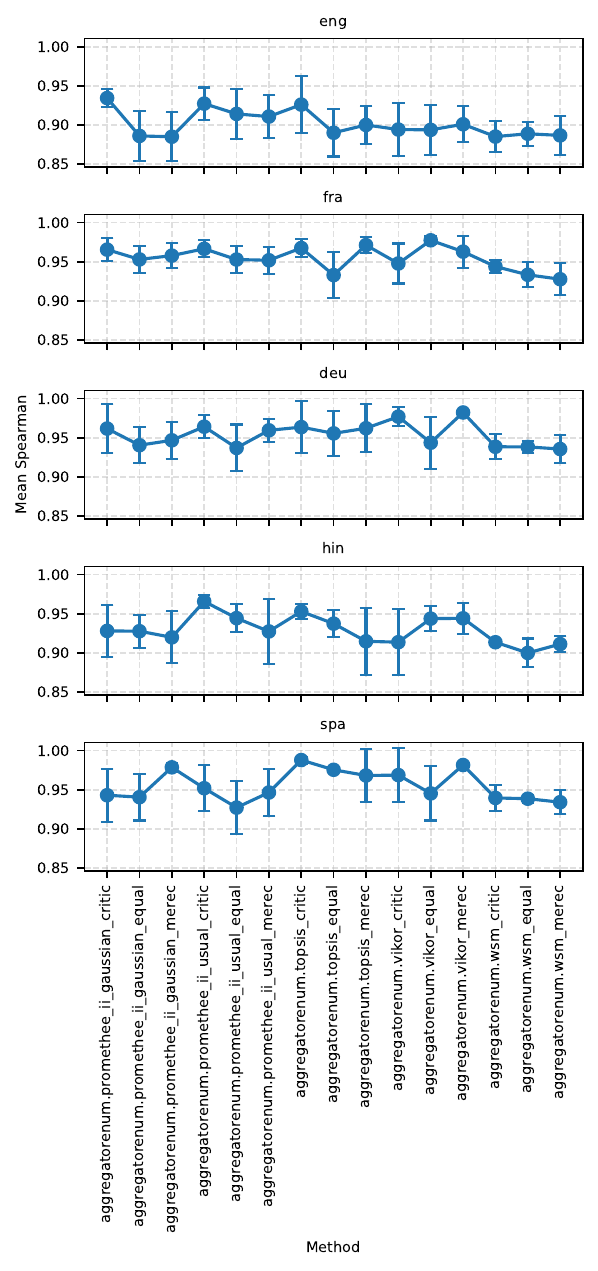}}
    \caption{
      Spearman correlation between the MTEB model rank and the rankings produced by different ranking schemes, averaged by ranking scheme, on the bitext mining task.
    }
    \label{fig:baseline_rank_comparison_method_bitextmining}
  \end{center}
\end{figure*}

In the bitext mining task (Figure~\ref{fig:baseline_rank_comparison_method_bitextmining}), the Spearman correlation with the MTEB baseline is high across all ranking schemes and languages (generally between 0.85 and 0.98). Moderate fluctuations across ranking schemes are visible, with more significant correlation drops for specific schemes. French, German, and Spanish maintain the highest correlations, while English and Hindi show slightly more variability but still remain within a high range. Overall, the choice of ranking scheme again has impact on the correlation in this task.

\begin{figure*}[ht]
  \begin{center}
    \centerline{\includegraphics[width=0.5\textwidth]{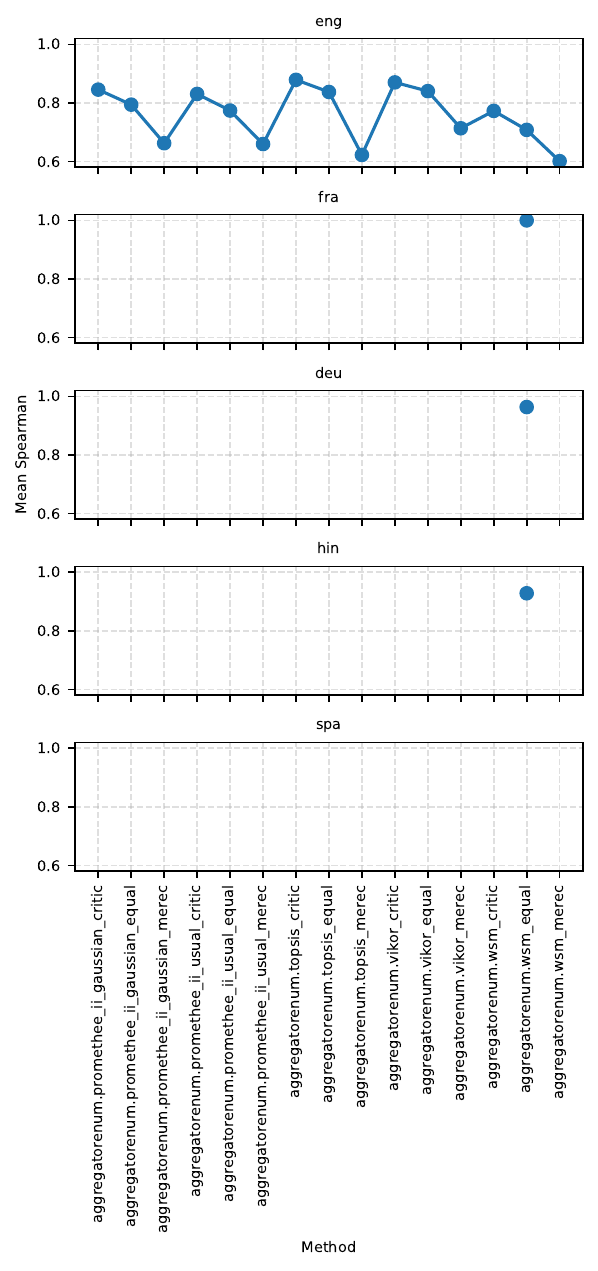}}
    \caption{
      Spearman correlation between the MTEB model rank and the rankings produced by different ranking schemes, averaged by ranking scheme, on the reranking task. Single point means that there is a single dataset for that language, and all ranking schemes produced the same results as the one reported.
    }
    \label{fig:baseline_rank_comparison_method_reranking}
  \end{center}
\end{figure*}

In the reranking task (Figure~\ref{fig:baseline_rank_comparison_method_reranking}), the Spearman correlation with the baseline shows strong method dependence for English, where values vary widely (0.60–0.88) across ranking schemes with several pronounced drops. For French, German, and Hindi, only a single dataset is available, all showing high correlation (0.93–1.0), while Spanish has no datasets in this task, so the analysis is inapplicable. Again the results suggest that ranking scheme can substantially impact the correlation with the MTEB baseline for the English language.

\begin{figure*}[ht]
  \begin{center}
    \centerline{\includegraphics[width=0.5\textwidth]{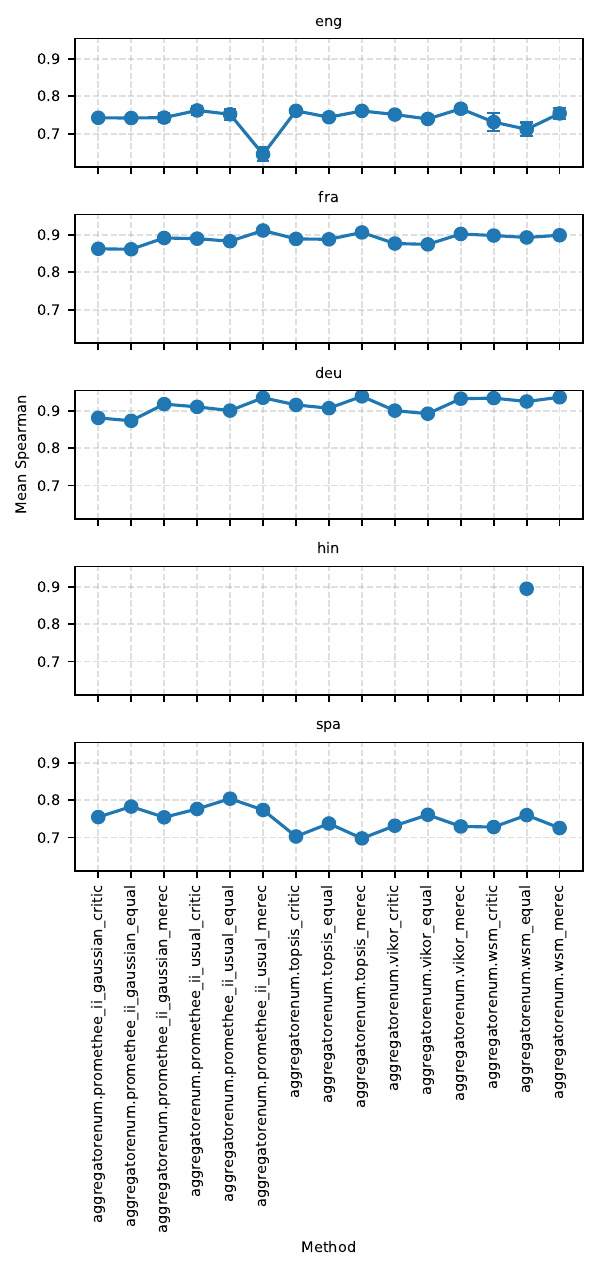}}
    \caption{
     Spearman correlation between the MTEB model rank and the rankings produced by different ranking schemes, averaged by ranking scheme, on the pair classification task. Single point means that there is a single dataset for that language, and all ranking schemes produced the same results as the one reported.
    }
    \label{fig:baseline_rank_comparison_method_pairclassification}
  \end{center}
\end{figure*}

In the pair classification task (Figure~\ref{fig:baseline_rank_comparison_method_pairclassification}), the Spearman correlation with the MTEB baseline is generally moderate to high but shows some variation across aggregation methods, particularly for English and Spanish. English exhibits relatively low and fluctuating correlation (0.65–0.77), including a clear drop for certain methods (e.g., PROMETHEE II with usual preference function and MEREC weighting method), while Spanish shows similar variability (0.70–0.80). In contrast, French and German maintain consistently high correlations (0.88–0.95) with only minor scheme-dependent changes, and Hindi has a single dataset with correlation value around 0.90.

\begin{figure*}[ht]
  \begin{center}
    \centerline{\includegraphics[width=0.5\textwidth]{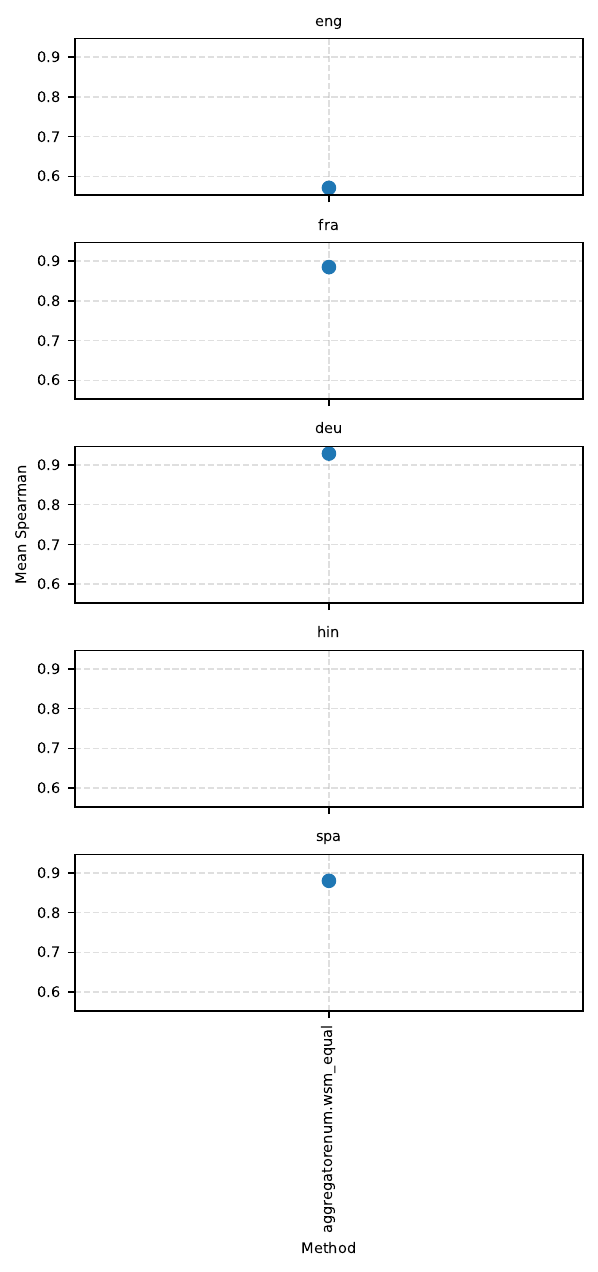}}
    \caption{
      Spearman correlation between the MTEB model rank and the rankings produced by different ranking schemes, averaged by ranking scheme, on the multilabel classification task. Single point means that there is a single dataset for that language, and all ranking schemes produced the same results as the one reported.
    }
    \label{fig:baseline_rank_comparison_method_multilabelclassification}
  \end{center}
\end{figure*}

In the multilabel classification task (Figure~\ref{fig:baseline_rank_comparison_method_multilabelclassification}), four languages (English, French, German, and Spanish) have a single dataset, while Hindi has no datasets. German shows the highest agreement (around 0.93), followed by French and Spanish (around 0.88), while English is substantially lower (0.57), indicating weaker correlation.

\begin{figure*}[ht]
  \begin{center}
    \centerline{\includegraphics[width=0.5\textwidth]{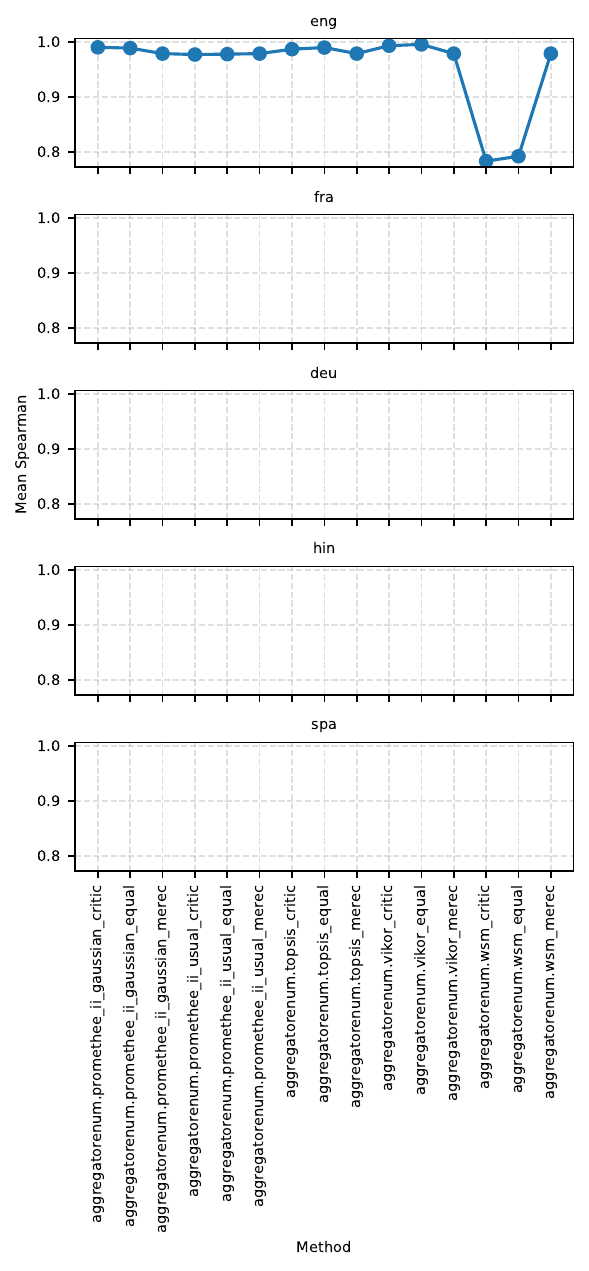}}
    \caption{
      Spearman correlation between the MTEB model rank and the rankings produced by different ranking schemes, averaged by ranking scheme, on the instruction reranking task. Empty plots indicate that there are no datasets available for that language.
    }
    \label{fig:baseline_rank_comparison_method_instructionreranking}
  \end{center}
\end{figure*}

In the instruction reranking task (Figure~\ref{fig:baseline_rank_comparison_method_instructionreranking}), there are datasets for the English language only, while no datasets for the other four languages. The Spearman correlation with the MTEB baseline is high for English (0.97–1.0) across most ranking schemes. However, a notable drop is observed for specific schemes (e.g., WSM variants), where the correlation decreases sharply to 0.78–0.80, highlighting ranking scheme sensitivity.

Overall, the impact of the ranking scheme on the Spearman correlation with the MTEB baseline varies substantially by task, ranging from negligible to highly pronounced. In tasks such as classification, multilabel classification, and bitext mining, the choice of ranking scheme has minimal effect on the correlation. In contrast, retrieval, clustering, and pair classification exhibit moderate sensitivity, where certain methods introduce visible fluctuations. The effect becomes most pronounced in STS, reranking, and instruction reranking, where correlations can vary widely across ranking schemes, including sharp drops for specific schemes, especially in languages like English. Overall, while many settings show stable correlation with the MTEB baseline regardless of the ranking scheme, there are clear cases where the choice of ranking scheme critically affects the resulting rankings.

\section{Practical guidance for hyperparameter choices}
\label{appendix:hyperparameters}

Based on the sensitivity analyses reported in Appendix~\ref{sec:sensitivty_correlation} and Appendix~\ref{app:sensitivty_top}, we offer the following practical recommendations for hyperparameter selection. For the correlation threshold $\tau$, we recommend $\tau = 0.9$ as the default. Lower thresholds ($\tau = 0.8$ or $0.85$) frequently merge datasets into a
single cluster, eliminating meaningful diversity in the resulting subsamples and limiting the dataset composition analysis.
Our sensitivity analysis (Appendix~\ref{sec:sensitivty_correlation}) confirms that rankings remain stable across all three thresholds (mean Spearman correlation $> 0.85$ in all task--language pairs), so the choice of $\tau$ does not affect conclusions; $\tau = 0.9$ simply produces the
clearest cluster structure.

For the top-$\eta$ parameter, we recommend $\eta = 10$. The top-5 and top-10 settings yield consistent task-agnostic conclusions (Appendix~\ref{app:sensitivty_top}), while the top-1 setting reflects
task-specific rather than task-agnostic behavior and is therefore not
appropriate for cross-task robustness analysis.

For the MCDM portfolio, practitioners wishing to minimize complexity can use WSM and PROMETHEE~II as a minimal representative pair, since these differ most fundamentally in their aggregation logic (value-based vs. outranking). The full portfolio of four methods is recommended for a comprehensive coverage of principal MCDM families (Appendix~\ref{app:portfolio}). Critically, these choices do not introduce opaque engineering decisions; they are fully documented, their sensitivity is quantified, and the results are stable across reasonable variations. This is precisely the transparency that simple averaging across datasets lacks - averaging embeds choices
(equal dataset weighting, metric comparability assumptions) but leaves them implicit and untested.

\end{document}